\documentclass[lettersize,journal]{IEEEtran}
\usepackage{amsmath,amsfonts}
\usepackage{algorithmic}
\usepackage{algorithm}
\usepackage{array}
\usepackage[caption=false,font=normalsize,labelfont=sf,textfont=sf]{subfig}
\usepackage{textcomp}
\usepackage{stfloats}
\usepackage{url}
\usepackage{verbatim}
\usepackage{graphicx}
\usepackage{cite}
\usepackage[percent]{overpic}
\usepackage{forest}
\usepackage{booktabs}
\usepackage{array}
\usepackage{multirow}
\usepackage{tabularx}
\usepackage{etoolbox}
\usepackage[table,xcdraw]{xcolor}
\usepackage{colortbl}
\usepackage{makecell}
\usepackage{hyperref}
\usepackage{fontawesome}
\usepackage[table,xcdraw]{xcolor}

\definecolor{red}{rgb}{0,0,0}
\definecolor{blue}{rgb}{0,0,0}
\newcommand{\fix}[1]{\textcolor{blue}{#1}}

\definecolor{customcyan}{HTML}{9DBCC6}

\usepackage[
enable,
labname={CMLab},
logo={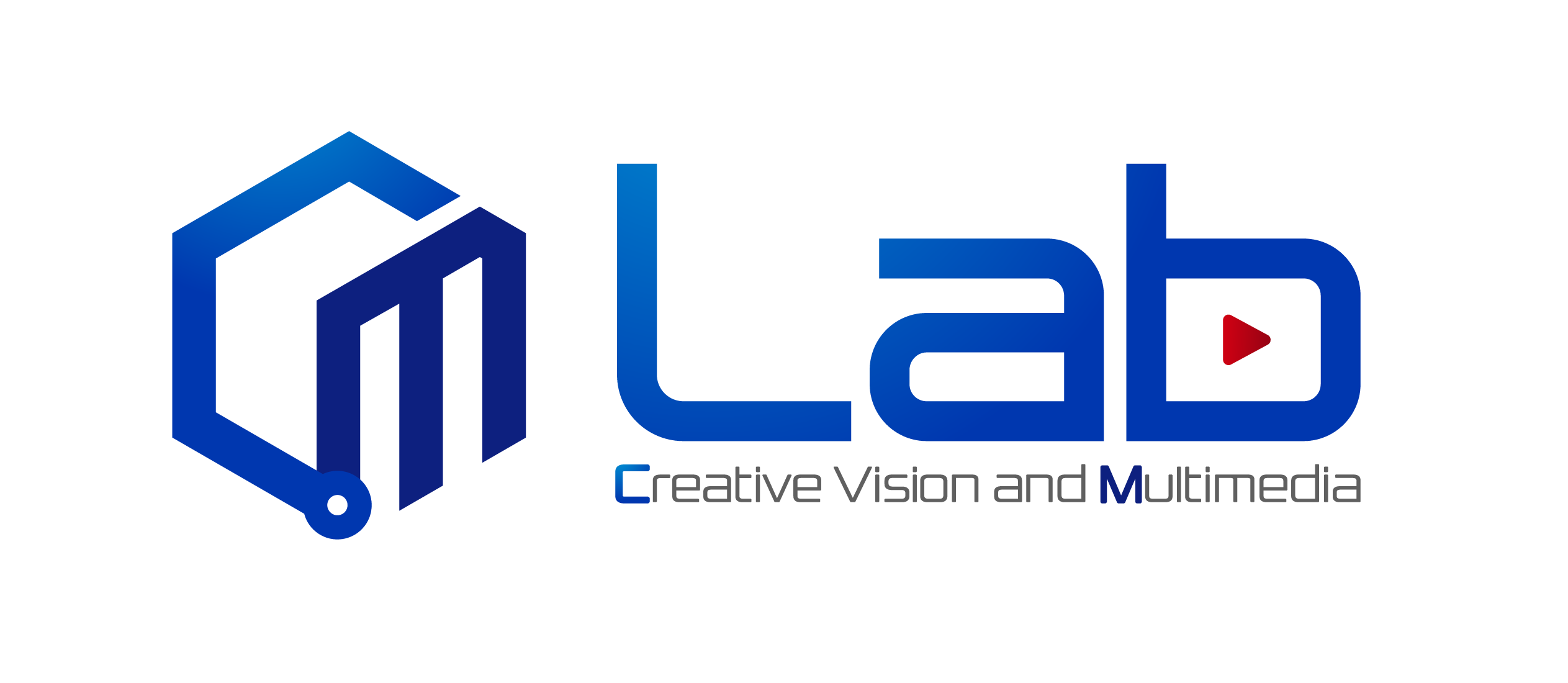},
titleicon={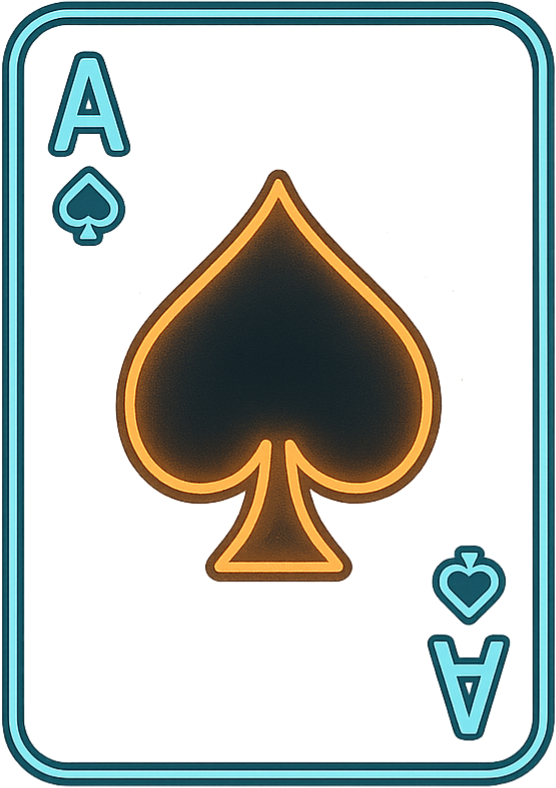},
logowidth={22mm},
projectpage={https://github.com/CMLab-Korea/Awesome-Video-Frame-Interpolation}
]{cmlab}

\begin{document}

\title{AceVFI: A Comprehensive Survey of Advances\\ in Video Frame Interpolation}


\markboth{IEEE Transactions on Circuits and Systems for Video Technology, 2026}%
{Kye \MakeLowercase{\textit{et al.}}: AceVFI: A Comprehensive Survey of Advances in Video Frame Interpolation}

\cmlabAuthors{Dahyeon Kye$^{1}$ \qquad Changhyun Roh$^{1}$ \qquad Sukhun Ko$^{1}$ \qquad Chanho Eom$^{2}$ \qquad Jihyong Oh$^{1, \dagger}$}
\cmlabAffiliations{$^{1}$CMLab, Chung-Ang University \qquad $^{2}$Perceptual AI LAB, Chung-Ang University}
\cmlabAuthorEmail{\{rpekgus, changhyunroh, looloo330, cheom, jihyongoh\}@cau.ac.kr}
\cmlabProjectPage{https://github.com/CMLab-Korea/Awesome-Video-Frame-Interpolation}

\maketitle

\begingroup
\renewcommand{\thefootnote}{}%
\footnotetext{Copyright~\textcopyright~2026 IEEE. Personal use of this material is permitted. However, permission to use this material for any other purposes must be obtained from the IEEE by sending an email to pubs-permissions@ieee.org.}%
\endgroup

\begingroup
\renewcommand{\thefootnote}{\fnsymbol{footnote}}%
\footnotetext[2]{Corresponding author.}%
\endgroup

\begin{abstract}
Video Frame Interpolation (VFI) is a core low-level vision task that synthesizes intermediate frames between existing ones while ensuring spatial and temporal coherence. Over the past decades, VFI methodologies have evolved from classical motion compensation-based approach to a wide spectrum of deep learning-based approaches, including kernel-, flow-, hybrid-, phase-, GAN-, Transformer-, Mamba-, and most recently, diffusion-based models. We introduce AceVFI, a comprehensive and up-to-date review of the VFI field, covering over 250 representative papers. We systematically categorize VFI methods based on their core design principles and architectural characteristics. Further, we classify them into two major learning paradigms: Center-Time Frame Interpolation (CTFI) and Arbitrary-Time Frame Interpolation (ATFI). We analyze key challenges in VFI, including large motion, occlusion, lighting variation, and non-linear motion. In addition, we review standard datasets, loss functions, evaluation metrics. We also explore VFI applications in other domains and highlight future research directions. This survey aims to serve as a valuable reference for researchers and practitioners seeking a thorough understanding of the modern VFI landscape. We maintain an up-to-date project page: \url{https://github.com/CMLab-Korea/Awesome-Video-Frame-Interpolation}.
\end{abstract}


\section{Introduction}
\label{sec:intro}

Video Frame Interpolation (VFI) aims to increase the temporal resolution (\textit{i.e.}, frame rate) of a video sequence by synthesizing one or more intermediate frames between given consecutive frames. This task serves a broad range of applications, including novel view synthesis~\cite{szeliski1999prediction, zhou2016view, flynn2016deepstereo, li2021neural}, slow-motion generation~\cite{jiang2018super, xue2019TOFlow, bao2019depth, xiang2020zooming, jeon2023dynamic, huang2024scale}, video compression~\cite{wu2018video, chun2020compressed, jia2022neighbor, takahashi2025coupled}, video prediction~\cite{mathieu2015deep, liu2017DVF, jia2022neighbor, hirose2024real}, and diverse generation tasks such as co-speech reenactment~\cite{liu2024tango}, human motion synthesis~\cite{liu2025video}, and facial animation~\cite{bigata2025keyface}.
In many of these scenarios, VFI is not merely an optional post-processing tool but a practically irreplaceable component: for slow-motion generation, high-frame-rate (HFR) capture typically requires specialized sensors, strong illumination, and large bandwidth or storage, which are often unavailable in consumer or legacy footage, so once a scene has been recorded at a low-frame-rate (LFR), additional real frames cannot be acquired retrospectively, so learned VFI becomes the only viable way to synthesize plausible in-between frames~\cite{ma2024timelens}. Likewise, in novel view synthesis and 4D scene reconstruction, densely sampling both viewpoints and time is often infeasible due to camera cost, calibration and synchronization overhead, and storage constraints, so VFI is used to temporally densify sparse input sequences, providing smoother motion trajectories and more continuous temporal coverage for downstream 3D/4D methods~\cite{lei2025mosca}. A key advantage of VFI lies in its ability to synthesize perceptually smooth and temporally coherent motion, aligning well with the temporal characteristics of the human visual system. HFR content reduces artifacts such as motion blur and judder~\cite{hou2022vfiqa, danier2022bvivfi}, thereby enhancing the visual quality in high-resolution (HR) and immersive media. This makes VFI particularly valuable in latency-sensitive and fidelity-critical scenarios such as sports broadcasting, interactive gaming, and virtual reality. Finally, in streaming pipelines, VFI also enables bandwidth-efficient video transmission by reconstructing intermediate frames locally, reducing the need to transmit full frame sequences~\cite{hou2022vfiqa}.
\IEEEpubidadjcol

\begin{figure}[!t]
    \centering
    \includegraphics[width=0.4\textwidth]{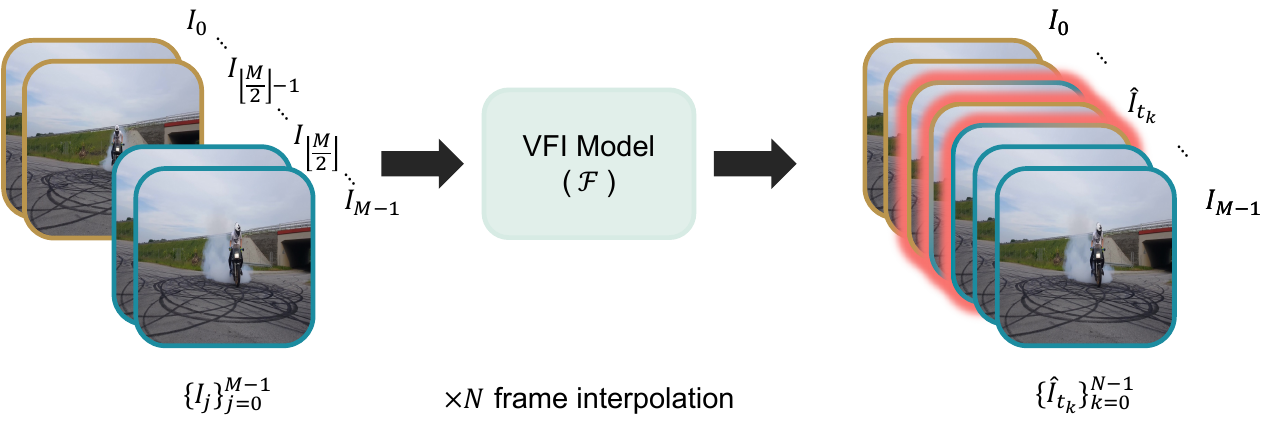}
    \caption{General process of VFI. Given $M$ consecutive input frames $\{I_j\}_{j=0}^{M-1}$, the VFI model $\mathcal{F}$ synthesizes one or more intermediate frames, producing an output sequence $\{\hat{I}_{t_k}\}_{k=0}^{N-1}$ with $t_k \in (\lfloor \frac{M}{2} \rfloor - 1, \lfloor \frac{M}{2} \rfloor)$, where $M \ge 2$ and $N \ge 1$.
    }
\label{fig:vfi}
\vspace{-0.5cm}
\end{figure}

As shown in Fig.~\ref{fig:vfi}, the general VFI formulation takes a sequence of $M$ consecutive frames $\{I_j\}_{j=0}^{M-1}$ to synthesize $N$ intermediate frames $\{\hat{I}_{t_k}\}_{k=0}^{N-1}$. The target time indices typically satisfy $t_k \in (\lfloor \frac{M}{2} \rfloor - 1, \lfloor \frac{M}{2} \rfloor)$. By generating $N$ frames within the central temporal interval, the video frame rate is increased by a factor of $N$. For instance, generating seven intermediate frames per interval transforms a 30fps video into 240fps. We begin by describing the most representative two-input-frame setting ($M=2$). Formally, given two adjacent frames $I_0$ and $I_1$, a VFI model $\mathcal{F}$ estimates the interpolated frame $\hat{I}_t$ at an arbitrary time $t \in (0,1)$:
\vspace{-0.1cm}
\begin{equation}
    \hat{I}_t = \mathcal{F}(I_0, I_1, t).
\end{equation}
\vspace{-0.9cm}

\subsection{Methodology Overview}
\label{subsec:methodology-overview}

\begin{figure}
  \centering
  \resizebox{0.99\columnwidth}{!}{
    \definecolor{colorRoot}{RGB}{36,118,129}
\definecolor{colorSec1}{RGB}{92,156,167}
\definecolor{colorSec2}{RGB}{124,180,186}
\definecolor{colorSec3}{RGB}{156,203,206}
\definecolor{colorSec4}{RGB}{181,217,215}
\definecolor{colorSec5}{RGB}{207,230,226}
\definecolor{colorSec6}{RGB}{227,240,238}
\definecolor{colorSec7}{RGB}{242,248,247}

\begin{forest}
for tree={
    grow=east,
    reversed=true,
    parent anchor=east,
    child anchor=west,
    rectangle,
    rounded corners,
    draw=colorRoot!50,
    edge+={black, line width=0.7pt},
    font=\sffamily\small,
    align=left,
    inner xsep=4pt,
    inner ysep=3pt,
    s sep=6pt,
    l sep=8mm,
    edge path={
        \noexpand\path [draw, \forestoption{edge}]
        (!u.parent anchor) -- ++(3.5mm,0) |- (.child anchor);
    }
},
where level=1{text width=10.5em}{},
where level=2{text width=13em}{},
where level=3{text width=11em}{},
[
    {Video Frame Interpolation},
    rotate=90,
    anchor=center,
    parent anchor=south, 
    fill=colorRoot!95,
    text=white,
    font=\sffamily\small,
    inner xsep=6pt,
    inner ysep=4pt
    [
        {I. Introduction (\S\ref{sec:intro})},
        fill=colorSec1
        [
            {Methodology Overview (\S\ref{subsec:methodology-overview})},
            fill=colorSec1!50
        ]
        [
            {General Pipeline of VFI (\S\ref{subsec:general-pipeline-of-vfi})},
            fill=colorSec1!50
        ]
    ]
    [
        {II. Methodology (\S\ref{sec:methodology})},
        fill=colorSec2
        [
            {Motion Compensation-based (\S\ref{subsec:motion-compensation-based})},
            fill=colorSec2!50
        ]
        [
            {Deep Learning-based (\S\ref{subsec:deep-learning-based})},
            fill=colorSec2!50
            [
                {Kernel-based (\S\ref{subsubsec:kernel-based})},
                fill=colorSec2!25
            ]
            [
                {Flow-based (\S\ref{subsubsec:flow-based})},
                fill=colorSec2!25
            ]
            [
                {Kernel- and Flow-based\\Combined (\S\ref{subsubsec:kernel-and-flow-combined})},
                fill=colorSec2!25
            ]
            [
                {Phase-based (\S\ref{subsubsec:phase-based})},
                fill=colorSec2!25
            ]
            [
                {GAN-based (\S\ref{subsubsec:gan-based})},
                fill=colorSec2!25
            ]
            [
                {Transformer-based (\S\ref{subsubsec:transformer-based})},
                fill=colorSec2!25
            ]
            [
                {Mamba-based (\S\ref{subsubsec:mamba-based})},
                fill=colorSec2!25
            ]
        ]
        [
            {Diffusion Model-based (\S\ref{subsec:diffusion-model-based})},
            fill=colorSec2!50
        ]
    ]
    [
        {III. Learning Paradigm (\S\ref{sec:learning-paradigm})},
        fill=colorSec3
        [
            {Center-Time Frame Interpolation\\(CTFI) (\S\ref{subsec:CTFI})},
            fill=colorSec3!50
        ]
        [
            {Arbitrary-Time Frame Interpolation\\(ATFI) (\S\ref{subsec:ATFI})},
            fill=colorSec3!50
        ]
        [
            {Training Strategy (\S\ref{subsec:training-strategy})},
            fill=colorSec3!30
            [
                {CTFI Training Strategy\\(CTFI-TS) (\S\ref{subsubsec:CTFI-TS})},
                fill=colorSec3!25
            ]
            [
                {ATFI Training Strategy\\(ATFI-TS) (\S\ref{subsubsec:ATFI-TS})},
                fill=colorSec3!25
            ]
        ]
        [
            {Loss Functions (\S\ref{subsec:loss-functions})},
            fill=colorSec3!50
            [
                {Reconstruction Loss (\S\ref{subsubsec:reconstruction-loss})},
                fill=colorSec3!25
            ]
            [
                {Perceptual Loss (\S\ref{subsubsec:perceptual-loss})},
                fill=colorSec3!25
            ]
            [
                {Adversarial Loss (\S\ref{subsubsec:adversarial-loss})},
                fill=colorSec3!25
            ]
            [
                {Flow Loss (\S\ref{subsubsec:flow-loss})},
                fill=colorSec3!25
            ]
        ]
    ]
    [
        {IV. VFI Challenges (\S\ref{sec:challenges})},
        fill=colorSec4
        [{Large Motion (\S\ref{subsec:large-motion})}, fill=colorSec4!50]
        [{Occlusion (\S\ref{subsec:occlusion})}, fill=colorSec4!50]
        [{Lighting Variation (\S\ref{subsec:lighting-variation})}, fill=colorSec4!50]
        [{Non-linear Motion (\S\ref{subsec:non-linear})}, fill=colorSec4!50]
    ]
    [
        {V. Datasets and\\Evaluation Metrics (\S\ref{sec:datasets-and-evaluation})},
        fill=colorSec5
        [
            {Datasets (\S\ref{subsec:datasets})},
            fill=colorSec5!50
            [{Triplet Datasets (\S\ref{subsubsec:tripelt-datasets})}, fill=colorSec5!25]
            [{Multi-frame Datasets (\S\ref{subsubsec:multiframe-datasets})}, fill=colorSec5!25]
        ]
        [{Data Augmentation (\S\ref{subsec:data-augmentation})}, fill=colorSec5!50]
        [
            {Evaluation Metrics (\S\ref{subsec:evaluation-metrics})},
            fill=colorSec5!50
            [{Image-level Metrics (\S\ref{subsubsec:image-level-metrics})}, fill=colorSec5!25]
            [{Perceptual Metrics (\S\ref{subsubsec:perceptual-metrics})}, fill=colorSec5!25]
            [{Video-level Metrics (\S\ref{subsubsec:video-level-metrics})}, fill=colorSec5!25]
        ]
        [{Summary of Comparisons (\S\ref{subsec:comparison})}, fill=colorSec5!50]
    ]
    [
        {VI. Applications (\S\ref{sec:applications})},
        fill=colorSec6
        [{Event-based VFI (\S\ref{subsec:event-based-vfi})}, fill=colorSec6!50]
        [{Cartoon VFI (\S\ref{subsec:cartoon-vfi})}, fill=colorSec6!50]
        [{Medical Image VFI (\S\ref{subsec:medical-image-vfi})}, fill=colorSec6!50]
        [{Joint VFI with LLV task (\S\ref{subsec:joint-task})}, fill=colorSec6!50]
    ]
    [
        {VII. Future Research\\Directions (\S\ref{sec:future})},
        fill=colorSec7
        [{Video Streaming Service (\S\ref{subsec:video-streaming-service})}, fill=colorSec7!50]
        [{All-in-One\\LLV Video Restoration (\S\ref{subsec:all-in-one-llv-video-restoration})}, fill=colorSec7!50]
        [{3D and\\4D Scene Understanding (\S\ref{subsec:3d-and-4d-scene-understanding})}, fill=colorSec7!50]
        [{Physics-Informed VFI\\for Extreme Environments (\S\ref{subsec:physics-informed})}, fill=colorSec7!50]
    ]
    [
        {VIII. Conclusion (\S\ref{sec:conclusion})},
        fill=colorSec7
    ]
]
\end{forest}
  }
  \caption{Overview of the survey structure.}
  \vspace{-0.5cm}
  \label{fig:overview}
\end{figure}

VFI methodologies can be broadly grouped into classical motion compensation-based~\cite{yu2013multi, haavisto1989fractional, nakaya1994motion, castagno1996method, lee2002adaptive, ha2004motion, choi2007motion, kang2008motion, huang2008multistage, wang2010motion1, wang2010motion2}, deep learning-based~\cite{park2020robust, jaderberg2015spatial, liu2017DVF, liu2019cyclicgen, bao2019depth, yuan2019zoom, reda2019unsupervised, xu2019quadratic, chi2020all, niklaus2020softmax, liu2020enhanced, zhang2020flexible, sim2021xvfi, hu2022many, kong2022ifrnet, park2021asymmetric, huang2022rife, shangguan2022learning, reda2022film, niklaus2023splatting, park2023biformer, jin2023unified, li2023amt, zhang2023extracting, hu2024iqvfi, jeong2024ocai, guo2024generalizable, wu2024perception, zhong2024clearer, jin2025unified, long2016learning, niklaus2017adaconv, niklaus2017sepconv, zhu2019deformable, peleg2019net, choi2020cain, cheng2020video, shi2021video, cheng2021multiple, ding2021cdfi, chen2021pdwn, danier2022enhancing, ding2022MSEConv, kalluri2023flavr, zhou2023exploring, niklaus2018context, bao2019memc, park2020bmbc, lee2020adacof, gui2020featureflow, niklaus2021revisiting, van2017FIGAN, xiao2020multi, xue2020frame, tran2020efficient, yuan2019zoom, danier2022stmfnet, meyer2015phase, meyer2018phasenet, shi2022video, lu2022video, zhang2023L2BEC2, liu2024sparse, gu2021efficiently, gu2023mamba, zhang2024vfimamba, jeong2025lc, briedis2025controllable, Wu_2023_BMVC}, and generative frameworks such as diffusion models (DMs)~\cite{wang2024framer, jain2024video, deng2025beyond, zhang2025arbitrary, tanveer2025multicoin, koren2017frame, tran2020efficient, tran2022video, wen2018generating, voleti2022mcvd, danier2024ldmvfi, jain2024video, huang2024motion, xing2024dynamicrafter, xing2024tooncrafter, shen2024dreammover, wang2024generative, feng2024explorative, lyu2024brownian, zhu2024generative, yang2024vibidsampler, zhang2025motion, zhang2025eden, hai2025hierarchical, hur2025high, wan2024unipaint, hwang2025diffuseslide, yang2024zerosmooth, lyu2025tlb, chen2025sci, lew2025disentangled, guo2025controllable, hong2025semantic}. The motion compensation-based approach dominated the pre-deep-learning era, offering a two-stage strategy: estimating motion explicitly and warping frames accordingly. While effective for simple motion, its reliance on hand-crafted rules and block-based assumptions limits its ability to handle occlusions and complex, non-rigid dynamics. With the advent of convolutional neural networks (CNNs)~\cite{krizhevsky2012cnn}, VFI shifted toward deep learning-based, replacing heuristic pipelines with end-to-end architectures. As a result, they significantly improve robustness under diverse and challenging conditions. Further methodological details are discussed in Sec.~\ref{subsec:deep-learning-based}. More recently, DMs have been introduced as a generative framework for VFI, framing the task as a conditional denoising process rather than a deterministic frame prediction. This expands the scope of VFI into the emerging concept of \textit{Generative Inbetweening}~\cite{feng2024explorative, wang2024generative, hong2025semantic}, enabling uncertainty-aware interpolation and semantically diverse frame synthesis. This shift not only enhances robustness in ambiguous motion scenarios but also opens the door to multi-modal guidance (\textit{e.g.}, text, depth, or motion priors), redefining the role of VFI in creative and interactive video generation.

In parallel to our work, two prior surveys~\cite{parihar2022comprehensive, dong2023video} have reviewed VFI techniques, focusing respectively on traditional interpolation methods and early deep learning–based approaches. 
Compared with these surveys, our paper provides a broader and more up-to-date coverage, including recent Transformer-, Mamba-, and DM-based VFI models that were not available or only briefly discussed in earlier works. 
Methodologically, we introduce a finer-grained taxonomy and explicitly relates each method to its underlying motion modeling strategy. 
In addition, we organize existing approaches through a learning-paradigm perspective that distinguishes Center-Time Frame Interpolation (CTFI) from Arbitrary-Time Frame Interpolation (ATFI), and connect these paradigms to their typical training strategies and loss formulations. 
Finally, our survey systematically compiles VFI datasets and evaluation metrics, providing a practical resource that complements the conceptual taxonomy and goes beyond the scope of previous VFI surveys.

\textbf{Overview.} Fig.~\ref{fig:overview} shows the overall structure of this paper. Sec.~\ref{sec:methodology} analyzes methodological taxonomies of VFI. Sec.~\ref{sec:learning-paradigm} introduces and compares the two principal learning paradigms of VFI, and further examines their corresponding training strategies and loss functions. Sec.~\ref{sec:challenges} discusses major challenges in VFI, along with how recent methods address them. Sec.~\ref{sec:datasets-and-evaluation} reviews common datasets and evaluation metrics. Section~\ref{sec:applications} explores applications of VFI across diverse domains. Finally, Sec.~\ref{sec:future} presents future research directions of VFI.
\begin{figure}[!t]
    \centering
    \includegraphics[width=0.9\linewidth]{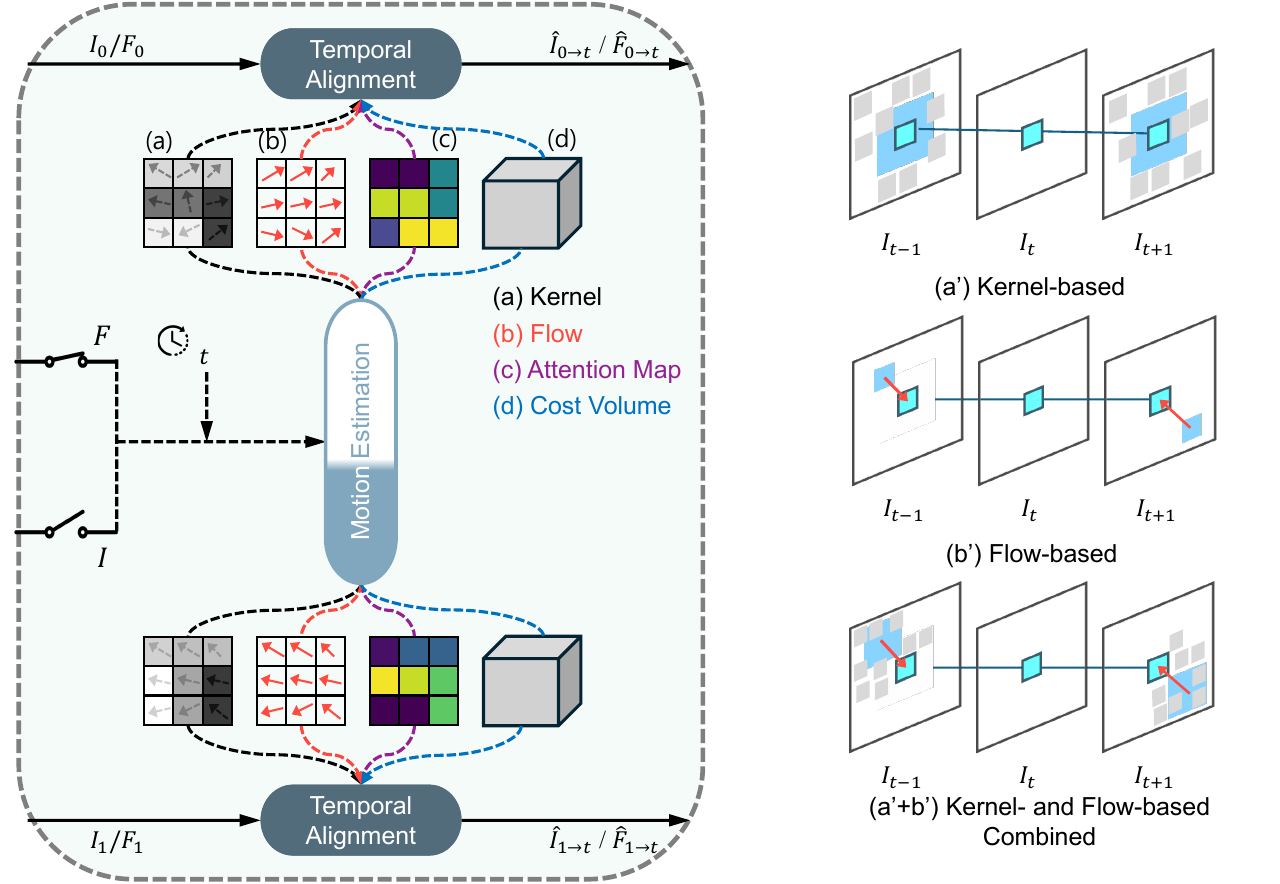} 
    \caption{Temporal alignment strategies. Input frames $(I_0, I_1)$ or features $(F_0, F_1)$ are aligned toward target time $t$ using four strategies: (a) kernel-based, (b) flow-based, (c) attention-based, and (d) cost volume-based. On the right, (a') (kernel-based), the blue square denotes the fixed kernel support window centered at the output location in $I_t$, while the gray patches indicate the actual sampling positions in $I_{t-1}$ and $I_{t+1}$ gathered via learned offsets, so that motion is encoded \emph{implicitly} through the offset pattern and kernel weights; (b') (flow-based), the light-blue marks the explicit reference location reached by a displacement vector, showing an \emph{explicit} motion field; (a'+b') shows the combined design, where an explicit flow first transports the support window toward a reference region and a local kernel is then applied around that flow-guided position for refinement.}
\label{fig:temporal_alignment}
\vspace{-0.5cm}
\end{figure}
\vspace{-12pt}

\begin{figure*}
    \centering
    \begin{overpic}[width=0.75\linewidth]{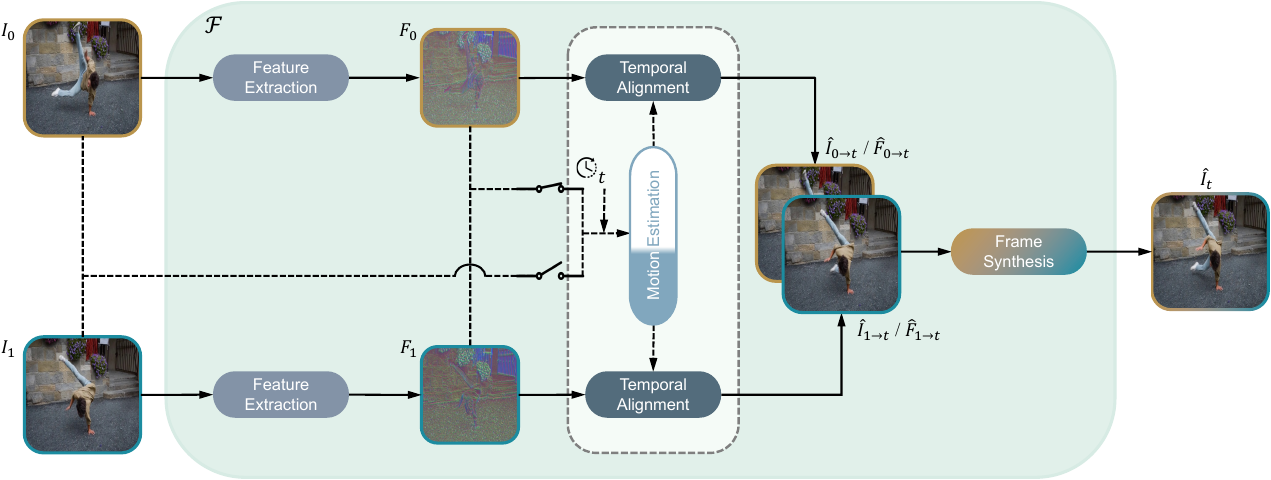}
        \put(44,36){%
        {\sffamily\scriptsize\textcolor{black}{Fig.~\ref{fig:temporal_alignment}}}
        }
    \end{overpic}
    \caption{General pipeline of VFI. Given two input frames $I_0$ and $I_1$, deep features $F_0$ and $F_1$ are first extracted. The features or pixels are then temporally aligned to the target time $t$ using estimated motion, producing $\hat{F}_{0 \rightarrow t}$, $\hat{F}_{1 \rightarrow t}$ or $\hat{I}_{0 \rightarrow t}$, $\hat{I}_{1 \rightarrow t}$. A Frame Synthesis module blends the aligned inputs to produce the final frame $\hat{I}_t$.}
\label{fig:pipeline}
\vspace{-0.5cm}
\end{figure*}
\vspace{-6pt}

\subsection{General Pipeline of VFI}
\label{subsec:general-pipeline-of-vfi}

The general VFI pipeline consists of four stages: \textbf{(i) Feature Extraction.}
Input frames $I_0$ and $I_1$ are passed through a feature extraction network~\cite{simonyan2014very, ronneberger2015u, cciccek20163d, xiang2020zooming}, yielding representations $F_0$ and $F_1$ for subsequent motion reasoning~\cite{nottebaum2022efficient}. \textbf{(ii) Motion Estimation.}  
Temporal correspondence (\textit{i.e.}, motion) is estimated either explicitly via optical flow~\cite{horn1981determining, niklaus2020softmax, sim2021xvfi} or implicitly using learned kernels~\cite{niklaus2017adaconv, niklaus2017sepconv}, phase cues~\cite{meyer2015phase, meyer2018phasenet}, attention maps~\cite{choi2020cain, shi2022video}, or cost volumes~\cite{park2020bmbc, park2021asymmetric}. \textbf{(iii) Temporal Alignment.}  
The estimated motion is used to temporally align the input pixels or features to the target time $t$, generating $\hat{F}_{0 \rightarrow t}$, $\hat{F}_{1 \rightarrow t}$ or $\hat{I}_{0 \rightarrow t}$, $\hat{I}_{1 \rightarrow t}$. As shown in Fig.~\ref{fig:temporal_alignment}, four major alignment strategies are employed. Although all of them rely on motion estimation, they differ in how motion is represented and applied during alignment. \textit{Kernel-based} alignment (Fig.~\ref{fig:temporal_alignment}~(a),(a')) aggregates local or non-local information from the inputs using  learned, spatially-adaptive kernels. These kernels implicitly encode motion by adapting their spatial weights based on local context, allowing motion-aware alignment without explicit flow estimation. As shown in Fig.~\ref{fig:temporal_alignment}~(a'), the network effectively selects reference locations by assigning larger weights to pixels that contribute most to the output, but the reachable motion range is constrained by the kernel support, making very large displacements harder to capture without excessively increasing the kernel size. \textit{Flow-based} alignment (Fig.~\ref{fig:temporal_alignment}~(b),(b')) warps inputs guided by the estimated flow. Forward warping~\cite{niklaus2020softmax} maps source pixels (\emph{i.e.}, input pixels) to their estimated locations in the target frame. Backward warping~\cite{jaderberg2015spatial} samples from the source based on coordinates in the target frame, effectively pulling information from the source toward the desired time. In Fig.~\ref{fig:temporal_alignment}~(b'), this is visualized as an explicit displacement vector that points from each output query to a reference position in the input frame, which is efficient and well suited to large motion but aggregates information from only a small number of source samples at each step, limiting flexibility under complex or noisy motion. The hybrid kernel-and-flow-based design in Fig.~\ref{fig:temporal_alignment}~(a'+b') combines these views by first using an explicit flow vector to transport the sampling window toward an appropriate reference region and then applying a local kernel around that flow-guided position, thereby increasing the number of effective reference samples while preserving the interpretability of the underlying motion field (Sec.~\ref{subsubsec:kernel-and-flow-combined}). \textit{Attention-based} alignment (Fig.~\ref{fig:temporal_alignment}~(c)) aggregates features based on attention-weighted correspondences~\cite{shi2022video, zhang2023extracting}. By computing soft correspondences between elements across input frames, this strategy can adaptively focus on semantically relevant regions and align contents even across large spatial-temporal gaps. \textit{Cost volume-based} alignment (Fig.~\ref{fig:temporal_alignment}~(d)) constructs dense similarity volumes between feature maps, enabling fine-grained correspondence modeling across space and time. \textbf{(iv) Frame Synthesis.}  
Finally, the aligned inputs are blended to synthesize the target frame $\hat{I}_t$ using simple averaging, weighted blending, or synthesis networks~\cite{chi2020all} as shown in Fig.~\ref{fig:pipeline}.

\section{Methodology}
\label{sec:methodology}

\subsection{Motion Compensation-based}
\label{subsec:motion-compensation-based}

Before the advent of deep-learning, VFI was mainly tackled via \textit{Motion-Compensated Frame Interpolation} (MCFI)~\cite{nakaya1994motion, ha2004motion, choi2007motion, huang2008multistage} or \textit{Frame Rate Up-Conversion} (FRUC)~\cite{haavisto1989fractional, castagno1996method, lee2002adaptive, kang2008motion, wang2010motion1, wang2010motion2}. These approaches, prevalent from the late 1990s through the early 2000s, estimate motion explicitly using block matching or parametric models, then synthesize intermediate frames by warping input frames based on the estimated motion fields. 

A typical MCFI pipeline involves two key steps: (i) block-based motion estimation and (ii) pixel-level warping for frame synthesis. In block-based estimation, each frame is partitioned into fixed-size rectangular blocks under the assumption of uniform motion within each block. While computationally efficient, this design often fails to capture non-rigid or object-specific motion, often resulting in artifacts such as holes (due to occlusions) and overlaps (due to many-to-one mappings). Rooted in classical video coding frameworks~\cite{jain1981displacement}, MCFI methods emphasize speed and simplicity, but inherently lack the capacity to handle fine-grained, non-linear motion. To address these issues, various extensions have been proposed, including multi-stage motion estimation~\cite{huang2008multistage}, adaptive motion models~\cite{lee2002adaptive}, and occlusion-aware warping~\cite{wang2010motion2}. Intermediate frame synthesis was generally performed through block-wise projection or forward warping, using the estimated motion vectors to guide the placement of each pixel. However, these methods typically operate in the pixel or block space without modeling complex motion patterns, and thus struggle to maintain spatial consistency under non-linear dynamics.

Despite their limited robustness in handling complex dynamics, MCFI and FRUC methods~\cite{haavisto1989fractional, nakaya1994motion, castagno1996method, lee2002adaptive, ha2004motion, choi2007motion, kang2008motion, huang2008multistage, wang2010motion1, wang2010motion2} laid the conceptual foundation for modern VFI. Their core principle, explicit motion estimation followed by motion-compensated warping, remains central to many modern learning-based models and is now enhanced with deep feature representations and end-to-end training. Importantly, classical motion-compensated strategies introduced valuable insights into the inductive biases that shape modern VFI architectures. Concepts such as motion locality, piecewise rigidity, and spatial warping, which originated from block-based estimation, are implicitly retained in modern mechanisms like deformable convolutions~\cite{dai2017dcn, zhu2019deformable} and local attention~\cite{vaswani2017attention}. Furthermore, challenges faced by early approaches, such as occlusion handling and motion discontinuity, have directly motivated the development of occlusion-aware blending, bidirectional flow formulations in contemporary VFI models. In this light, traditional motion models serve as both a historical foundation and conceptual framework for the progressive development of VFI architectures.
\vspace{-6pt}

\subsection{Deep Learning-based}
\label{subsec:deep-learning-based}

\subsubsection{\textbf{Kernel-based}}
\label{subsubsec:kernel-based}

\begin{figure}[!t]
    \centering
    \begin{overpic}[width=0.95\linewidth]{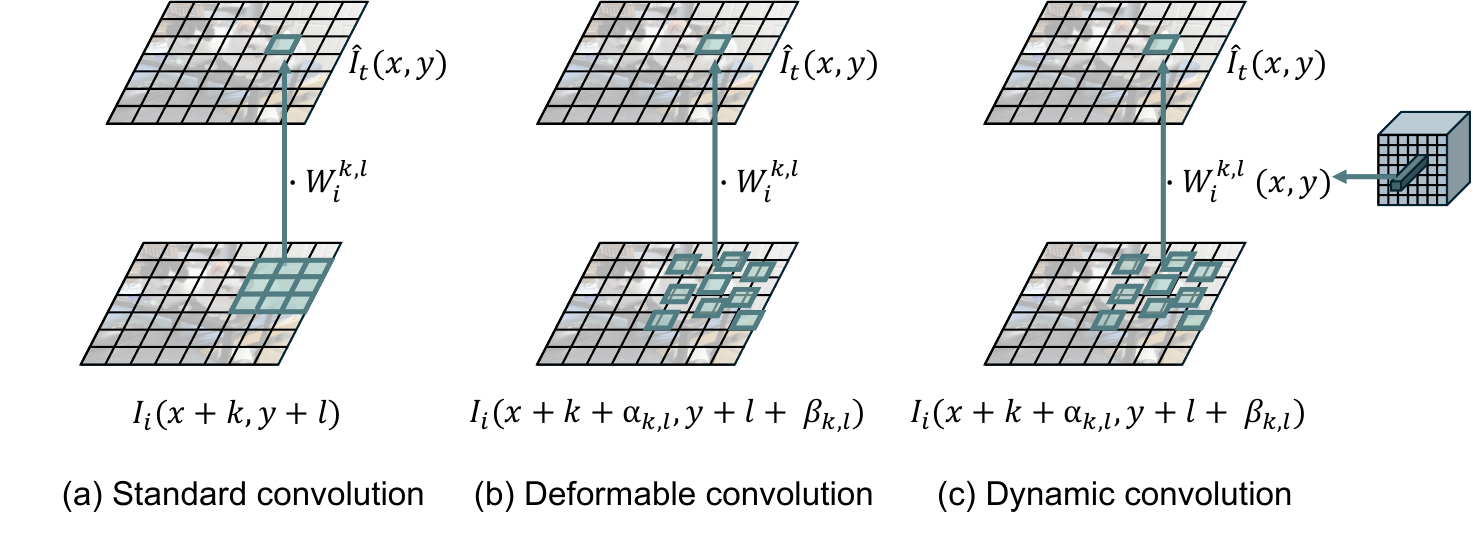}
        \put(12,37){%
        {\sffamily\tiny\textcolor{black}{Eq~(\ref{eq:standard-kernel})}}
        }
        \put(43,37){%
        {\sffamily\tiny\textcolor{black}{Eq~(\ref{eq:deformable-kernel})}}
        }
        \put(76,37){%
        {\sffamily\tiny\textcolor{black}{Eq~(\ref{eq:dynamic-kernel})}}
        }
    \end{overpic}
    \vspace{-0.3cm}
    \caption{Comparison of different convolution types. (a) Standard convolution samples at a fixed grid location $(x+k, y+l)$.
(b) Deformable convolution introduces learnable offsets $(\alpha_{k,l}, \beta_{k,l})$, enabling adaptive sampling at $(x+k+\alpha_{k,l},\ y+l+\beta_{k,l})$.
(c) Dynamic convolution further generalizes this by predicting the kernel weights $W_i^{k,l}(x, y)$ dynamically for each output position, allowing for spatially-variant filtering.}
    \label{fig:convolutions}
    \vspace{-0.5cm}
\end{figure}

Kernel-based VFI methods~\cite{niklaus2017adaconv, niklaus2017sepconv, niklaus2018context, peleg2019net, zhu2019deformable, bao2019memc, cheng2020video, park2020bmbc, lee2020adacof, gui2020featureflow, xiang2020zooming, niklaus2021revisiting, shi2021video, cheng2021multiple, ding2021cdfi, chen2021pdwn, danier2022enhancing, ding2022MSEConv, kalluri2023flavr, zhou2023exploring, zhang2023L2BEC2, shen2024ladder} synthesize intermediate frames by predicting spatially-adaptive convolutional \textit{kernels}, which are applied to local patches extracted from the input frames. Motion information is implicitly encoded in these kernel weights, enabling motion-aware pixel aggregation. In other words, this approach still estimates and exploits motion, but represents it implicitly through the spatial pattern and support of the learned kernels rather than as an explicit dense flow field or motion-vector map. A standard kernel-based interpolation can be formulated as
\begin{equation}
    \hat{I}(x, y) = \sum_{i=0}^{N-1}\sum_{k=0}^{R-1} \sum_{l=0}^{R-1} W_{k, l}I_i(x + k, y + l),
\label{eq:standard-kernel}
\end{equation}
where $N$ is the number of input frames, $R$ is the kernel size, and $W_{k,l}$ denotes the learned kernel weight at offset $(k, l)$, as shown in Fig.~\ref{fig:convolutions}~(a). AdaConv~\cite{niklaus2017adaconv} utilizes a U-Net-like architecture~\cite{ronneberger2015u} to predict spatially-varying 2D kernels for each output pixel. This enables local, pixel-wise motion-aware aggregation that can implicitly handle both alignment and occlusion~\cite{zhou2023exploring}. SepConv~\cite{niklaus2017sepconv} further reduces the computational overhead by decomposing the 2D kernel into separable 1D kernels:
\begin{equation}
    W = W_v * W_h,
\end{equation}
where $W_v \in \mathbb{R}^{R \times 1}$ and $W_h \in \mathbb{R}^{1 \times R}$ are vertical and horizontal 1D kernels respectively. The $*$ denotes the outer product between the two 1D kernels, resulting in a full 2D kernel $W \in \mathbb{R}^{R \times R}$. This reduces the number of learnable parameters from $R^2$ to $2R$, while preserving a comparable receptive field. Despite their simplicity, these methods are inherently limited in handling large displacements due to their fixed receptive fields~\cite{yuan2019zoom}. Such constraints stem from the content-agnostic nature of CNNs, which uniformly apply learned filters across spatial locations~\cite{shi2022video}. While this weight-sharing inductive bias proves effective in recognition tasks, it becomes suboptimal in VFI, where fine-grained motion modeling is essential. To overcome this problem, deformable kernel-based methods~\cite{zhu2019deformable, cheng2020video, gui2020featureflow, xiang2020zooming, cheng2021multiple, chen2021pdwn, shi2021video, ding2022MSEConv, shi2022video, zhang2023L2BEC2} introduce learnable offsets~\cite{dai2017dcn} as shown Fig.~\ref{fig:convolutions}~(b), which allow sampling beyond the static grid:
\begin{align}
    \hat{I}(x, y)  =\ & \sum_{i=0}^{N-1}\sum_{k=0}^{R-1} \sum_{l=0}^{R-1} W_{k,l}\notag\\
    &\cdot I_i(x + k + \alpha_{k,l},\ y + l + \beta_{k,l}),
\label{eq:deformable-kernel}
\end{align}
where $(\alpha_{k,l}, \beta_{k,l})$ are the learnable offsets. From a motion-representation viewpoint, these offsets can be interpreted as kernel-centered displacement vectors, while the associated kernel weights modulate how much each sampled location contributes, yielding an implicit analogue of a local flow field. 
AdaCoF~\cite{lee2020adacof} jointly predicts both kernel weights and sampling offsets for each output pixel, improving flexibility over static kernels. However, its limited offset range and time-invariant sampling pattern limits its expressiveness under complex motion. To further enhance spatial adaptivity, dynamic kernel-based methods~\cite{park2021asymmetric, park2020bmbc, ding2021cdfi, ding2022MSEConv, liu2023ttvfi} as shown in Fig.~\ref{fig:convolutions}~(c) generate location-dependent kernel weights:
\begin{align}
    \hat{I}(x, y) =\ & \sum_{i=0}^{N-1}\sum_{k=0}^{R-1} \sum_{l=0}^{R-1}
    W_{k,l}(x, y)\notag\\
    &\cdot I_i(x + k + \alpha_{k,l},\ y + l + \beta_{k,l}),
\label{eq:dynamic-kernel}
\end{align}
where $W_{k,l}(x,y)$ denotes a dynamically predicted kernel at location $(x, y)$. Methods such as CDFI~\cite{ding2021cdfi} and MSEConv~\cite{ding2022MSEConv} jointly learn spatially-varying weights and offsets, resulting in enhanced flexibility and improved interpolation accuracy.

Kernel-based models adopt a simple single-stage formulation that combines motion estimation and frame synthesis into a one-step process~\cite{niklaus2017adaconv, niklaus2017sepconv}. Instead of first regressing an explicit flow field and then warping the inputs, the network directly predicts sampling kernels whose spatial support implicitly specifies where information is gathered from the input frames, thereby coupling motion inference and reconstruction in a unified operation. This implicit formulation enhances robustness in motion-ambiguous or low-texture regions by avoiding reliance on external optical flow estimators. However, a notable limitation arises from their temporal rigidity. Most methods are trained to interpolate at fixed time steps (\textit{e.g.}, $t=0.5$) and lack generalization to arbitrary times $t \in (0,1)$. As a result, they are typically restricted to CTFI (Sec.~\ref{subsec:CTFI}) and fail to support ATFI (Sec.~\ref{subsec:ATFI}), limiting their applicability in real-world scenarios requiring temporal flexibility.

\subsubsection{\textbf{Flow-based}}
\label{subsubsec:flow-based}

Flow-based methods~\cite{park2020robust, liu2017DVF, liu2019cyclicgen, yuan2019zoom, reda2019unsupervised, xu2019quadratic, chi2020all, niklaus2020softmax, liu2020enhanced, zhang2020flexible, sim2021xvfi, shi2021video, hu2022many, kong2022ifrnet, park2021asymmetric, huang2022rife, shangguan2022learning, reda2022film, niklaus2023splatting, jin2023enhanced, park2023biformer, jin2023unified, li2023amt, zhang2023extracting, jeong2024ocai, guo2024generalizable, wu2024perception, zhong2024clearer, niklaus2018context, bao2019memc, park2020bmbc, gui2020featureflow, niklaus2021revisiting, shen2024ladder, briedis2025controllable} explicitly estimate dense motion in the form of \textit{optical flow}, a dense motion field representing the pixel-wise displacements between two frames, and use it to align inputs temporally for intermediate frame synthesis. Advances in optical flow estimation~\cite{dosovitskiy2015flownet, ilg2017flownet, sun2018pwc, teed2020raft} have directly propelled the performance of flow-based VFI models. A typical pipeline consists of three stages: (1) estimating either \textit{anchor flows} ($\mathcal{V}_{0 \rightarrow t}, \mathcal{V}_{1 \rightarrow t}$) or \textit{intermediate flows} ($\mathcal{V}_{t \rightarrow 0}, \mathcal{V}_{t \rightarrow 1}$), (2) warping~\cite{niklaus2020softmax, jaderberg2015spatial} the inputs or their features ($I_0, I_1$ or $F_0, F_1$) using the predicted flow fields, and (3) synthesizing the target frame ($\hat{I}_t$) by blending the warped results ($\hat{I}_{0 \rightarrow t} / \hat{F}_{0 \rightarrow t}$ and $\hat{I}_{1 \rightarrow t} / \hat{F}_{1 \rightarrow t}$).

The accuracy of the flow critically impacts interpolation quality in this approach, as misalignment directly causes blur and artifacts. Early works~\cite{niklaus2018context, bao2019depth, niklaus2020softmax} adopt off-the-shelf optical flow networks~\cite{jiang2021learning, weinzaepfel2013deepflow, dosovitskiy2015flownet, ilg2017flownet, ranjan2017optical, sun2018pwc, hui2018liteflownet, bar2020scopeflow, teed2020raft, huang2022flowformer, xu2022gmflow} to estimate the initial flows. While these networks offer strong general-purpose motion estimation, they are not specifically optimized for the VFI task and struggle to handle motions that lie outside the training distribution~\cite{jin2023enhanced}. Moreover, their large parameter count introduces unnecessary overhead. To address these issues, a number of methods~\cite{liu2017DVF, jiang2018super, xue2019TOFlow, yuan2019zoom, chi2020all, park2020bmbc, huang2022rife, kong2022ifrnet, danier2022stmfnet, reda2022film, park2023biformer, li2023amt, zhang2023extracting, jin2023enhanced, liu2024sparse, zhang2024vfimamba} estimate their own task-oriented flow within their framework, which is optimized jointly with the frame interpolation objective. BiM-VFI~\cite{seo2024bim} distills flow knowledge from an ensemble of flow networks into a compact flow estimator aligned with the interpolation task. GIMM-VFI~\cite{guo2024generalizable} mitigates the noise in flows from pre-trained flow estimator (\textit{e.g.}, RAFT~\cite{teed2020raft}, FlowFormer~\cite{huang2022flowformer}) by refining them through a coordinate-based implicit networks. Pseudo ground-truth (GT) strategies are also common, where pseudo GT flow is generated by existing flow networks and used as weak supervision to bootstrap VFI training~\cite{liu2017DVF, sim2021xvfi}. These help produce temporally consistent and semantically aligned flows customized for interpolation. With the estimated flow, warping is implemented via either forward~\cite{niklaus2020softmax} or backward~\cite{jaderberg2015spatial} warping. Depending on the warping operation, it requires different types of flows.

\begin{figure}
    \centering
    \begin{overpic}[width=0.85\linewidth]{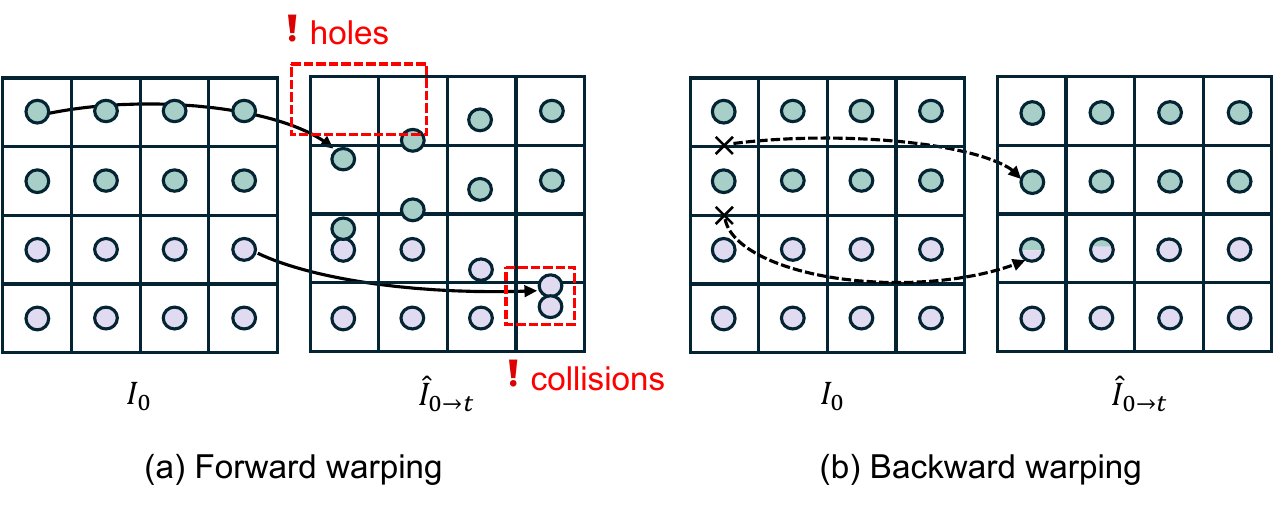}
    \put(36,3){%
    {\sffamily\tiny\textcolor{black}{(Eq~(\ref{eq:FW}))}}}
    \put(91,3){%
    {\sffamily\tiny\textcolor{black}{(Eq~(\ref{eq:BW}))}}}
    \end{overpic}
    \vspace{-0.4cm}
    \caption{Comparison of forward and backward warping strategies.
    (a) Forward warping~\cite{niklaus2020softmax} projects source pixels ($I_0$) to their estimated positions in the target frame using $\mathcal{V}_{0 \rightarrow t}$. 
    (b) Backward warping~\cite{jaderberg2015spatial} samples each pixel in the target frame from the source using $\mathcal{V}_{t \rightarrow 0}$.}
\label{fig:FW_BW}
\vspace{-0.5cm}
\end{figure}

\textbf{Forward warping.} Forward warping-based methods~\cite{niklaus2018context, xue2019TOFlow, liu2019cyclicgen, yuan2019zoom, chi2020all, niklaus2020softmax, park2020bmbc, niklaus2021revisiting, hu2022many, jin2023enhanced, jin2023unified} first estimate \textit{bidirectional flows} $(\mathcal{V}_{0 \rightarrow 1}, \mathcal{V}_{1 \rightarrow 0})$, from which the intermediate flows $(\mathcal{V}_{0 \rightarrow t}, \mathcal{V}_{1 \rightarrow t})$ are linearly interpolated:
\begin{equation}
    \hat{\mathcal{V}}_{0 \rightarrow t} = t \cdot \mathcal{V}_{0 \rightarrow 1}, \quad
    \hat{\mathcal{V}}_{1 \rightarrow t} = (1 - t) \cdot \mathcal{V}_{1 \rightarrow 0}.
    \label{eq:flow_interp}
\end{equation}
These flows are then used to project source pixels to the target frame:
\begin{equation}
    \hat{I}_{0 \rightarrow t} = \overset{\rightarrow}{\mathcal{W}_f}(I_0, \hat{\mathcal{V}}_{0 \rightarrow t}), \quad
    \hat{I}_{1 \rightarrow t} = \overset{\rightarrow}{\mathcal{W}_f}(I_1, \hat{\mathcal{V}}_{1 \rightarrow t}).
    \label{eq:FW}
\end{equation}
While conceptually simple, forward warping suffers from two inherent issues: \textit{holes} (unmapped regions) and \textit{collisions} (multiple pixels mapped to the same location), especially near motion boundaries~\cite{jiang2018super}, as shown in Fig.~\ref{fig:FW_BW}~(a). These artifacts arise from the intrinsic asymmetry of optical flow, where inverse consistency does not generally hold (\textit{i.e.}, $\mathcal{V}_{i \rightarrow j} \neq -\mathcal{V}_{j \rightarrow i}$), particularly under occlusions or non-rigid motion. Moreover, forward warping is generally non-differentiable due to discrete sampling, hindering gradient-based optimization. To address this, SoftSplat~\cite{niklaus2020softmax} proposes a differentiable softmax-based splatting mechanism:
\begin{equation}
     \overset{\rightarrow}{\mathcal{W}_f}(I_0, \mathcal{V}_{0 \rightarrow t}) = 
    \frac{\vec{\sum} (\exp(Z) \cdot I_0, \mathcal{V}_{0 \rightarrow t})}
         {\vec{\sum} (\exp(Z), \mathcal{V}_{0 \rightarrow t})},
\label{eq:softsplat}
\end{equation}
where $Z$ denotes a learned importance map (\textit{e.g.}, depth), and $\vec{\sum}$ denotes a differentiable splatting with soft aggregation. This formulation not only mitigates the aforementioned artifacts but also improves the gradient flow by making warping fully differentiable, in contrast to standard splatting operations which are piecewise constant and non-smooth. Despite this, the inherent artifacts make naive forward warping a less favored primary choice.

\textbf{Backward warping.} Backward warping~\cite{jiang2018super, bao2019memc, xu2019quadratic, kong2022ifrnet, huang2022rife, park2023biformer} reconstructs each target pixel by sampling from the input frames using estimated intermediate flows $(\mathcal{V}_{t \rightarrow 0}, \mathcal{V}_{t \rightarrow 1})$, which represent motion from the unknown target frame to each input frame. Since the target frame is unavailable, it is not straightforward to obtain these flows. It can be approximated via direct prediction~\cite{liu2017DVF, xue2019TOFlow, park2020bmbc, park2021asymmetric, huang2022rife, lu2022video, kong2022ifrnet, li2023amt, zhang2023extracting, park2023biformer, liu2024sparse, zhang2024vfimamba, zhong2024clearer}, flow interpolation~\cite{jiang2018super, bao2019depth}, or flow reversal techniques~\cite{xu2019quadratic, liu2020enhanced, sim2021xvfi, wu2024perception}. For instance, SuperSloMo~\cite{jiang2018super} employs linear approximations and further refines them via dedicated subnetworks:
\begin{equation}
\begin{aligned}
    \hat{\mathcal{V}}_{t \rightarrow 0} &= - t \cdot {\mathcal{V}}_{0 \rightarrow 1} \,\, \textrm{ or } \,\, t \cdot {\mathcal{V}}_{1 \rightarrow 0}
    \label{eq:m_approx_0}
\end{aligned}
\end{equation}
\begin{equation}
\begin{aligned}
    \hat{\mathcal{V}}_{t \rightarrow 1} &= (1-t) \cdot {\mathcal{V}}_{0 \rightarrow 1} \,\, \textrm{ or } \,\, -(1-t) \cdot {\mathcal{V}}_{1 \rightarrow 0}.
    \label{eq:m_approx_1}
\end{aligned}
\end{equation}

To enhance robustness against ambiguities near motion boundaries, XVFI~\cite{sim2021xvfi} introduces Complementary Flow Reversal (CFR), which aggregates multiple reversed flows to construct more stable motion fields. Given the intermediate flows, backward warping is applied as:
\begin{equation}
\hat{I}_{0 \rightarrow t} = \overset{\leftarrow}{\mathcal{W}_b}(I_0, \hat{\mathcal{V}}_{t \rightarrow 0}), \quad
\hat{I}_{1 \rightarrow t} = \overset{\leftarrow}{\mathcal{W}_b}(I_1, \hat{\mathcal{V}}_{t \rightarrow 1}),
\label{eq:BW}
\end{equation}
where $\overset{\leftarrow}{\mathcal{W}_b}$ denotes the backward warping operator~\cite{jaderberg2015spatial}. The warped results are blended using occlusion-aware mask $M$ and residual refinement term $R$:
\begin{equation}
    I_t = M \odot \hat{I}_{0 \rightarrow t} + (1 - M) \odot \hat{I}_{1 \rightarrow t} + R.
\end{equation}
The operator $\odot$ denotes element-wise multiplication, or the Hadamard product, which blends the warped frames proportionally based on the occlusion-aware confidence map. Some methods~\cite{jiang2018super, xu2019quadratic, reda2019unsupervised, sim2021xvfi} further incorporates $(1{-}t)$ and $t$ as scalar weights into $M$ to guide time-aware blending.  
Several methods also exploit auxiliary priors such as depth~\cite{bao2019depth}, contextual features~\cite{niklaus2018context, bao2019memc, bao2019depth, park2020bmbc, niklaus2020softmax, shi2021video}, or edge information~\cite{liu2019cyclicgen, liu2020enhanced, gui2020featureflow} to further guide interpolation. Learnable synthesis networks~\cite{fourure2017residual} are also commonly employed to further sharpen the output and correct residual artifacts.

\textbf{Modeling non-linear motion.} Many early methods~\cite{liu2017DVF, jiang2018super, niklaus2018context, bao2019memc, bao2019depth, liu2019cyclicgen, yuan2019zoom, xue2019TOFlow, park2020bmbc, niklaus2020softmax, jin2023unified} assume linear motion and brightness constancy, meaning that objects move along a straight trajectories at constant speed, and pixel intensities remain unchanged. However, these assumptions often fail under real-world scenarios involving acceleration, occlusion, or dynamic lighting. Quadratic~\cite{xu2019quadratic, liu2020enhanced, zhang2020video, hu2024iqvfi} or cubic~\cite{chi2020all} motion modeling has been proposed to account for acceleration. QVI~\cite{xu2019quadratic} and EQVI~\cite{liu2020enhanced} estimate acceleration-aware flows utilizing four input frames. While recent works~\cite{seo2024bim, zhong2024clearer} further explore \textit{velocity ambiguity}~\cite{zhong2024clearer}, which refers to the ill-posed nature of intermediate motion inference where multiple trajectories yield the same intermediate position, especially under occlusion or acceleration. BiM-VFI~\cite{seo2024bim} and Zhong \textit{et al}~\cite{zhong2024clearer} introduce bidirectional motion fields and time-aware reasoning mechanisms to disambiguate such cases, enabling robust interpolation under occlusion, acceleration, and non-linear motion. 

Overall, flow-based methods remain one of the most extensively explored and practically adopted approaches in VFI, owing to their explicit and interpretable modeling of motion trajectories. Their ability to flexibly generate intermediate frames for arbitrary timestamps makes them well-suited for applications such as variable frame-rate generation and slow-motion rendering. Despite these strengths, their performance is sensitive to flow estimation accuracy, particularly under conditions of occlusion, large motion, lighting variation or non-linear motion. As research in optical flow continues to evolve~\cite{du2025mambaflow, dong2024memflow, bargatin2025memfof}, flow-based VFI is expected to further benefit from these developments and remain a foundational component of future VFI approach. In contrast to the kernel-based approach in Sec.~\ref{subsubsec:kernel-based}, which encodes motion implicitly in spatially adaptive kernels and typically focuses on single-step CTFI, flow-based methods maintain an \textit{explicit} dense flow field that can be reused to synthesize multiple timestamps and to inspect failure cases. This explicit representation provides strong interpretability and a large effective motion range, whereas kernel-based models, by tying their sampling patterns to local content, are often more robust in low-texture or motion-ambiguous regions but are less straightforward to extend to arbitrary timestamps or to diagnose at the level of explicit motion fields.

\subsubsection{\textbf{Kernel- and Flow-based Combined}}
\label{subsubsec:kernel-and-flow-combined}

Kernel- and flow-based approaches each offer distinct strengths in VFI. Flow-based methods explicitly model pixel-wise motion to enable temporally consistent frame alignment, but are sensitive to inaccuracies in optical flow estimation. In contrast, kernel-based methods directly synthesize pixels using learned, spatially adaptive convolutional kernels, where motion cues are encoded implicitly in the sampling support and kernel weights rather than in an explicit dense flow field, offering greater robustness in regions with complex motion. However, they are limited by their local receptive field and thus struggle with large displacements.

From a motion-representation standpoint, hybrid designs make explicit that kernel- and flow-based strategies are complementary rather than mutually exclusive. 
Flow fields provide an \emph{explicit} description of pixel-wise displacement that is easy to visualize, debug, and reuse for arbitrary timestamps, 
whereas kernel-based sampling offers an \emph{implicit} representation in which motion is captured by content-adaptive sampling locations and aggregation weights, often yielding more stable behavior in low-texture or motion-ambiguous regions. 
Hybrid architectures exploit this complementarity by using flow to provide globally coherent displacement guidance, while kernels refine local appearance and compensate for residual misalignment.

Hybrid methods combine these advantages by leveraging optical flow to guide the placement and orientation of learned convolutional kernels, achieving global motion alignment while enabling local refinement. This combined approach~\cite{niklaus2018context, bao2019depth, bao2019memc, yuan2019zoom, park2020bmbc, lee2020adacof, niklaus2021revisiting, danier2022stmfnet} typically begins with estimating optical flows using dedicated or pre-trained flow networks~\cite{jiang2021learning, weinzaepfel2013deepflow, dosovitskiy2015flownet, ilg2017flownet, ranjan2017optical, sun2018pwc, hui2018liteflownet, bar2020scopeflow, teed2020raft, huang2022flowformer, xu2022gmflow}, which provide the sampling offsets for adaptive kernels. These kernels are then applied along flow-guided paths to aggregate motion-aware pixel neighborhoods. MEMC-Net~\cite{bao2019memc}, for example, combines PWC-Net~\cite{sun2018pwc} for flow estimation and deformable convolution~\cite{dai2017dcn} for localized refinement. In this setup, flow fields define the sampling offsets, while the kernel weights are learned to capture residual motion and restore high-frequency content.

Despite their accuracy, hybrid approach typically introduces significant computational costs due to the dual pipelines for flow and kernel prediction~\cite{ding2021cdfi}. To alleviate this, several works~\cite{danier2022stmfnet, shen2024ladder} adopt encoder-sharing strategies to reduce redundancy and latency. These designs enhance interpolation robustness in scenarios with large displacements, motion ambiguities, or complex occlusion, where single approach-based models often fail. As hybrid architectures continue to evolve, balancing the performance and efficiency remains a central challenge and a promising direction.

\subsubsection{\textbf{Phase-based}}
\label{subsubsec:phase-based}

An alternative direction in VFI exploits the phase information to capture motion cues. In the frequency domain, pixel-wise representations can be decomposed into amplitude and phase components, where temporal phase shifts across frames encode the apparent motion of underlying structures. To extract and manipulate phase information, most phase-based methods~\cite{meyer2015phase, meyer2018phasenet} adopt multi-scale frequency representations such as complex steerable pyramids~\cite{simon1992shiftable, simon1995steerable, portilla2000parametric}. Motion is then modeled by interpolating both phase and amplitude at each pyramid level. Meyer \textit{et al}.~\cite{meyer2015phase} solves this optimization problem explicitly, while PhaseNet~\cite{meyer2018phasenet} adopts end-to-end learning strategies. These methods offer robustness to lighting changes and subpixel motion, without relying on explicit pixel correspondence.

However, the underlying assumption that motion can be approximated as local phase shift, fails under large displacement. It leads to phase ambiguity and aliasing artifacts~\cite{wadhwa2013phasebase, didyk2013jointview}. As a result, phase-based methods often struggle to handle high-speed motion and often produce blurry results around sharp edges or occlusion boundaries. Nonetheless, phase representations remain a valuable signal modality and, when combined with other learning-based methods, may help enhance robustness against photometric and structural distortions.

\subsubsection{\textbf{GAN-based}}
\label{subsubsec:gan-based}

Conventional learning-based VFI approaches predominantly rely on pixel-wise losses such as $\ell_1$, $\ell_2$, or deep feature-based perceptual losses (e.g., VGG~\cite{liu2015vgg}). Although these objectives effectively reduce reconstruction errors, they often lead to over-smoothed textures and lack of fine details, thereby compromising perceptual realism~\cite{men2020VQA, danier2022VQA}. To overcome this limitation, several methods adopt Generative Adversarial Networks (GANs)~\cite{goodfellow2020generative}, which demonstrate remarkable performance in synthesizing visually plausible content~\cite{gulrajani2017improved, berthelot2017began}. GAN-based VFI methods~\cite{van2017FIGAN, wen2018generating, yuan2019zoom, lee2020adacof, danier2022stmfnet} employ a generator $G$ to synthesize the intermediate frame $\hat{I}_t$, while a discriminator $D$ distinguishes between the GT $I_t$ and $\hat{I}_t$. The generator is trained with both reconstruction and adversarial losses, enabling it to preserve structural consistency with the input frames while enhancing visual fidelity. Such formulations are particularly effective in hallucinating plausible textures in disoccluded or low-texture regions~\cite{larsen2016autoencoding, yuan2019zoom}.

Despite their potential, this approach introduces new challenges, including training instability, mode collapse~\cite{arjovsky2017wasserstein}, and limited generalization to unseen motion dynamics or scene layouts. These issues can lead to artifacts or unrealistic interpolations, especially when the training data lacks sufficient diversity. Consequently, domain adaptation or fine-tuning is often required when deploying these models in novel settings~\cite{chen2020generative}, raising concerns about scalability and robustness.

\subsubsection{\textbf{Transformer-based}} 
\label{subsubsec:transformer-based}

Originally proposed for sequence modeling in natural language processing (NLP)~\cite{vaswani2017attention}, the Transformer architecture has been successfully adapted to VFI~\cite{choi2020cain, shi2022video, lu2022video, zhang2023extracting, li2023amt, park2023biformer, zhang2023L2BEC2, liu2023ttvfi, lyu2024brownian, liu2024sparse, zhang2025eden} owing to its strong capacity for capturing long-range dependencies through the attention mechanism~\cite{vaswani2017attention, liu2021swin}. In the context of VFI, where motion often spans large spatial and temporal regions with occlusions and deformations, this ability is particularly advantageous. The attention mechanism adaptively weighs features by their relevance to selectively attend to distant yet semantically relevant regions. This is an essential property for synthesizing temporally coherent intermediate frames. The core attention operation is defined as:
\begin{equation}
    \text{Attn}(Q, K, V) = \text{Softmax}\left( \frac{QK^\top}{\sqrt{d}} \right)V,
\end{equation}
where $Q$, $K$, and $V$ denote the query, key, and value matrices respectively and $d$ is the dimension of the feature space. This formulation enables the model to focus on spatial-temporal regions that are informative for interpolation, while effectively handling occlusions and appearance changes~\cite{gui2020featureflow}.

Transformer-based VFI methods primarily differ in how they structure attention and encode temporal dependencies. VFIFormer~\cite{lu2022video} introduces a cross-scale window-based attention (CSWA) mechanism to capture multi-scale dependencies without relying on flow-based motion estimation. Queries are computed from features at the target time, while keys and values are derived from neighboring input frames, enabling direct temporal associations. The multi-scale windowing expands the receptive field, enhancing robustness to complex motion. EMA-VFI~\cite{zhang2023extracting} integrates attention modules with CNNs to reduce overhead, using inter-frame attention to jointly extract motion and appearance cues with improved efficiency.

Despite their advantages, this approach faces high computational burden because standard self-attention scales quadratically with input size. To overcome this, efficient attention designs have been proposed. Swin Transformer~\cite{liu2021swin} reduces complexity via windowed self-attention and shifted windows, while Restormer~\cite{zamir2022restormer} introduces transposed attention to achieve linear complexity with respect to spatial dimensions. These developments point to a promising direction in which Transformer-based architectures may effectively balance global context modeling with computational efficiency, enabling real-time HR frame interpolation in practical applications.

\subsubsection{\textbf{Mamba-based}}
\label{subsubsec:mamba-based}

Structured State Space Models (SSMs)~\cite{gu2021efficiently} offer a principled approach to sequence modeling by formulating input–output dynamics through linear dynamical systems. Mamba~\cite{gu2023mamba} introduces selective state-space parameterization with input-dependent gating and linear recurrence, enabling efficient long-range dependency modeling with linear time complexity. This architectural simplicity and scalability make Mamba a compelling alternative to conventional attention-based models, particularly in scenarios where both temporal context length and computational budget are critical.

VFIMamba~\cite{zhang2024vfimamba} is the first to adopt Mamba as a core temporal modeling backbone. Its hierarchical SSM-based design allows bidirectional recurrence across multiple spatial scales, facilitating robust motion feature propagation and long-range temporal alignment while maintaining low memory usage. In this sense, VFIMamba is representative of a new line of lightweight VFI architectures that replace global attention with structured recurrence, achieving competitive interpolation accuracy under substantially reduced computational cost. LC-Mamba~\cite{jeong2025lc} further refines this idea by incorporating shifted-window mechanisms and Hilbert-curve-based spatial scanning to preserve locality and continuity, which are crucial for high-resolution frame synthesis. These designs highlight the capacity of structured recurrence to capture both global motion trends and fine-grained dynamics necessary for high-fidelity interpolation. Beyond VFI, Mamba-based models such as MambaIR~\cite{guo2024mambair} and MambaIRv2~\cite{guo2024mambairv2} have shown promising results in image restoration, suggesting that the core modeling principles behind Mamba generalize well across vision domains and tasks.

Taken together, these developments suggest that Mamba is emerging as a promising backbone for spatio-temporal modeling in VFI, particularly when lightweight deployment and long-range temporal context are required. At the same time, its behavior under challenging conditions such as severe occlusion or highly non-rigid motion remains relatively unexplored. Potential directions include localized or deformable recurrence, motion-aware conditioning, and hybrid designs that combine Mamba with complementary modules (e.g., occlusion-aware or non-linear-motion modeling components) to better handle complex video dynamics. We anticipate that future Mamba-based VFI architectures will increasingly exploit such combinations to improve expressiveness while preserving the favorable efficiency properties of structured state-space models.

\subsection{Diffusion Model-based}
\label{subsec:diffusion-model-based}

Diffusion Models (DMs)~\cite{ho2020denoising, dhariwal2021diffusion, rombach2022high} have become a dominant generative framework in image~\cite{dhariwal2021diffusion, rombach2022high}, video~\cite{ho2022video, blattmann2023align}, and multimodal synthesis~\cite{bar2024lumiere, zhang2023i2vgenxl}. Compared to GANs~\cite{goodfellow2020generative} and VAEs~\cite{kingma2013auto}, DMs offer more stable training, higher fidelity outputs, and better temporal coherence. Following their success in text-to-video (T2V)~\cite{wu2023tune, yang2024cogvideox, yuan2024idpreserve} and image-to-video generation (I2V)~\cite{ren2024consisti2v}, recent research has extended DMs to VFI~\cite{wang2024framer, deng2025beyond, zhang2025arbitrary, tanveer2025multicoin, voleti2022mcvd, danier2024ldmvfi, jain2024video, huang2024motion, zhu2024generative, zhang2025eden, wan2024unipaint, hwang2025diffuseslide, yang2024zerosmooth, lyu2025tlb, chen2025sci, lew2025disentangled, guo2025controllable, hong2025semantic}.
\begin{figure}[!t]
    \centering
\includegraphics[width=0.88\linewidth]{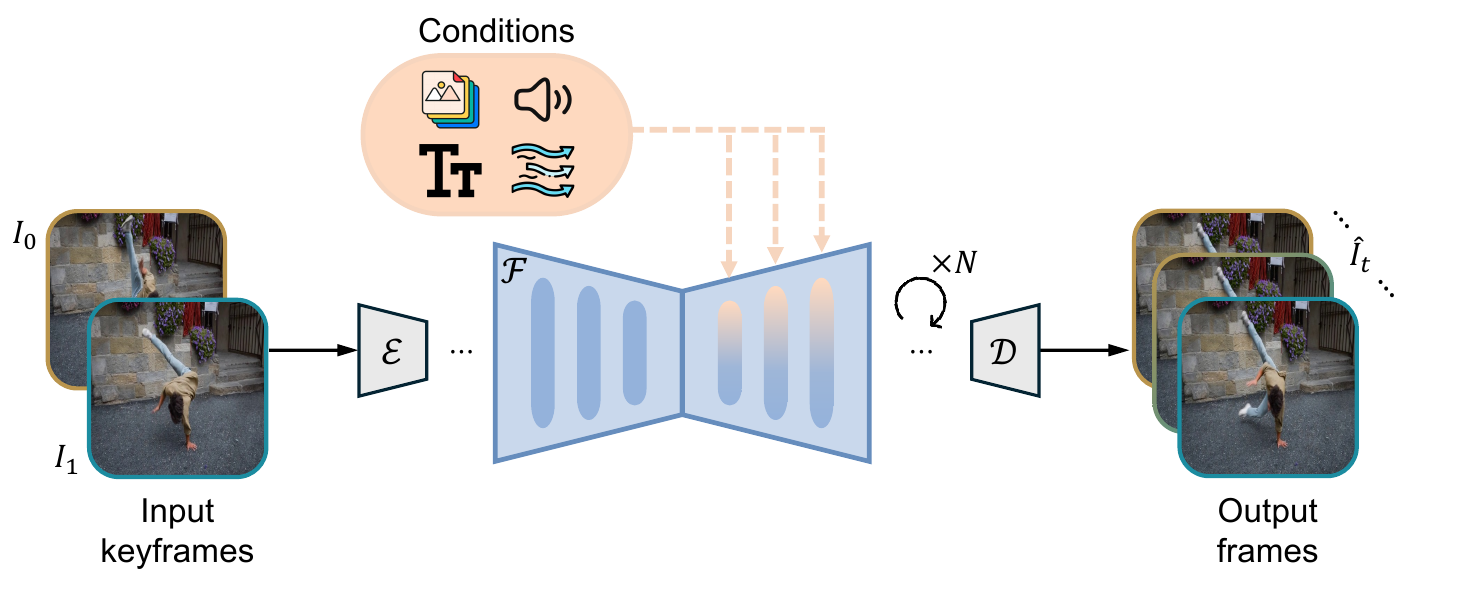} 
    \vspace{-0.3cm}
    \caption{General structure of DM-based VFI. The model receives input keyframes ($I_0$, $I_1$) and generates intermediate frames ($I_t$) through a denoising process. In addition to input keyframes, the model can accept various auxiliary conditioning signals such as images, text, audio, optical flow, or semantic maps via adapter modules or attention mechanisms.}
\label{fig:DM}
\vspace{-0.5cm}
\end{figure}

Among diffusion backbones, Stable Video Diffusion (SVD)~\cite{blattmann2023stable} first encodes video frames into a latent space via an encoder $\mathcal{E}(\cdot)$, adds Gaussian noise, and then denoises the representation using a 3D U-Net~\cite{cciccek20163d}. A typical loss formulation is the $\mathbf{v}$-prediction objective~\cite{salimans2022progressive}:
\begin{equation}
    \mathcal{L} = \mathbb{E}_{\mathbf{z}, \mathbf{c}_{\text{image}}, \epsilon, t} \left[ \left\| \mathbf{v} - f_\theta(\mathbf{z}_t, \mathbf{c}_{\text{image}}, t) \right\|_2^2 \right],
\end{equation}
where $\mathbf{v} = \alpha_t \epsilon - \sigma_t \mathbf{z}_t$, $\mathbf{z}_t$ is the noisy latent at timestep $t$, $\epsilon$ is the GT noise, and $\alpha_t$, $\sigma_t$ are variance schedule parameters that define the weighting between signal and noise. $\mathbf{c}_{\text{image}}$ denotes input frames as condition. This reframes VFI as a conditional generation task in latent space. 

Early DM-based VFI works such as MCVD~\cite{voleti2022mcvd} and LDMVFI~\cite{danier2024ldmvfi} generate intermediate frames directly from noise conditioned on keyframes, without explicitly modeling motion. As shown in Fig.~\ref{fig:DM}, a key strength of DMs is their ability to support flexible conditioning on auxiliary signals such as optical flow, semantic maps, audio, or text through adapters~\cite{zhang2023adding, peng2024controlnext, hu2022lora} or cross-attention mechanisms. For example, MoG~\cite{huang2024motion} and FCVG~\cite{zhu2024generative} employ ControlNet~\cite{zhang2023adding} to inject motion priors into the denoising process, while Framer~\cite{wang2024framer} integrates spatial priors via attention-based guidance.

Recently, VFI has been generalized into a broader task termed \textit{Generative Inbetweening}~\cite{feng2024explorative, wang2024generative, hong2025semantic}. Unlike conventional VFI, which assumes short temporal gaps between similar input frames, this formulation  handles \textit{sparse} and semantically distant input keyframes. However, this new concept introduces greater motion ambiguity, making temporal alignment more challenging. To address this, bidirectional strategies like TRF~\cite{feng2024explorative} and ViBiDSampler~\cite{yang2024vibidsampler} fuse forward and backward sampling trajectories. On the architectural side, EDEN~\cite{zhang2025eden} employs a spatiotemporal encoder to enhance global consistency, while TLB-VFI~\cite{lyu2025tlb} utilizes 3D-wavelet gating and temporal-aware autoencoding for motion fidelity. Another recent development is the emergence of Diffusion Transformers (DiTs)~\cite{peebles2023scalable}-based VFI methods~\cite{hong2025semantic, zhang2025arbitrary}. ArbInterp~\cite{zhang2025arbitrary} further emphasizes temporally flexible sampling, enabling generation at arbitrary timestamps with content-aware temporal control. Together with TRF~\cite{feng2024explorative} and ViBiDSampler~\cite{yang2024vibidsampler}, it exemplifies diffusion-guided temporal modeling in which the sampling trajectory is regularized by motion-aware priors rather than being purely stochastic, leading to improved global coherence over long sequences.

In parallel, cross-modal and multi-modal conditioning has been explored as a means to reduce motion ambiguity. MultiCOIN~\cite{tanveer2025multicoin} performs controllable video inbetweening by injecting textual, visual, and structural controls (e.g., trajectories or layout cues) that jointly steer geometry and appearance, while BBF~\cite{deng2025beyond} leverages audio–visual semantic guidance to improve context-aware interpolation around boundary frames, indicating that audio cues can help disambiguate motion direction and event timing in complex scenes.

Despite their potential, DM-based approach faces several limitations: high computational cost, slow sampling, and limited scalability to long-range or HR settings. To mitigate these issues, lightweight or training-free designs~\cite{yang2024zerosmooth, wan2024unipaint} have been explored, along with frameworks that decouple motion estimation from generative synthesis~\cite{guo2025controllable, lew2025disentangled}, offering improved efficiency and controllability.

Overall, DM-based VFI has progressed from straightforward conditional denoising between nearby frames to generative inbetweening frameworks that incorporate diffusion-guided temporal modeling, multi-modal conditioning, and increasingly efficient architectures. We anticipate that DMs will play an important role in VFI, particularly in scenarios involving sparse keyframes, ambiguous motion, and high-level user controls, while further advances in efficiency and controllability will be crucial for practical deployment.
\begin{figure}[!t]
    \centering
    \includegraphics[width=0.85\linewidth]{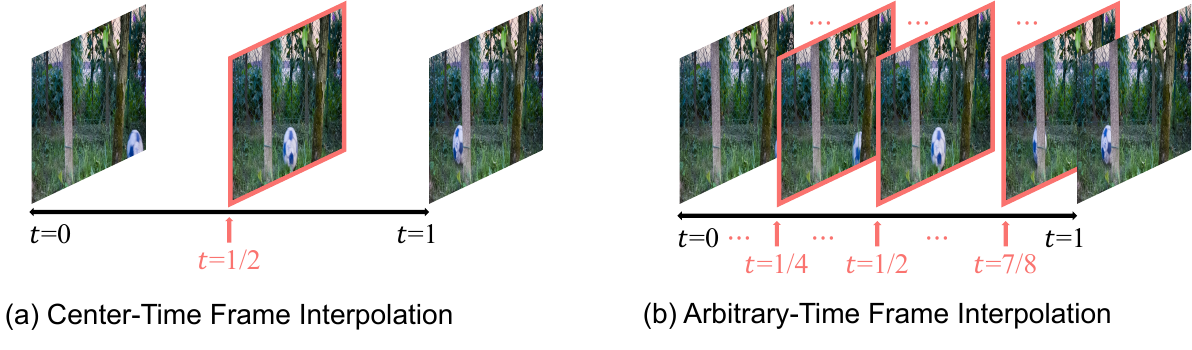} 
    \vspace{-0.3cm}
    \caption{Comparison of CTFI and ATFI. (a) CTFI only generates a single center-frame at $t{=}0.5$ given two inputs. (b) ATFI can synthesize frames at arbitrary $t \in (0,1)$.}
\label{fig:CTFI_ATFI}
\vspace{-0.5cm}
\end{figure}

\section{Learning Paradigm}
\label{sec:learning-paradigm}

\subsection{Center-Time Frame Interpolation (CTFI)}
\label{subsec:CTFI}

Center-Time Frame Interpolation (CTFI) as shown in Fig.~\ref{fig:CTFI_ATFI}~(a), also known as \textit{fixed-time interpolation}, is a widely adopted learning paradigm in VFI. Here, models are trained on triplets $(I_0, I_{\frac{1}{2}}, I_1)$~\cite{soomro2012ucf101, xue2019TOFlow, su2017adobe, choi2020cain, siyao2021deep}, with $I_0$ and $I_1$ as inputs and $I_{\frac{1}{2}}$ as the GT center-frame. Owing to the simplicity of supervision and precise GT alignment, this paradigm has been dominant in early studies~\cite{long2016learning, niklaus2017adaconv, niklaus2017sepconv, liu2017DVF, xu2019quadratic, choi2020cain, lee2020adacof, huang2022rife, shi2022video, lyu2024brownian}.

However, CTFI is inherently limited in generating intermediate frames at arbitrary timestamps. Since models are trained exclusively for the center-frame at $t{=}\frac{1}{2}$, they inherently lack temporal flexibility for generating frames at other timestamps. For example, to generate a frame at $t{=}\frac{1}{4}$, the model first synthesizes $\hat{I}_{\frac{1}{2}}$, and then recursively generates $\hat{I}_{\frac{1}{4}}$ conditioned on $(I_0, \hat{I}_{\frac{1}{2}})$. This sequential process incurs two key drawbacks~\cite{jiang2018super, sim2021xvfi, oh2022demfi}. First, it increases computational latency and prevents parallel generation, as each intermediate frame depends on the previously synthesized result. Second, it leads to cumulative errors where artifacts in earlier frames propagate through the inference chain, degrading temporal consistency and overall quality. Additionally, CTFI restricts the temporal upsampling factor to powers of two ($2^n$), thereby limiting adaptability in diverse frame-rate conversion scenarios such as real-time video streaming or arbitrary slow-motion synthesis. 

\subsection{Arbitrary-Time Frame Interpolation (ATFI)}
\label{subsec:ATFI}

In contrast, Arbitrary-Time Frame Interpolation (ATFI) or \textit{multi-frame interpolation} as shown in Fig.~\ref{fig:CTFI_ATFI}~(b), generalizes the task to arbitrary timestamps $t \in (0,1)$ between two given frames~\cite{jiang2018super, niklaus2018context, bao2019depth, xu2019quadratic, reda2019unsupervised, chi2020all, park2020bmbc, sim2021xvfi, huang2022rife, zhang2023extracting, li2023amt, kalluri2023flavr, zhu2024generative, wang2024framer}. This paradigm explicitly receives $t$ as input, enabling direct and continuous-time interpolation without recursion. Earlier methods~\cite{bao2019depth, liu2019cyclicgen} perform iterative ATFI in a frame-by-frame fashion, often leading to temporal jitter due to a lack of continuity modeling. In contrast, temporally-aware models~\cite{chi2020all, reda2019unsupervised} predict multiple intermediate frames in one pass, promoting temporal coherence and computational efficiency.

While ATFI offers superior flexibility, it introduces new challenges. First, training requires HFR datasets to provide dense supervision at various timestamps. Second, ATFI is inherently susceptible to the velocity ambiguity problem, where multiple motion trajectories can lead to the same intermediate position. This often leads models to average over alternatives, resulting in temporal blur. Third, ATFI must account for non-linear motion such as acceleration or abrupt direction changes, which are difficult to model under constant-velocity assumptions. These challenges are further analyzed in Sec.~\ref{subsec:non-linear}. Despite these issues, ATFI remains a versatile and powerful paradigm for real-world applications, offering improved flexibility for slow-motion generation, dynamic frame-rate adaptation, and user-controllable playback.

\begin{figure}[!t]
    \centering
    \includegraphics[width=0.9\linewidth]{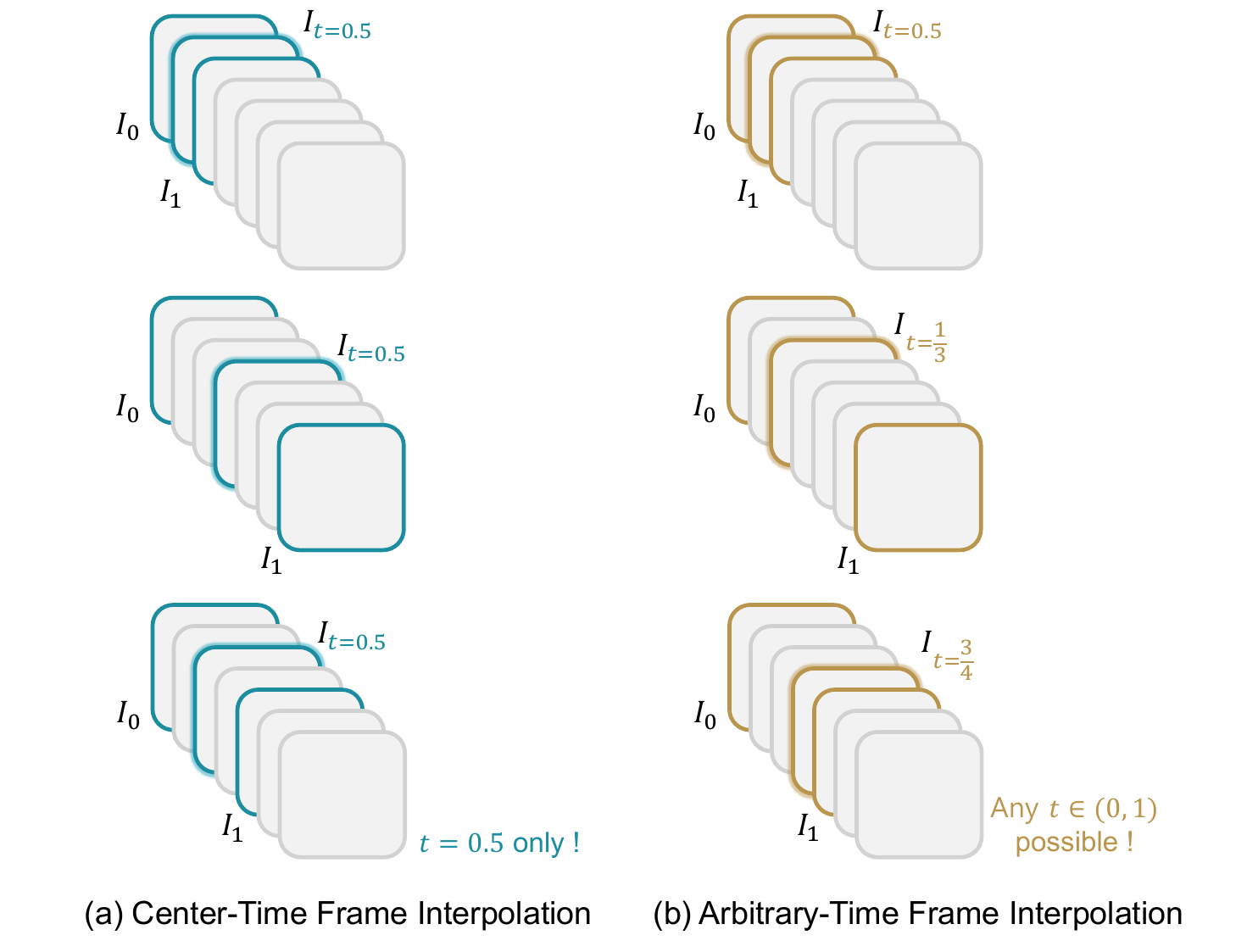} 
    \vspace{-0.3cm}
    \caption{Comparison of CTFI-TS and ATFI-TS.
    (a) CTFI-TS samples exactly three uniformly spaced frames per training example, with only the center-frame used as supervision. 
    (b) ATFI-TS uses $(n{+}1)$ uniformly spaced frames from HFR videos, allowing an intermediate frame at arbitrary timestamp $t \in (0,1)$ to serve as supervision target.}
\label{fig:CTFITS_ATFITS}
\vspace{-0.5cm}
\end{figure}

\vspace{-3pt}
\subsection{Training Strategy}
\label{subsec:training-strategy}

\subsubsection{CTFI Training Strategy (CTFI-TS)}
\label{subsubsec:CTFI-TS}

CTFI-TS builds training triplets $(I_0, I_t, I_1)$ with $I_t$ positioned at the center-point between $I_0$ and $I_1$. These triplets can be generated by uniformly sampling three consecutive frames as shown in Fig.~\ref{fig:CTFITS_ATFITS}~(a). This enables the construction of large-scale training datasets without dense manual annotation. During training, models are supervised exclusively at $t{=}0.5$, and no explicit temporal encoding is involved. At inference, the model predicts only the center-frames at each step. While efficient, this strategy inherently lacks the flexibility to synthesize frames at arbitrary timestamps and requires recursive inference.

\subsubsection{ATFI Training Strategy (ATFI-TS)} 
\label{subsubsec:ATFI-TS}

ATFI-TS constructs training samples from $(n{+}1)$ consecutive frames, using the first and last as inputs $(I_0, I_1)$ and the $(n{-}1)$ intermediate frames as supervision targets for their respective times $t \in (0,1)$. Each $t$ is either provided directly or encoded via temporal embeddings~\cite{kong2022ifrnet, li2023amt, jain2024video}. When HFR videos are available, training data can be flexibly constructed by uniformly sub-sampling frames at a desired interval as shown in Fig.~\ref{fig:CTFITS_ATFITS}~(b). As long as the original frame rate of the video is divisible by the desired interpolation factor, any pair of frames can be selected as inputs, and the frames that lie temporally between them can serve as GT supervision targets. This strategy allows models to learn from a wide distribution of motions and time intervals. Unlike CTFI-TS, inference in ATFI-TS is fully parallelizable, frames at any $t \in (0,1)$ can be generated independently. By explicitly modeling time and enabling continuous supervision, ATFI-TS forms the backbone of modern interpolation frameworks seeking generalizability, temporal coherence, and fine-grained control.

\subsection{Loss Functions}
\label{subsec:loss-functions}

Loss functions play a critical role in guiding VFI models toward producing temporally coherent and perceptually realistic outputs. They are broadly categorized into reconstruction, perceptual, adversarial, and flow-based losses, each addressing different aspects of the interpolation objective.

\subsubsection{Reconstruction Loss}
\label{subsubsec:reconstruction-loss}

Reconstruction losses supervise the model to minimize the pixel-wise discrepancy between the predicted frame $\hat{I}_t$ and the GT frame $I_t^{\text{GT}}$. These losses are typically applied in the RGB space.

\begin{itemize}
    \item\textbf{$\mathcal{L}_1$ Loss} computes the pixel-wise absolute difference between frames, defined as: $\mathcal{L}_1 = \left\lVert \hat{I}_t - I_t^{GT} \right\rVert_1$

    \item\textbf{$\mathcal{L}_2$ loss} computes the squared error, defined as: $\mathcal{L}_2 = \left\lVert \hat{I}_t - I_t^{GT} \right\rVert_2^2$, 
    yielding smoother gradients but often producing overly smoothed outputs, particularly in high-frequency regions or under motion-induced misalignments~\cite{jain2024video}.

    \item\textbf{Charbonnier Loss}~\cite{charbonnier1994two} is a differentiable variant of the $\mathcal{L}1$ loss, defined as: $\mathcal{L}_\text{char} = \rho(I_t^{GT} - \hat{I}_t)$ 
    where $\rho(x) = (x^2 + \epsilon^2)^\alpha$ is the Charbonnier function, with a small constant $\epsilon$ (typically $10^{-3}$) and $\alpha=0.5$. The loss provides smoother gradients than the $\mathcal{L}_1$ loss. Owing to its smooth gradients and robustness to outliers, it is widely used in VFI for stable optimization and accurate reconstruction.

    \item\textbf{Laplacian loss}~\cite{bojanowski2017optimizing} compares the Laplacian pyramid decompositions of the interpolated and GT frames to supervise frame synthesis across multiple spatial scales: $\mathcal{L}_{\text{lap}} = \sum_{i=1}^{l} 2^{i-1} \left\lVert L^{i}(\hat{I}_t) - L^{i}((I_{t}^{GT}) \right\rVert_1$
    where $L^i(\cdot)$ denotes the $i$-th level of the Laplacian pyramid. This encourages alignment of both global structure and fine detail, and is often used in conjunction with $\mathcal{L}_1$ loss.

    \item\textbf{Census loss}~\cite{meister2018unflow}, also referred to as ternary loss, evaluates the structural consistency of local image patches under census transformation~\cite{zabih1994non}. It is defined as: $L_{\text{cen}} = \psi(I_{t}^{GT}, \hat{I}_t)$
    where $\psi(\cdot,\cdot)$  is a Hamming-like distance function over census-encoded patches. Due to its robustness against illumination and photometric noise, it is particularly effective in unsupervised or self-supervised VFI frameworks.
\end{itemize}

\subsubsection{Perceptual Loss}
\label{subsubsec:perceptual-loss}

To enhance perceptual realism, VFI models often incorporate high-level perceptual losses in addition to pixel-wise criteria. A widely adopted formulation computes feature-level distances using a pre-trained VGG network~\cite{simonyan2014very}: $\mathcal{L}_\text{per} = \left\lVert \phi(\hat{I}_t) - \phi(I_{t}^{GT}) \right\rVert_2^2$
where $\phi$ denotes the feature extractor. This loss promotes structural consistency and encourages synthesis of semantically aligned textures, especially in challenging visual regions.

\subsubsection{Adversarial Loss}
\label{subsubsec:adversarial-loss}

Adversarial loss enhances realism by training a discriminator $D$ to distinguish interpolated frames from GT. The typical GAN objective is: $\mathcal{L}_{\text{GAN}} = \mathbb{E}_{I_t^{\mathrm{GT}}}[\log D(I_t^{\mathrm{GT}})] + 
\mathbb{E}_{\hat{I}_t}[\log(1 - D(\hat{I}_t))]$

\begin{figure}[!t]
    \centering
    \includegraphics[width=0.9\columnwidth]{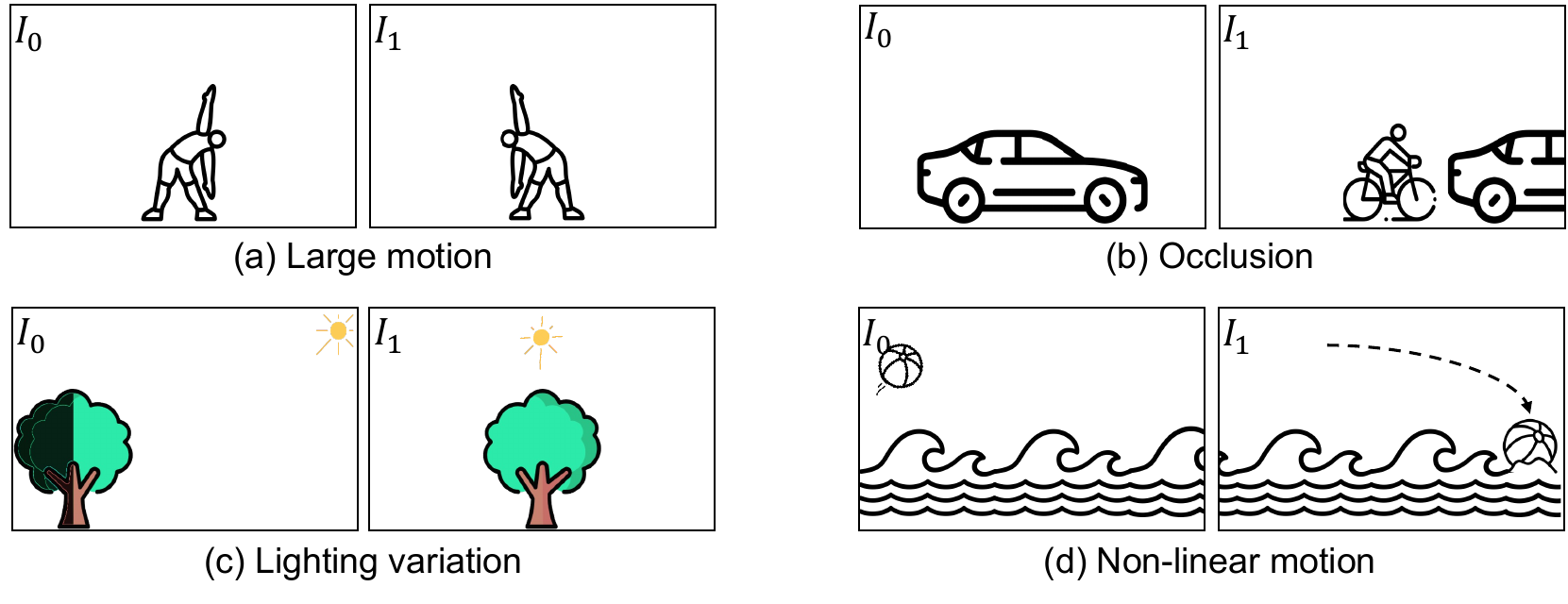} 
    \vspace{-0.3cm}
    \caption{Representative challenges in VFI. 
    (a) A person bends between $I_0$ and $I_1$, creating large articulated motion that makes it difficult to establish accurate correspondences. 
    (b) A cyclist is partially hidden by a passing car, illustrating occlusion where intermediate content is not directly visible in either input frame. 
    (c) Illumination changes from shadowed to fully lit, showing that lighting variation breaks brightness constancy and degrades motion estimation. 
    (d) A beach ball moves along a curved trajectory over wavy water, exemplifying non-linear motion and dynamic textures in which both object paths and stochastic wave patterns cannot be well described by simple linear-flow assumptions.}
\label{fig:challenges}
\vspace{-0.5cm}
\end{figure}

\subsubsection{Flow Loss}
\label{subsubsec:flow-loss}

Given that many VFI models rely on motion estimation as an intermediate step, flow supervision becomes critical for improving temporal alignment. Several loss terms are used to regularize or supervise flow prediction.
\begin{itemize}
    \item\textbf{Smoothness Loss~}\cite{liu2017DVF} encourages piecewise smooth flow by penalizing abrupt spatial changes: $\mathcal{L}_{smooth} = \left\lVert \nabla \mathcal{V}_{0 \to 1} \right\rVert_1 + \left\lVert \nabla \mathcal{V}_{1 \to 0} \right\rVert_1$
    
    \item\textbf{Warping Loss}~\cite{jiang2018super} measures the reconstruction error after warping one frame to the other using estimated flow: $\mathcal{L}_{warp} = \| I_0 - \mathcal{W}(I_1, \mathcal{V}) \|_1 + \| I_1 - \mathcal{W}(I_0, \mathcal{V}) \|_1$
    where $\mathcal{W}$ denotes the warping operator.
    
    \item\textbf{First-order Edge-aware Smoothness Loss}~\cite{sim2021xvfi} is designed to preserve sharp motion discontinuities, this loss attenuates regularization near edges: $\mathcal{L}_{edge} = \sum_{i=0,1} \exp\left( -e^2 \sum_c |\nabla_x I^0_{tc}| \right)^\top \cdot |\nabla_x \mathcal{V}^0_{ti}|$
    where edge strengths are computed via image gradients and used to modulate the smoothness penalty.
\end{itemize}
\section{VFI Challenges}
\label{sec:challenges}

\begin{figure}[!t]
    \centering
    \includegraphics[width=0.9\columnwidth]{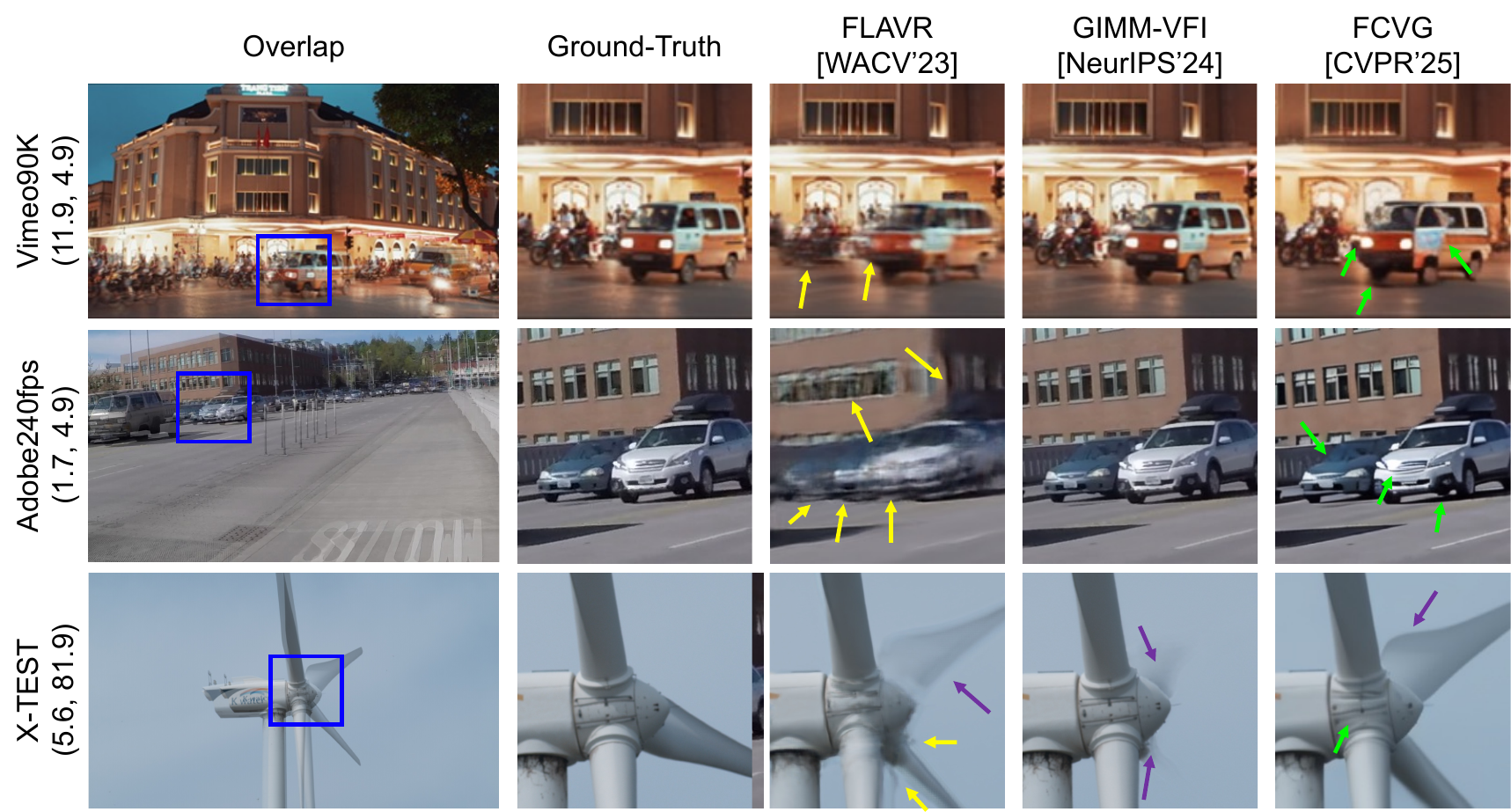} 
    \vspace{-0.3cm}
    \caption{Visual comparison of VFI results under representative challenges. Rows correspond to Vimeo90K (occlusion-dominated), Adobe240fps (moderate motion and blur), and X-TEST (large motion). The first column (“Overlap”) shows the two input frames with a blue crop region; the following columns show the ground truth and the outputs of FLAVR (kernel-based), GIMM-VFI (flow-based), and FCVG (DM-based). 
    Yellow, purple, and green arrows denote motion blur, ghosting artifacts, and severely distorted structures, respectively. (*,*) indicate the $50^{\text{th}}$-percentile occlusion and optical flow magnitudes~\cite{sim2021xvfi}.}
\label{fig:failure_cases}
\vspace{-0.5cm}
\end{figure}

Despite extensive progress in VFI, several representative challenges consistently remain difficult across approaches, limiting real-world performance. As shown in Fig.~\ref{fig:challenges}, these include large motion~\cite{niklaus2020softmax, park2021asymmetric}, occlusion~\cite{jeong2024ocai}, lighting variation, and non-linear motion~\cite{zhong2024clearer, seo2024bim}.

\subsection{Large Motion}
\label{subsec:large-motion}
Large motion refers to scenarios where objects undergo substantial displacement between consecutive frames, as shown in Fig.~\ref{fig:challenges}~(a). This includes articulated movements (\textit{e.g.}, a person leaning left to right), abrupt camera motion, or rapid object translations, all of which cause wide spatial shifts on the image plane. Such motion is common in real-world videos and presents a fundamental challenge in VFI due to the difficulty of establishing accurate correspondences over long spatial ranges. 

To accurately synthesize an intermediate frame, the model must identify where each pixel from the first frame $(I_0)$ has moved in the following frame $(I_1)$ which denotes motion or correspondence estimation. When the motion is small, this is relatively straightforward because corresponding pixels remain close. However, large motion induces long-range dependencies that exceed the receptive field of standard networks. Moreover, appearance changes and occlusions further hinder accurate estimation by introducing discontinuities in motion and visibility. 

To address this, many VFI models adopt a coarse-to-fine hierarchical framework, estimating large displacements at low-resolution (LR) feature maps and progressively refining them at higher resolutions. RIFE~\cite{huang2022rife} employs multi-scale residual flow refinement for robust alignment over wide motion ranges, while FILM~\cite{reda2022film} leverages a feature pyramid to improve flow estimation in fast motion and blur scenarios. IFRNet~\cite{kong2022ifrnet} enhances motion encoding with a motion-aware feature extractor and an intermediate flow refinement block. In addition to these designs, bidirectional motion modeling~\cite{jiang2018super, park2021asymmetric, zhang2023L2BEC2, liu2024sparse} and attention mechanisms~\cite{danier2022stmfnet, park2023biformer, wang2024framer} further improve alignment under extreme motion. ABME~\cite{park2021asymmetric} proposes asymmetric bilateral estimation, predicting forward and backward flows independently to improve robustness under occlusion. BiFormer~\cite{park2023biformer} incorporates deformable attention across bidirectional contexts, enabling the model to dynamically attend to semantically relevant but spatially distant regions, an effective strategy for capturing non-local motion patterns.

These models all share a common objective of expanding the receptive field effectively while maintaining spatial precision. Combining multi-scale refinement, global attention-based matching, and motion-aware modules has proven especially effective in handling large motion. Their advantages are evident on challenging benchmarks such as X4K1000FPS~\cite{sim2021xvfi}, which offers 4K videos at 1000fps with dense GT for precise evaluation. Following this, several HR datasets~\cite{nan2024openvid, stergiou2024lavib, hong2025semantic} have been proposed to further benchmark performance under high-speed and large-displacement conditions. By providing more realistic and challenging settings, these datasets enable better training and evaluation of VFI models in unconstrained environments. As a result, the availability of such benchmarks has accelerated the development of more robust architectures capable of preserving motion detail and fidelity under large displacements. To concretize these behaviors, Fig.~\ref{fig:failure_cases} (third row) compares a kernel-based model (FLAVR~\cite{kalluri2023flavr}), a flow-based model (GIMM-VFI~\cite{guo2024generalizable}), and a diffusion-based model (FCVG~\cite{wu2024perception}) on an X-TEST~\cite{sim2021xvfi} example with extreme motion. Here the kernel-based FLAVR exhibits pronounced motion trails and over-smoothed structures around the turbine hub and blade (yellow arrows), reflecting its limited effective motion range. 
The flow-based GIMM-VFI substantially reduces blur but still shows noticeable geometric distortions near the blade tip and hub (red arrows), indicating sensitivity to flow errors under very long-range displacements. 
The DM-based FCVG, in contrast, preserves sharp and coherent blade and hub structures (green arrows) while occasionally deviating slightly from the exact GT contour, as it prioritizes perceptual plausibility over strict pixel-wise alignment. 
Together with the middle row from Adobe240fps, where motion and occlusion lie between these two regimes and FLAVR tends to show stronger blur around moving objects whereas GIMM-VFI and FCVG maintain relatively clearer object boundaries, these qualitative comparisons indicate that explicit flow modeling and generative refinement both provide advantages over purely kernel-based alignment when large displacements dominate, while still exhibiting characteristic trade-offs between robustness and geometric fidelity.

\subsection{Occlusion}
\label{subsec:occlusion}
Achieving high-quality (HQ) interpolation demands accurate motion estimation as well as a proper understanding of occlusions. Otherwise, severe artifacts are likely to appear in the predicted frames, particularly near motion boundaries. For two consecutive input frames, certain pixels in the intermediate frame may not correspond to any observable region in either input, creating ambiguity in determining the correct content for these occluded regions~\cite{wang2018occlusionaware}. As shown in Fig.~\ref{fig:challenges}~(b), such occlusions can occur when previously hidden objects become visible or when objects move toward the camera, revealing regions that were not seen in either input. Naively blending warped inputs often results in severe artifacts, most notably ghosting artifacts~\cite{jeong2024ocai}, where an object is not only incorrectly projected from its previous location but also appears as a duplicate at its correct position due to the lack of sufficient visual cues. This is especially problematic in disoccluded regions, areas newly revealed in the intermediate frame but absent in both inputs, such as when an object emerges from behind another or moves directly toward the viewpoint. In these cases, the absence of visual evidence introduces ambiguity, making it unclear what content should be synthesized. To resolve this, modern VFI methods incorporate explicit occlusion reasoning to guide the synthesis process.

A common approach involves estimating soft occlusion masks that weight the pixel contributions from each frame~\cite{jiang2018super, yuan2019zoom, bao2019depth, bao2019memc, xue2019TOFlow}. SuperSloMo~\cite{jiang2018super} jointly predicts bidirectional flow and occlusion masks to exclude unreliable pixels during frame blending. SoftSplat~\cite{niklaus2020softmax} improves upon this by introducing a differentiable softmax visibility map that enables confidence-weighted forward warping. OCAI~\cite{jeong2024ocai} further incorporates forward-backward consistency checks~\cite{meister2018unflow} to identify unreliable flow regions and applies targeted masking and flow inpainting to recover missing structures. In addition to visibility maps, auxiliary cues such as context and depth also improve occlusion handling. CtxSyn~\cite{niklaus2018context} integrates warped context features alongside frames to guide synthesis with spatial awareness. DAIN~\cite{bao2019depth} estimates occlusion areas using depth information and leverages neighboring contextual cues to fill the missing regions.

Overall, occlusion-aware VFI remains a critical challenge, particularly in dynamic scenes with depth discontinuities or disoccluded motion. As such, SOTA models increasingly combine multiple strategies, such as masking, depth priors, feature similarity, or forward-backward consistency~\cite{meister2018unflow, jeong2024ocai} to recover plausible content in ambiguous regions and maintain temporal coherence in the output. To complement the quantitative statistics, Fig.~\ref{fig:failure_cases} (top row) presents a failure-case comparison on Vimeo90K~\cite{xue2019TOFlow}, which contains significant occlusions~\cite{sim2021xvfi}. 
Here the flow-based GIMM-VFI produces noticeable ghosting artifacts around disocclusion boundaries (purple arrows), while the kernel-based FLAVR yields smoother although slightly blurred transitions inside the occluded region (yellow arrows). 
The DM-based FCVG is able to hallucinate a plausible object shape behind the occluder, improving perceptual completeness but deviating mildly from the GT geometry (green arrows). 
Together with the Adobe240fps example, which lies between extreme occlusion and large-motion regimes, these qualitative behaviors highlight that kernel-based sampling can be relatively more stable in heavily occluded or texture-poor regions, DM-based models favor perceptual plausibility even when exact geometry is ambiguous, and flow-based models remain sensitive to flow estimation errors near occlusion boundaries.

\subsection{Lighting Variation}
\label{subsec:lighting-variation}
Lighting variation refers to temporal changes in illumination, shadows, reflections, or exposure across consecutive frames as shown in Fig.~\ref{fig:challenges}~(c). These variations can significantly degrade the quality of interpolation, as they violate the basic assumption of brightness constancy~\cite{horn1981determining, baker2011middlebury}, which is widely adopted in many optical flow and motion estimation methods. This assumption presumes that the intensity of a surface patch remains constant across time as it moves, allowing pixel-wise correspondences to be inferred from photometric similarity. However, in practice, lighting changes can cause the same object to appear drastically different between frames, resulting in erroneous motion estimation and visually inconsistent interpolations.

To mitigate this, alternative representations have been proposed. Phase-based methods~\cite{meyer2015phase, meyer2018phasenet} operate in the frequency domain, where motion is encoded as phase shifts rather than intensity differences. These models leverage phase information that remains stable under lighting fluctuations, yielding temporally coherent interpolations even in the presence of flickering or exposure variation. More recently, Transformer-based architectures have shown robustness to photometric inconsistencies. TTVFI~\cite{liu2023ttvfi} aligns motion features across temporal trajectories using attention, enabling the model to blend semantically aligned tokens rather than relying on raw pixel intensities. This higher-level representation effectively helps suppress errors induced from inconsistent lighting, producing perceptually coherent results.

Although lighting variation has received less attention than large motion or occlusion problem, existing methods suggest that photometric-invariant features, frequency-domain modeling, and attention-based alignment provide viable solutions. Continued exploration of these strategies could further enhance the robustness of VFI models in unconstrained environments.

\newcommand{\lnk}[1]{\href{#1}{\faExternalLink}}
\newcommand{\crbx}[2]{%
\scalebox{0.75}{%
\fcolorbox{#1}{#1}{\makebox[1.6ex][c]{\rule{0pt}{1.6ex}#2}}}}
\newcommand{\smath}[1]{$#1$}

\definecolor{Triplet}{RGB}{28, 140, 160}
\definecolor{Multi}{RGB}{186, 149, 77}

\begin{table}[t]
\centering
\scriptsize
\renewcommand{\arraystretch}{1.0}
\setlength{\tabcolsep}{3pt}

\caption{\textbf{Summary and comparison of popular datasets for VFI.}
The dataset types \crbx{Triplet!70}{T} represent Triplet datasets, and
\crbx{Multi!70}{M} represent Multi-frame datasets.}
\label{tab:dataset}

\resizebox{0.9\columnwidth}{!}{%
\begin{tabular}{c c c c c c c}
\toprule
\textbf{Dataset} &
\textbf{Venue} &
\textbf{Type} &
\textbf{Resolution} &
\textbf{Split} &
\textbf{\#Videos / \#Triplets} &
\textbf{URL} \\
\midrule

\multirow{2}{*}{Middlebury\textsuperscript{\cite{baker2011middlebury}}}
& \multirow{2}{*}{IJCV'11}
& \multirow{2}{*}{\crbx{Triplet!70}{T}}
& \multirow{2}{*}{\smath{\leq} 640 $\times$ 480 (VGA)}
& train & -
& \multirow{2}{*}{\lnk{https://vision.middlebury.edu/stereo/data/}} \\
& & & & \cellcolor{pink!20}test & \cellcolor{pink!20}12 & \\

\hline
\multirow{2}{*}{UCF101\textsuperscript{\cite{soomro2012ucf101}}}
& \multirow{2}{*}{CRCV'12}
& \multirow{2}{*}{\crbx{Triplet!70}{T}}
& \multirow{2}{*}{256 $\times$ 256}
& train & -
& \multirow{2}{*}{\lnk{https://www.crcv.ucf.edu/data/UCF101.php}} \\
& & & & \cellcolor{pink!20}test & \cellcolor{pink!20}379 & \\

\hline
\multirow{2}{*}{Vimeo90K\textsuperscript{\cite{xue2019TOFlow}}}
& \multirow{2}{*}{IJCV'19}
& \multirow{2}{*}{\crbx{Triplet!70}{T}}
& \multirow{2}{*}{448 $\times$ 256}
& train & 51,312
& \multirow{2}{*}{\lnk{http://toflow.csail.mit.edu/}} \\
& & & & \cellcolor{pink!20}test & \cellcolor{pink!20}3,782 & \\

\hline
\multirow{2}{*}{SNU-FILM\textsuperscript{\cite{choi2020cain}}}
& \multirow{2}{*}{AAAI'20}
& \multirow{2}{*}{\crbx{Triplet!70}{T}}
& \multirow{2}{*}{\smath{\leq} 1280 $\times$ 720 (HD)}
& train & -
& \multirow{2}{*}{\lnk{https://github.com/myungsub/CAIN}} \\
& & & & \cellcolor{pink!20}test & \cellcolor{pink!20}1,240 & \\

\hline
\multirow{2}{*}{ATD-12K\textsuperscript{\cite{siyao2021deep}}}
& \multirow{2}{*}{CVPR'21}
& \multirow{2}{*}{\crbx{Triplet!70}{T}}
& \multirow{2}{*}{
\begin{tabular}[c]{@{}l@{}}
1280 $\times$ 720 \\
1920 $\times$ 1080 (FHD)
\end{tabular}}
& train & 10,000
& \multirow{2}{*}{\lnk{https://github.com/lisiyao21/AnimeInterp}} \\
& & & & \cellcolor{pink!20}test & \cellcolor{pink!20}2,000 & \\

\hline
\multirow{2}{*}{Xiph\textsuperscript{\cite{niklaus2020softmax}}}
& \multirow{2}{*}{--}
& \multirow{2}{*}{\crbx{Multi!70}{M}}
& \multirow{2}{*}{
\begin{tabular}[c]{@{}l@{}}
2048 $\times$ 1080 (2K) \\
4096 $\times$ 2160 (4K)
\end{tabular}}
& train & -
& \multirow{2}{*}{\lnk{https://media.xiph.org/}} \\
& & & & \cellcolor{pink!20}test & \cellcolor{pink!20}8 & \\

\hline
\multirow{2}{*}{KITTI\textsuperscript{\cite{geiger2012kitti}}}
& \multirow{2}{*}{CVPR'12}
& \multirow{2}{*}{\crbx{Multi!70}{M}}
& \multirow{2}{*}{1240 $\times$ 376}
& train & 194
& \multirow{2}{*}{\lnk{https://www.cvlibs.net/datasets/kitti/}} \\
& & & & \cellcolor{pink!20}test & \cellcolor{pink!20}195 & \\

\hline
\multirow{2}{*}{DAVIS\textsuperscript{\cite{perazzi2016benchmark}}}
& \multirow{2}{*}{CVPR'16}
& \multirow{2}{*}{\crbx{Multi!70}{M}}
& \multirow{2}{*}{1920 $\times$ 1080}
& train & 30
& \multirow{2}{*}{\lnk{https://davischallenge.org/}} \\
& & & & \cellcolor{pink!20}test & \cellcolor{pink!20}20 & \\

\hline
\multirow{2}{*}{HD\textsuperscript{\cite{bao2019memc}}}
& \multirow{2}{*}{TPAMI'19}
& \multirow{2}{*}{\crbx{Multi!70}{M}}
& varied
& train & -
& \multirow{2}{*}{\lnk{https://github.com/baowenbo/MEMC-Net}} \\
& & & & \cellcolor{pink!20}test & \cellcolor{pink!20}11 & \\

\hline
\multirow{2}{*}{Sintel\textsuperscript{\cite{butler2012naturalistic}}}
& \multirow{2}{*}{ECCV'12}
& \multirow{2}{*}{\crbx{Multi!70}{M}}
& \multirow{2}{*}{1024 $\times$ 436}
& train & 23
& \multirow{2}{*}{\lnk{http://sintel.is.tue.mpg.de/}} \\
& & & & \cellcolor{pink!20}test & \cellcolor{pink!20}12 & \\

\hline
\multirow{2}{*}{Adobe240\textsuperscript{\cite{su2017adobe}}}
& \multirow{2}{*}{CVPR'17}
& \multirow{2}{*}{\crbx{Multi!70}{M}}
& \multirow{2}{*}{1280 $\times$ 720}
& train & 61
& \multirow{2}{*}{\lnk{https://github.com/shuochsu/DeepVideoDeblurring}} \\
& & & & \cellcolor{pink!20}test & \cellcolor{pink!20}10 & \\

\hline
\multirow{2}{*}{GOPRO\textsuperscript{\cite{nah2017gopro}}}
& \multirow{2}{*}{CVPR'17}
& \multirow{2}{*}{\crbx{Multi!70}{M}}
& \multirow{2}{*}{1280 $\times$ 720}
& train & 22
& \multirow{2}{*}{\lnk{https://seungjunnah.github.io/Datasets/gopro}} \\
& & & & \cellcolor{pink!20}test & \cellcolor{pink!20}11 & \\

\hline
\multirow{2}{*}{X4K1000FPS\textsuperscript{\cite{sim2021xvfi}}}
& \multirow{2}{*}{ICCV'21}
& \multirow{2}{*}{\crbx{Multi!70}{M}}
& \multirow{2}{*}{4096 $\times$ 2160}
& train & 4,408
& \multirow{2}{*}{\lnk{https://github.com/JihyongOh/XVFI}} \\
& & & & \cellcolor{pink!20}test & \cellcolor{pink!20}15 & \\

\hline
\multirow{2}{*}{WebVid-10M\textsuperscript{\cite{bain2021frozen}}}
& \multirow{2}{*}{ICCV'21}
& \multirow{2}{*}{\crbx{Multi!70}{M}}
& varied
& train & 10M
& \multirow{2}{*}{\lnk{https://github.com/m-bain/webvid}} \\
& & & & \cellcolor{pink!20}test & \cellcolor{pink!20}- & \\

\hline
\multirow{2}{*}{LAVIB\textsuperscript{\cite{stergiou2024lavib}}}
& \multirow{2}{*}{NeurIPS'24}
& \multirow{2}{*}{\crbx{Multi!70}{M}}
& \multirow{2}{*}{4096 $\times$ 2160}
& train & 188,644
& \multirow{2}{*}{\lnk{https://alexandrosstergiou.github.io/datasets/LAVIB/}} \\
& & & & \cellcolor{pink!20}test & \cellcolor{pink!20}53,494 & \\

\hline
\multirow{2}{*}{OpenVid\textsuperscript{\cite{nan2024openvid}}}
& \multirow{2}{*}{ICLR'25}
& \multirow{2}{*}{\crbx{Multi!70}{M}}
& \multirow{2}{*}{
\begin{tabular}[c]{@{}l@{}}
\smath{\geq} 512 $\times$ 512 \\
1920 $\times$ 1080
\end{tabular}}
& train & 1M
& \multirow{2}{*}{\lnk{https://huggingface.co/datasets/nkp37/OpenVid-1M}} \\
& & & & \cellcolor{pink!20}test & \cellcolor{pink!20}- & \\

\bottomrule
\end{tabular}
}
\vspace{-0.5cm}
\end{table}

\subsection{Non-linear Motion}
\label{subsec:non-linear}
Many early VFI methods~\cite{liu2017DVF, jiang2018super, niklaus2018context, bao2019memc, bao2019depth, liu2019cyclicgen, xu2019quadratic, yuan2019zoom, xue2019TOFlow, park2020bmbc, niklaus2020softmax, jin2023unified} assume linear or uniform motion between input frames. Under this assumption, objects move along straight trajectories at constant velocity, allowing motion estimation based on simple temporal interpolation. Flow-based~\cite{liu2017DVF, jiang2018super, bao2019memc, xue2019TOFlow}, kernel-based~\cite{bao2019memc}, and even phase-based models~\cite{meyer2018phasenet} often rely on this assumption implicitly. However, in real-world scenarios, motion is frequently non-linear due to acceleration, deceleration, directional changes, or complex local dynamics. As shown in Fig.~\ref{fig:challenges}~(d), a beach ball is thrown along a curved trajectory over a wavy sea surface; both the curved object path and the rapidly evolving wave patterns violate the linear-motion prior and lead to significant estimation errors when only two frames are observed. In practice, such non-linear behavior can be broadly observed in three forms: (i) curved or acceleration-driven object trajectories, (ii) fine-grained \emph{dynamic textures} such as water or foliage, and (iii) \emph{velocity ambiguity}, where multiple plausible velocity profiles or motion paths produce the same observed displacement between two frames.

To address the limitations of linear motion assumptions, researchers have proposed higher-order motion modeling that extends beyond simple first-order trajectories. Since most existing methods operate on only two input frames, they are inherently under-constrained and often forced to assume simple motion. To overcome this, several methods~\cite{xu2019quadratic, chi2020all, shang2023joint, liu2020enhanced} incorporate multiple input frames (typically four) to capture richer temporal variations and better approximate non-linear motion. QVI~\cite{xu2019quadratic} introduces a quadratic motion model that fits second-order trajectories over four consecutive input frames. Specifically, it takes $(I_{-1}, I_0, I_1, I_2)$ as inputs and predicts an intermediate frame $I_t$ for arbitrary $t \in (0, 1)$. By modeling both velocity and acceleration from surrounding frames, QVI enables the network to better handle curved or time-varying motion paths. This parametric formulation allows the model to explicitly account for motion curvature. EQVI~\cite{liu2020enhanced} further refines this idea by combining offset-based warping with temporal embeddings, improving precision and robustness. More recently, IQ-VFI~\cite{hu2024iqvfi} introduces an implicit motion representation using a coordinate-based MLP that adapts to arbitrary motion patterns without requiring predefined trajectory assumptions. These works collectively emphasize the importance of modeling non-linear motion directly, especially in multi-frame settings. Nevertheless, explicitly parameterized motion models such as quadratic trajectories still cannot fully capture the complexity and irregularity of real-world motions, particularly under strong occlusions or highly non-rigid dynamics.

Non-linear motion is also prominent in \emph{dynamic textures}, including water, fire, smoke, or foliage~\cite{danier2022stmfnet}. These regions contain high spatial-frequency content with temporally stochastic and spatially irregular motion, where local displacements are multi-directional and only weakly correlated over time. Conventional flow-based~\cite{jiang2018super, liu2017DVF} and kernel-based models~\cite{niklaus2017sepconv} that assume smooth, one-to-one motion fields often produce temporal incoherence and “boiling” artifacts on such content. ST-MFNet~\cite{danier2022stmfnet} mitigates this by combining multi-flow warping with a 3D CNN-based texture enhancement branch, while TAFI~\cite{danier2021texture} conditions on texture classes (static, dynamic continuous / discrete) to specialize interpolation behavior.

Finally, \textit{velocity ambiguity}~\cite{zhong2024clearer} has recently been recognized as another key challenge. When only two frames are available, multiple plausible motion trajectories can explain the observed displacement, making the underlying motion inherently under-constrained. This ambiguity becomes especially pronounced in scenes involving curved motion or directional switches, such as bouncing balls or rotating limbs. To tackle this, Zhong \textit{et al.}~\cite{zhong2024clearer} learns a velocity embedding that jointly reasons about motion direction and temporal consistency, while BiM-VFI~\cite{seo2024bim} introduces Bidirectional Motion Fields (BiM) encoding angular and magnitude differences around the intermediate time to better model curved and asymmetric trajectories.

These recent advances collectively signal a shift from rigid linear motion priors toward flexible, context-aware motion modeling in VFI. By extending temporal supervision and refining motion representations, either through higher-order parametric formulations, implicit coordinate-based embeddings, or bidirectional velocity fields, modern VFI methods achieve significantly improved performance in complex non-linear motion scenarios. At the same time, such regimes often expose discrepancies between distortion-based metrics (e.g., PSNR) and perceptual or temporal-consistency measures, suggesting that fully understanding and benchmarking non-linear motion remains an open and practically important challenge for VFI.
\section{Datasets and Evaluation}
\label{sec:datasets-and-evaluation}

\subsection{Datasets}
\label{subsec:datasets}

To facilitate training and evaluation across varying temporal resolutions and motion complexities, numerous VFI datasets have been developed. Tab.~\ref{tab:dataset} provides a high-level summary of commonly used datasets categorized into triplet and multi-frame types. We describe each dataset in detail below.

\subsubsection{Triplet Datasets}
\label{subsubsec:tripelt-datasets}

Early deep learning-based VFI approaches primarily rely on triplet datasets, where two input frames are used to predict the temporally centered GT frame. This configuration aligns with CTFI settings (Sec.~\ref{subsec:CTFI}). Some datasets are further extended to seven-frame sequences~\cite{xue2019TOFlow} for evaluating frame-rate upsampling.

\begin{itemize}
\item \textbf{Middlebury}~\cite{baker2011middlebury}: Originally designed for optical flow, it contains short video clips with moderate complexity.
\item \textbf{UCF101}~\cite{soomro2012ucf101, liu2017DVF}: A human action dataset from which a small subset of triplets is used for VFI.
\item \textbf{Vimeo90K}~\cite{xue2019TOFlow}: A widely adopted benchmark with diverse scenes and consistent format. It offers clean supervision and balanced motion complexity.
\item \textbf{SNU-FILM}~\cite{choi2020cain}: Constructed from high-speed footage and categorized by motion difficulty, it enables evaluation across varying levels of motion, occlusion, and blur.
\item \textbf{ATD-12K}~\cite{siyao2021deep}: A large-scale animation dataset with rich stylistic diversity.
\end{itemize}

\subsubsection{Multi-frame Datasets}
\label{subsubsec:multiframe-datasets}

Multi-frame datasets enable dense temporal supervision and are commonly used in both CTFI and ATFI (Sec.~\ref{subsec:ATFI}) settings. They support flexible frame sampling and facilitate evaluation under diverse temporal intervals.

\begin{itemize}
\item\textbf{Xiph}~\cite{montgomery1994xiph, niklaus2020softmax}: A curated 4K video suite with fine, predominantly small motions, widely used to assess interpolation fidelity under high-resolution, subtle-motion conditions.

\item\textbf{KITTI}~\cite{geiger2012kitti}: Driving scenes captured from a moving vehicle, characterized by sparse ground truth, large ego-motion, and strong perspective changes, forming a challenging benchmark for motion estimation and interpolation.

\item\textbf{Sintel}~\cite{butler2012naturalistic}: A synthetic dataset rendered from the \textit{Sintel} film, providing photorealistic appearance, complex dynamics, and dense optical flow annotations for controlled evaluation.

\item\textbf{DAVIS}~\cite{perazzi2016benchmark}: Originally proposed for video object segmentation, DAVIS features complex object motion, occlusions, and non-rigid deformations, making it suitable for analyzing interpolation behavior in realistic dynamic scenes.

\item\textbf{Adobe240}~\cite{su2017adobe}: Real-world videos captured at 240~fps, containing motion blur and illumination changes, commonly used to evaluate fine-grained temporal modeling and slow-motion synthesis.

\item\textbf{GOPRO}~\cite{nah2017gopro}: High-frame-rate recordings from handheld cameras with non-linear motion and defocus blur, providing a realistic testbed for VFI under camera shake and depth-varying blur.

\item\textbf{HD}~\cite{bao2019memc}: A high-resolution subset derived from Xiph, including sharper content and more pronounced motion, suitable for realistic HR interpolation evaluation.

\item\textbf{X4K1000FPS}~\cite{sim2021xvfi}: A 4K, 1000~fps dataset designed for ultra–slow motion and long-range interpolation, offering dense temporal sampling for high-fidelity frame synthesis studies.

\item\textbf{WebVid-10M}~\cite{bain2021frozen}: A large-scale web video corpus originally curated for text–video learning; after appropriate filtering, it provides diverse, in-the-wild content for generative and data-driven VFI.

\item\textbf{LAVIB}~\cite{stergiou2024lavib}: A large-scale, diverse-domain benchmark with balanced splits and curated subsets, enabling systematic evaluation of in-distribution and out-of-distribution VFI performance.

\item\textbf{OpenVid}~\cite{nan2024openvid}: A text–video dataset with densely aligned samples that supports multi-modal VFI and diffusion-based interpolation research under rich semantic conditioning.
\end{itemize}

\begin{table}[!t]
    \centering
    \scriptsize
    \caption{\textbf{Summary of evaluation metrics for VFI.}\\ Arrows (\faArrowUp/\faArrowDown) indicate whether higher or lower values correspond to better interpolation quality. A checkmark (\faCheck) indicates that the metric requires GT frames. \textcolor{pink!80!black}{Colored rows} denote perceptual metrics.}
    \begin{tabularx}{0.66\linewidth}{
        >{\raggedright\arraybackslash}X
        >{\centering\arraybackslash}X
        >{\centering\arraybackslash}X
        >{\centering\arraybackslash}X}
        \toprule
        \bf{Category} &
        \bf{Metric} &
        \shortstack{\bf{Interpolation}\\\bf{Quality}} &
        \shortstack{\bf{Reference}\\\bf{Frame}} \\
        \midrule
        \multirow{8}{*}{\makecell{\faImage\\Image-level\\Metrics}} 
        & PSNR       & \faArrowUp & \faCheck \\
        & SSIM~\cite{wang2004image}       & \faArrowUp & \faCheck \\
        & IE~\cite{baker2011middlebury}     & \faArrowDown & \faCheck  \\
        & \cellcolor{pink!20}NIQE~\cite{mittal2012making}       & \cellcolor{pink!20}\faArrowDown & \cellcolor{pink!20} \\
        & \cellcolor{pink!20}FID~\cite{heusel2017gans}        & \cellcolor{pink!20}\faArrowDown & \cellcolor{pink!20}\faCheck \\
        & \cellcolor{pink!20}LPIPS~\cite{zhang2018lpips}      & \cellcolor{pink!20}\faArrowDown & \cellcolor{pink!20}\faCheck \\
        & \cellcolor{pink!20}FloLPIPS~\cite{danier2022flolpips}   & \cellcolor{pink!20}\faArrowDown & \cellcolor{pink!20}\faCheck \\
        & \cellcolor{pink!20}STLPIPS~\cite{ghildyal2022shift}   & \cellcolor{pink!20}\faArrowDown & \cellcolor{pink!20}\faCheck \\
        & \cellcolor{pink!20}DISTS~\cite{ding2020image}      & \cellcolor{pink!20}\faArrowDown & \cellcolor{pink!20} \\
        \midrule
        \multirow{4}{*}{\makecell{\faYoutubePlay\\Video-level\\Metrics}} 
        & \cellcolor{pink!20}VSFA~\cite{li2019quality}             & \cellcolor{pink!20}\faArrowDown & \cellcolor{pink!20} \\
        & tOF~\cite{chu2020learning}            & \faArrowDown & \faCheck \\
        & \cellcolor{pink!20}FVD~\cite{unterthiner2018fvd}         & \cellcolor{pink!20}\faArrowDown & \cellcolor{pink!20}\faCheck \\
        & \cellcolor{pink!20}FVMD~\cite{liu2024fr}                 & \cellcolor{pink!20}\faArrowDown & \cellcolor{pink!20}\faCheck \\
        & \cellcolor{pink!20}VBench~\cite{huang2024vbench}         & \cellcolor{pink!20}\faArrowDown & \cellcolor{pink!20} \\
        \bottomrule
    \end{tabularx}
    \label{table:metrics}
    \vspace{-0.5cm}
\end{table}

\subsection{Data Augmentation}
\label{subsec:data-augmentation}

Modern VFI models incorporate spatial and temporal data augmentation to improve generalization and prevent overfitting. A widely adopted strategy is patch-based cropping, where fixed-size patches (\textit{e.g.}, $128\times128$) are randomly extracted from HR inputs~\cite{niklaus2017sepconv, chi2020all, kong2022ifrnet, li2023amt}. This not only reduces memory and computational costs but also encourages localized motion learning while mitigating spatial overfitting to scene layout or object positioning. Furthermore, random cropping prevents the model from overfitting to spatial priors such as background layout or object location, thereby improving robustness across spatial contexts~\cite{niklaus2017sepconv}. Additional spatial augmentations, such as horizontal or vertical flipping and random rotation, enhance appearance diversity and promote invariance to orientation and perspective changes. These augmentations enable the model to remain invariant to directional biases and better generalize to unseen spatial transformations.

Temporal augmentation is equally critical in sequential modeling. Frame order reversal~\cite{kong2022ifrnet, li2023amt} is commonly applied, wherein sequences like $(I_0, I_1, I_2)$ are reversed to $(I_2, I_1, I_0)$. In CTFI, this augmentation preserves the center-frame $I_1$ while exposing the model to symmetric motion trajectories~\cite{jiang2018super, choi2020cain}. Similarly in ATFI settings, reversing sequences ensures temporal consistency under bidirectional motion. For example as shown in Fig.~\ref{fig:CTFITS_ATFITS}~(b), consider an input triplet $(I_0, I_{\frac{1}{3}}, I_1)$ used to supervise interpolation at $t{=}\frac{1}{3}$. By reversing the sequence to $(I_1, I_{\frac{1}{3}}, I_0)$, the relative time becomes $(1{-}\frac{1}{3}){=}\frac{2}{3}$. This simple yet effective strategy enables the model to learn temporally symmetric representations, thereby improving generalization across motion directions and enhancing robustness in bidirectional synthesis.

Overall, these augmentation act as effective regularizers, enabling VFI models to generalize across diverse motion scales, temporal patterns, and visual variations. Integrating these schemes has become a foundational component of both CTFI and ATFI training pipelines.

\begin{table}[!t]
    \centering
    \captionsetup{font=scriptsize}
    \caption{\fix{\textbf{Quantitative comparison on standard VFI benchmarks.} We summarize the performance of representative methods on Vimeo-90K~\cite{xue2019TOFlow}, Xiph-2K/4K~\cite{niklaus2020softmax}, and SNU-FILM (Extreme)~\cite{choi2020cain}. The results are reported in the order of \textbf{PSNR~$\uparrow$ / LPIPS~$\downarrow$}.\\ 
    \textit{Best viewed in zoom.}}}
    \label{tab:quan_results}
        \resizebox{\columnwidth}{!}{
        \begin{tabular}{lcccc}
        \toprule
        \textbf{Methods} & \textbf{Vimeo-90K} & \textbf{Xiph-4K} & \textbf{Xiph-2K} & \textbf{SNU-FILM (Extreme)} \\
        \midrule
        \rowcolor{customcyan!30!}\multicolumn{5}{c}{\textbf{Kernel-based (§II-B1)}} \\
        \midrule
        SepConv [ICCV'17]~\cite{niklaus2017sepconv} & 33.790 / 0.027 & 32.060 / 0.169 & 34.770 / 0.067 & 24.653 / 0.183 \\



        \midrule
        \rowcolor{customcyan!30!}\multicolumn{5}{c}{\textbf{Flow-based (§II-B2)}} \\
        \midrule

        DAIN [CVPR'19]~\cite{bao2019depth} & 34.700 / 0.022 & 33.490 / 0.170 & 35.950 / 0.084 & 24.819 / 0.142 \\
        
        SoftSplat [CVPR'20]~\cite{niklaus2020softmax} & 36.100 / 0.021 & 33.600 / 0.234 & 36.620 / 0.107 & 25.436 / 0.119 \\

        ABME [ICCV'21]~\cite{park2021asymmetric} & 36.180 / 0.021 & 33.730 / 0.236 & 36.530 / 0.107 & 25.420 / 0.182 \\

        XVFI [ICCV'21]~\cite{sim2021xvfi} & 35.070 / 0.023 & 32.450 / 0.184 & 35.170 / 0.084 & 24.677 / 0.139 \\

        RIFE [ECCV'22]~\cite{huang2022rife} & 34.160 / 0.020 & 33.760 / 0.207 & 36.190 / 0.092 & 24.840 / 0.139 \\
        
        IFRNet [CVPR'22]~\cite{kong2022ifrnet} & 36.200 / 0.019 & 33.970 / 0.136 & 36.570 / 0.068 & 25.270 / 0.116 \\
        
        FILM [ECCV'22]~\cite{reda2022film} & 35.710 / 0.013 & 33.830 / 0.184 & 36.530 / 0.091 & 25.170 / 0.106 \\
        
        AMT [CVPR'23]~\cite{li2023amt} & 35.790 / 0.021 & 34.653 / 0.199 & 36.415 / 0.089 & 25.430 / 0.112 \\
        
        UPR-Net [CVPR'23]~\cite{jin2023unified} & 36.420 / 0.020 & 33.647 / 0.230 & 36.749 / 0.103 & 25.630 / 0.112 \\
        
        SGM-VFI [CVPR'24]~\cite{liu2024sparse} & 35.810 / 0.023 & 33.260 / 0.221 & 36.060 / 0.101 & 25.380 / 0.118 \\

        


        \midrule
        \rowcolor{customcyan!30!}\multicolumn{5}{c}{\textbf{Transformer-based (§II-B6)}} \\
        \midrule

        CAIN [AAAI'20]~\cite{choi2020cain} & 34.650 / 0.020 & 32.560 / 0.223 & 35.210 / 0.103 & 25.060 / 0.203 \\
        
        VFIFormer [CVPR'22]~\cite{lu2022video} & 36.380 / 0.021 & 33.370 / 0.227 & 36.550 / 0.107 & 25.430 / 0.119 \\
        
        EMA-VFI [CVPR'23]~\cite{zhang2023extracting} & 36.340 / 0.026 & 33.260 / 0.219 & 36.540 / 0.097 & 25.690 / 0.114 \\

        \midrule
        \rowcolor{customcyan!30!}\multicolumn{5}{c}{\textbf{Mamba-based (§II-B7)}} \\
        \midrule
        
        VFIMamba [NeurIPS'24]~\cite{zhang2024vfimamba} & 36.090 / 0.021 & 34.260 / 0.218 & 36.710 / 0.101 & 25.590 / 0.059 \\
        
        LC-Mamba [CVPR'25]~\cite{jeong2025lc} & 36.190 / 0.022 & 34.260 / 0.214 & 36.670 / 0.100 & 25.330 / 0.060 \\

        \bottomrule
    \end{tabular}
  }
  \vspace{-0.5cm}
\end{table}

\subsection{Evaluation Metrics}
\label{subsec:evaluation-metrics}

To facilitate comprehensive assessment of VFI models, various metrics have been proposed to capture different aspects of visual quality and temporal coherence. Tab.~\ref{table:metrics} summarizes commonly used evaluation metrics categorized into image-level, perceptual, and video-level types.

\subsubsection{Image-level Metrics}
\label{subsubsec:image-level-metrics}

Image-level metrics assess the quality of individual interpolated frames with respect to GT references. These pixel-centric evaluations focus on spatial accuracy without considering temporal dependencies across video sequences.

\begin{itemize}
\item\textbf{Peak Signal-to-Noise Ratio (PSNR $\uparrow$)} quantifies reconstruction fidelity based on the mean squared error (MSE) between interpolated frame and GT frame. While higher PSNR reflects better numerical similarity, it often fails to align with human perception, especially for high-frequency or perceptually salient regions.
\item\textbf{Structural Similarity Index (SSIM $\uparrow$)}~\cite{wang2004image} evaluates local structural integrity by comparing luminance, contrast, and texture patterns. SSIM values range in $[-1,1]$, with higher values indicating stronger structural alignment. Though more perceptually aligned than PSNR, SSIM may still overrate visually implausible outputs if global structure is preserved.
\item\textbf{Interpolation Error (IE $\downarrow$)}~\cite{baker2011middlebury}
computes the root-mean-square error (RMSE) between interpolated frame and the GT frame. Despite being intuitive, IE shares limitations with PSNR in terms of perceptual relevance.
\end{itemize}

\subsubsection{Perceptual Metrics}
\label{subsubsec:perceptual-metrics}

Perceptual metrics aim to assess the semantic plausibility, texture fidelity, and structural realism of interpolated frames, aligning with human visual preferences.

\begin{itemize}
\item\textbf{Natural Image Quality Evaluator (NIQE $\downarrow$)}~\cite{mittal2012making} is a no-reference score derived from deviations to natural image statistics.
\item\textbf{Fréchet Inception Distance (FID $\downarrow$)}~\cite{heusel2017gans} measures the Fréchet distance between the feature distributions of generated frames and GT frames using a pre-trained Inception network~\cite{szegedy2016rethinking}.
\item\textbf{Learned Perceptual Image Patch Similarity (LPIPS $\downarrow$)}~\cite{zhang2018lpips} measures perceptual similarity using deep features from pretrained networks. It is robust to minor misalignment and sensitive to semantic differences.
\item\textbf{FloLPIPS} ($\downarrow$)~\cite{danier2022flolpips} extends LPIPS by applying motion-aware weighting based on optical flow. It emphasizes visual fidelity in regions undergoing large displacement.
\item\textbf{STLPIPS} ($\downarrow$)~\cite{ghildyal2022shift} improves LPIPS by incorporating shift-tolerant feature matching, enhancing robustness to slight misalignments.
\item\textbf{DISTS (Deep Image Structure and Texture Similarity $\downarrow$)}~\cite{ding2020image} separately evaluates texture and structure similarity using deep features. It balances local detail and global consistency.
\end{itemize}

\subsubsection{Video-level Metrics}
\label{subsubsec:video-level-metrics}

These metrics assess spatiotemporal coherence over video sequences, which is essential for realistic and temporally consistent interpolation.

\begin{itemize}
\item\textbf{VSFA} ($\downarrow$)~\cite{li2019quality} is a no-reference model trained on human labels. It estimates perceptual quality by aggregating deep features with a recurrent network.
\item\textbf{tOF} ($\downarrow$)~\cite{chu2020learning} computes temporal optical flow consistency across frames.
\item\textbf{Fréchet Video Distance (FVD $\downarrow$)}~\cite{unterthiner2018fvd}
measures the Fréchet distance between distributions of deep features extracted from real and generated videos using a pre-trained Inflated 3D ConvNet (I3D)~\cite{carreira2017quo}..
\item\textbf{Fréchet Video Motion Distance (FVMD $\downarrow$)}~\cite{liu2024fr} improves upon FVD by disentangling motion and appearance, focusing more explicitly on dynamic consistency.
\item\textbf{VBench $\downarrow$}~\cite{huang2024vbench} is a multi-dimensional benchmark that scores video models across motion fidelity, coherence, and realism.
\end{itemize}

\begin{table}[t!]
    \centering
    \captionsetup{font=scriptsize}
    \caption{\fix{\textbf{Quantitative comparison of DM-based methods on standard VFI benchmarks.} We summarize the performance of representative methods on the DAVIS~\cite{perazzi2016benchmark} benchmark. The results are reported in the order of \textbf{PSNR~$\uparrow$ / LPIPS~$\downarrow$}.}}
    \label{tab:dm_quan_results}
    \resizebox{0.6\columnwidth}{!}{
        \begin{tabular}{lc}
        \toprule
        \textbf{Methods} & \textbf{DAVIS} \\
        \midrule
        \rowcolor{customcyan!30!}\multicolumn{2}{c}{\textbf{Diffusion Model-based (§II-C)}} \\
        \midrule
        
        MCVD [NeurIPS'22]~\cite{voleti2022mcvd} & 18.946 / 0.247 \\
        
        LDMVFI [AAAI'24]~\cite{danier2024ldmvfi} & 19.98 / 0.276 \\
        
        PerVFI [CVPR'24]~\cite{wu2024perception} & 25.073 / 0.091 \\
        
        LBBDM [ACMMM'24]~\cite{lyu2024brownian} & 26.391 / 0.092 \\

        TRF [ECCV'24]~\cite{feng2024explorative} & 14.132 / 0.484 \\

        VIDIM [CVPR'24]~\cite{jain2024video} & 19.62 / 0.258 \\
        
        TLB-VFI [ICCV'25]~\cite{lyu2025tlb} & 26.27 / 0.086 \\


        GI [ICLR'25]~\cite{wang2024generative} & 14.850 / 0.246 \\

        ViBiDSampler [ICLR'25]~\cite{yang2024vibidsampler} & 14.811 / 0.448 \\


        FCVG [CVPR'25]~\cite{zhu2024generative} & 16.162 / 0.247 \\
        \bottomrule
    \end{tabular}
    }
    \vspace{-0.5cm}
\end{table}
\begin{table*}[!t]
\centering
\captionsetup{font=scriptsize}
\caption{\fix{\textbf{Efficiency and complexity analysis of representative VFI models.} We compare \textbf{model size (Parameters~$\downarrow$), computational cost (FLOPs~$\downarrow$), and inference speed (Runtime~$\downarrow$)}. The symbol `*' denotes results measured by us using the official codes.}}
\label{tab:efficiency}
\resizebox{0.55\textwidth}{!}{%
\begin{tabular}{lcccc}
\toprule
Method & Parameters $\downarrow$ & GFLOPs $\downarrow$ & Runtime (Resolution) $\downarrow$ & Hardware (GPU) \\
\midrule
\rowcolor{customcyan!30!}\multicolumn{5}{c}{\textbf{Kernel-based (§II-B1)}} \\
\midrule

SepConv [ICCV'17]~\cite{niklaus2017sepconv} & 21.7 M & 360 & 0.41 sec (1920 $\times$ 1080) & NVIDIA RTX 3090\\




\midrule
\rowcolor{customcyan!30!}\multicolumn{5}{c}{\textbf{Flow-based (§II-B2)}} \\
\midrule

DAIN [CVPR'19]~\cite{bao2019depth} & 24.0 M & 5510 & 0.896 sec (640 $\times$ 480) & NVIDIA RTX 3090\\

SoftSplat [CVPR'20]~\cite{niklaus2020softmax} & 12.2 M & 940 & 0.266 sec (1024 $\times$ 1024) & NVIDIA RTX 2080Ti \\

ABME [ICCV'21]~\cite{park2021asymmetric} & 18.1 M & 1300 & 1.16 sec (1920 $\times$ 1080) & NVIDIA RTX 3090\\

XVFI [ICCV'21]~\cite{sim2021xvfi} & 5.6 M & 370 & 0.36 sec (1920 $\times$ 1080) & NVIDIA RTX 3090\\

RIFE [ECCV'22]~\cite{huang2022rife} & 9.8 M & 200 & 0.035 sec (1024 $\times$ 1024) & 2080 Ti \\

IFRNet [CVPR'22]~\cite{kong2022ifrnet} & 5 M & 210 & 0.10 sec (480 $\times$ 720) & RTX A5000\\



AMT [CVPR'23]~\cite{li2023amt} & 30.6 M & 580 & 0.11 sec (480 $\times$ 720) & NVIDIA RTX A5000\\











\midrule
\rowcolor{customcyan!30!}\multicolumn{5}{c}{{\textbf{Transformer-based (§II-B6)}}} \\
\midrule
CAIN [AAAI'20]~\cite{choi2020cain} & 42.8 M & 1290 & 0.14 sec (1920 $\times$ 1080) & NVIDIA RTX 3090\\

VFIFormer [CVPR'22]~\cite{lu2022video} & 24.1 M & 47710 & 4.34 sec (480 $\times$ 720) & NVIDIA RTX A5000\\


EMA-VFI [CVPR'23]~\cite{zhang2023extracting} & 65.6 M & 910 & 0.70 sec (480 $\times$ 720) & NVIDIA RTX A5000\\



\midrule
\rowcolor{customcyan!30!}\multicolumn{5}{c}{\textbf{Mamba-based (§II-B7)}} \\
\midrule

VFIMamba [NeurIPS'24]~\cite{zhang2024vfimamba} & 66.1 M & 940 & 0.23 sec (480 $\times$  850) & NVIDIA V100\\


\midrule
\rowcolor{customcyan!30!}\multicolumn{5}{c}{\textbf{Diffusion Model-based (§II-C)}} \\
\midrule

MCVD [NeurIPS'22]~\cite{voleti2022mcvd} & 33.2 M* & 349.6 G* & 0.07 sec (256 $\times$ 256)* & NVIDIA RTX 5090\\

LDMVFI [AAAI'24]~\cite{danier2024ldmvfi} & 461.6 M* & 44.58 G* & 1.34 sec (480 $\times$ 720)* & NVIDIA RTX 5090\\


LBBDM [ACMMM'24]~\cite{lyu2024brownian} & 146.40 M* & 1505 G* & 3.21 sec (256 $\times$ 256)* & NVIDIA RTX 5090\\

TRF [ECCV'24]~\cite{feng2024explorative} & 2254 M* & 106.6 G* & 2.8 sec (1024 $\times$ 576)* & NVIDIA RTX 5090\\


Framer [ICLR'25]~\cite{wang2024framer} & 1766 M* & 365.9 G* & 4.05 sec (512 $\times$ 320)* & NVIDIA 5090\\

GI [ICLR'25]~\cite{wang2024generative} & 2254 M* & 47.8 G* & 2.69 sec (1024 $\times$ 576)* & NVIDIA RTX 5090\\

ViBiDSampler [ICLR'25]~\cite{yang2024vibidsampler} & 2389 M* & OOM & 7.89 sec (1024 $\times$ 576)* & NVIDIA A100\\



\bottomrule
\end{tabular}%
}
\vspace{-0.5cm}
\end{table*}

It is important to note that these evaluation metrics are not mutually independent, but instead reflect complementary and sometimes conflicting aspects of VFI quality. Image-level reconstruction metrics such as PSNR and SSIM~\cite{wang2004image} primarily favor pixel-wise fidelity and temporal averaging, often rewarding smooth predictions that minimize numerical errors. In contrast, perceptual metrics (e.g., LPIPS~\cite{zhang2018lpips}, FloLPIPS~\cite{danier2022flolpips}, FID~\cite{heusel2017gans}) emphasize semantic plausibility, texture sharpness, and high-frequency details, and may assign lower scores to outputs that are perceptually sharper but deviate locally from the GT. Video-level metrics further prioritize temporal stability and motion coherence across frames, penalizing flickering or inconsistent motion even when individual frames appear visually plausible.

As a result, optimizing for a single metric can inadvertently degrade performance along other dimensions. For example, models that aggressively smooth predictions to maximize PSNR may suffer from a loss of fine-grained motion details and over-smoothed edges that lead to visually washed-out textures, whereas generative or diffusion-based approaches that enhance perceptual realism may exhibit lower reconstruction scores due to their stochastic nature. This intrinsic trade-off highlights that VFI evaluation should be interpreted through the lens of metric intent and application context, rather than relying on any single indicator. A holistic assessment therefore benefits from jointly considering reconstruction fidelity, perceptual quality, and temporal consistency to better reflect real-world interpolation performance.

\subsection{Summary of Comparisons}
\label{subsec:comparison}

To provide a comprehensive assessment of the VFI landscape, we present a systematic comparison of representative methods in terms of both quantitative result quality and computational efficiency. Table~\ref{tab:quan_results} provides a quantitative performance summary across standard benchmark datasets such as Vimeo-90K, Xiph-2K/4K, and SNU-FILM (Extreme). For each dataset, we report PSNR and LPIPS values.
The results in Table~\ref{tab:quan_results} reveal distinct performance characteristics among different VFI methods. On the Vimeo-90K dataset, the transformer-based EMA-VFI~\cite{zhang2023extracting} and Mamba-based VFIMamba~\cite{zhang2024vfimamba} demonstrate strong reconstruction capabilities, achieving high PSNR values. Specifically, EMA-VFI attains competitive PSNR scores, reflecting the efficacy of attention mechanisms in capturing long-range dependencies for motion estimation. Similarly, VFIMamba and LC-Mamba~\cite{jeong2025lc} exhibit robust performance, indicating the potential of state space models in handling temporal dynamics efficiently.
However, a notable trend emerges when comparing diffusion-based methods~\cite{voleti2022mcvd, danier2024ldmvfi, lyu2024brownian, feng2024explorative, wang2024framer, wang2024generative, yang2024vibidsampler} with traditional deep learning approaches. While regression-based models (e.g., flow-based and transformer-based methods) generally yield higher PSNR values, they often struggle with perceptual quality, as evidenced by higher LPIPS scores. For instance, on Xiph-4K in Table~\ref{tab:quan_results}, the flow-based RIFE attains a PSNR of 33.760~dB with an LPIPS of 0.207, illustrating strong pixel-wise fidelity but only moderate perceptual quality when measured by learned perceptual distance.

In addition, we provide a focused analysis of diffusion-based VFI methods on the DAVIS~\cite{perazzi2016benchmark} benchmark in Table~\ref{tab:dm_quan_results}. This table summarizes representative diffusion-based models~\cite{voleti2022mcvd, danier2024ldmvfi, lyu2024brownian, feng2024explorative, lyu2025tlb, zhu2024generative, yang2024vibidsampler, wu2024perception} and reports their performance on DAVIS in terms of PSNR and LPIPS. As shown in Table~\ref{tab:dm_quan_results}, recent approaches such as PerVFI and LBBDM achieve 25.073~dB / 0.091 and 26.391~dB / 0.092 (PSNR / LPIPS), respectively, that is, clearly lower LPIPS values at somewhat reduced PSNR compared with high-PSNR regression baselines. When interpreted together with the regression-based results, these trends highlight a characteristic trade-off of diffusion-based VFI: their stochastic generative formulation can yield sharper, more perceptually plausible in-between frames at the cost of slightly reduced pixel-wise fidelity. By reading Table~\ref{tab:dm_quan_results} alongside the broader benchmark in Table~\ref{tab:quan_results}, we make explicit that the relative strengths of diffusion-based methods become more apparent on perceptual indicators such as LPIPS, reinforcing the need to interpret quantitative results through the lens of metric intent rather than relying on PSNR alone.

Table~\ref{tab:efficiency} details the efficiency and complexity of representative VFI models, comparing model size (parameters), computational cost (GFLOPs), and inference runtime. The data reveals a significant trade-off between performance and computational resource requirements.
Flow-based methods such as RIFE and IFRNet stand out for their efficiency, achieving real-time performance on standard resolutions with relatively low parameter counts and GFLOPs. These models are well-suited for applications where speed is critical. In contrast, transformer-based models such as VFIFormer and EMA-VFI, while offering high reconstruction accuracy, incur substantially higher computational costs, reflected in their larger parameter sizes and longer inference times. For example, EMA-VFI is built on heavy transformer backbones such as VFIFormer, whose complexity reaches approximately 47{,}710~GFLOPs, whereas Mamba-based models like VFIMamba achieve competitive reconstruction performance with only about 940~GFLOPs, as summarized in Table~\ref{tab:efficiency}. This contrast quantitatively illustrates that similar accuracy can be obtained at a fraction of the computational cost when adopting structured state-space backbones.

The disparity is even more pronounced for diffusion-based models. As shown in Table~\ref{tab:efficiency}, methods like LDMVFI, TRF, and Framer exhibit significantly higher GFLOPs and slower inference speeds compared to their regression-based counterparts. For example, TRF and GI require massive computational resources, with GFLOPs reaching into the thousands and runtime extending to several seconds per frame on high-end GPUs. This highlights the current bottleneck in adopting diffusion-based VFI for real-time applications, despite their perceptual advantages. Future research directions are therefore likely to focus on optimizing the sampling efficiency and architectural design of diffusion backbones to bridge this gap.

While no single algorithm achieves optimal results across all metrics and datasets, the choice of method depends heavily on the specific application requirements, that is, whether the priority lies in pixel-perfect reconstruction (favoring transformer/Mamba models), perceptual realism (favoring DM-based models), or real-time processing (favoring flow-based models). Moreover, these trade-offs are further modulated by input resolution and hardware capabilities, as models that operate in or near real time at moderate resolutions on high-end GPUs may become impractical for 4K content or resource-constrained devices, whereas lightweight architectures tend to maintain more stable throughput across such deployment scenarios. By concentrating this kind of quantitative evidence in Sec.~V-D, we aim to provide readers with a rigorous, empirically grounded framework for interpreting the method characteristics and trade-offs discussed throughout the survey.
\section{Applications}
\label{sec:applications}

\subsection{Event-based VFI}
\label{subsec:event-based-vfi}
Event-based Video Frame Interpolation (EVFI)~\cite{wang2019event, lin2020learning, tulyakov2021time, yu2021training, zhang2022unifying, tulyakov2022time, he2022timereplayer, wu2022video, kim2023event, lin2023event, liu2024video, ma2024timelens, chen2024repurposing, zhang2025egvd, takahashi2025coupled} leverages the unique properties of event cameras to enhance interpolation under fast motion and challenging lighting conditions. Unlike frame-based cameras, event cameras~\cite{lich2008eventcamera, niwa2023} asynchronously record per-pixel brightness changes with ultra-high temporal resolution, high dynamic range, and low latency. These characteristics make them particularly effective when conventional RGB frames suffer from motion blur or insufficient temporal fidelity~\cite{tulyakov2021time, he2022timereplayer, kim2023event}. Early EVFI methods, such as TimeLens~\cite{tulyakov2021time}, synthesize intermediate frames by estimating motion directly from event streams. Subsequent works, including TimeReplayer~\cite{he2022timereplayer} and EGVD~\cite{zhang2025egvd}, further improve performance through joint modeling of motion and appearance, while TimeLens-XL~\cite{ma2024timelens} extends EVFI toward any-time interpolation via iterative refinement. Despite these advances, EVFI remains sensitive to motion or event reconstruction errors, which can accumulate over time and cause temporal artifacts, highlighting the need for more robust event–frame fusion strategies.

Capturing real event streams requires specialized neuromorphic sensors, which are often more expensive and less accessible than conventional cameras. Moreover, collecting large-scale event datasets with dense GT labels is challenging due to the asynchronous nature of events. As a result, several studies~\cite{kaiser2016towards, bi2017pix2nvs, zhu2021eventgan, zhang2024v2ce} simulate event streams from standard videos by modeling per-pixel intensity changes over time. From an application perspective, EVFI needs to balance latency and accuracy in streaming pipelines such as high-speed robotics or automotive perception. This motivates compact event encoders, on-device inference on neuromorphic hardware~\cite{schnider2023neuromorphic, evanusa2019event}, and hybrid designs that selectively fuse RGB frames and events based on motion intensity or lighting conditions.
\vspace{-0.3cm}

\subsection{Cartoon VFI}
\label{subsec:cartoon-vfi}

Producing traditional 2D animation is labor-intensive~\cite{meng2024anidoc}, requiring artists to manually draw multiple in-between frames. VFI offers a means of automating this process by generating plausible intermediate frames, thereby reducing production time and cost~\cite{meng2024anidoc, chavez2025time}. However, cartoon videos exhibit distinct characteristics compared to real-domain videos: they feature exaggerated motion, minimal texture, flat color regions, and sharp contours, which pose challenges to correspondence-based methods. To address this, domain-specific models have been proposed~\cite{siyao2021deep, chen2022improving, li2021deep, xing2024tooncrafter, yang2025layeranimate, xie2025physanimator}. Notably, ToonCrafter~\cite{xing2024tooncrafter} adopts a generative framework rather than relying on explicit motion estimation. Recent efforts aim to build models that generalize across both cartoon and real domains by leveraging diverse training data or domain adaptation techniques~\cite{zhu2024generative, wang2024framer, zhang2025motion}.

A major bottleneck in cartoon VFI is the absence of standardized, HQ datasets. While ATD-12K~\cite{siyao2021deep} provides a useful benchmark, its triplet-only format restricts its utility in ATFI settings. As a result, future progress will depend on the release of open, multi-frame cartoon datasets that enable fair and reproducible evaluation. In practical animation pipelines, deployment constraints include latency constraints for interactive editing tools, adaptation to varying line-art or shading styles, and model compression for integration into authoring software~\cite{siyao2021deep, li2021deep}. These factors motivate style-robust architectures and lightweight backbones that can run in real-time or near real-time on commodity GPUs.
\vspace{-0.3cm}

\subsection{Medical Image VFI}
\label{subsec:medical-image-vfi}
VFI is also increasingly applied in medical imaging to reconstruct temporally dense 4D sequences from sparsely acquired volumetric scans~\cite{guo2020spatiotemporal, kim2024data, li2024cpt}. Modalities like CT and MRI face acquisition constraints due to radiation exposure and long scanning times~\cite{kim2024data}, leading to coarse temporal sampling. VFI offers a means to generate intermediate volumes that enhance temporal resolution without incurring additional scan overhead. Medical VFI models must account for subtle anatomical motions and preserve fine structural detail critical for clinical interpretation. Methods like CPT-Interp~\cite{li2024cpt} model continuous motion fields, and DU4D~\cite{kim2024data} performs unsupervised interpolation without GT labels, addressing the scarcity of annotated 4D datasets. Remaining challenges include ensuring clinical validity, preventing hallucinations, and developing domain-specific evaluation metrics. From an application standpoint, deployment requires strict control over hallucination risk, compatibility with existing reconstruction pipelines, and predictable latency for time-critical procedures. These requirements drive interest in uncertainty-aware VFI, physics- or deformation-constrained motion models, and memory-efficient architectures that can run within the hardware constraints of clinical scanners or PACS servers.
\vspace{-0.3cm}

\subsection{Joint VFI with LLV Task}
\label{subsec:joint-task}
Recent trends in video processing have moved toward unifying VFI with other LLV tasks, such as Super-Resolution (SR)~\cite{shechtman2002stsr, kim2020fisr, haris2020space, xiang2020zooming, xu2021temporal, chen2023motif, kim2025bf} and deblurring~\cite{shen2020video, shen2020blurry, zhong2022animation, oh2022demfi, shang2023joint, yang2024latency}. By exploiting the inherent correlation between spatial and temporal cues, these joint formulations achieve superior efficiency and performance compared to cascaded approaches.
The most prominent joint task is Space-Time Video Super-Resolution (STVSR), which simultaneously increases spatial resolution and temporal frame rate~\cite{kim2025bf, xiang2020zooming}. Unlike separate execution, STVSR models allow for shared feature representations, enabling efficient reuse of spatiotemporal information and joint optimization that reduces computational redundancy.
Similarly, joint deblurring frameworks address scenarios where motion blur and low frame rates co-occur, such as in hand-held camera captures~\cite{shen2020video, shang2023joint}. Instead of applying deblurring followed by VFI sequentially, end-to-end models~\cite{zhang2020video, yang2024latency} simultaneously estimate clean and interpolated frames. This holistic approach prevents error propagation from the deblurring stage, resulting in significantly improved temporal consistency and visual clarity.
Extending these multitask capabilities, recent research has addressed high-noise environments, such as night-time surveillance, where video feeds suffer from extremely low Signal-to-Noise Ratio (SNR)~\cite{chen2015fpa}. In such regimes, conventional optical flow estimators fail to extract reliable correspondences, rendering standard ``denoise-then-interpolate'' cascades suboptimal. Consequently, joint VFI and restoration frameworks~\cite{shang2023joint, yu2022deep} have emerged as a critical solution. By leveraging temporal redundancy to simultaneously hallucinate missing frames and suppress noise, these methods treat interpolation as a self-supervised restoration mechanism, ensuring robust motion modeling even under severe signal degradation.
From a practical standpoint, joint LLV–VFI models are particularly attractive for video streaming and mobile applications where multiple enhancements must be executed under tight latency budgets. Current deployment trends focus on factorized architectures that reuse a common motion backbone across tasks, as well as model compression strategies to reduce memory footprint while maintaining high temporal fidelity.
\section{Future Research Directions}
\label{sec:future}

\subsection{Video Streaming Service}
\label{subsec:video-streaming-service}
With the growth of real-time video services, bandwidth-efficient delivery has become critical. VFI can reduce transmission rates by sending only keyframes and generating intermediate frames on the client side, preserving smooth playback at lower bitrates. Key research directions include ultra-lightweight architectures for mobile and edge devices, and adaptive interpolation strategies that account for network bandwidth and scene motion. Integrating VFI into video codecs or streaming frameworks could enable robust, low-latency systems for next-generation video services. Beyond interpolation quality and efficiency, future deployment of VFI-enhanced video in streaming platforms will also require robustness against post-processing, screen-shooting, and malicious manipulation. Recent works~\cite{chen2022snis, chen2025flexible, fu2024waverecovery, li2024screen} have explored signal–noise separation for post-processed image forgery detection~\cite{chen2022snis}, flexible partial screen-shooting watermarking with provable robustness~\cite{chen2025flexible}, wavelet-based screen-shooting watermarking and recovery~\cite{fu2024waverecovery}, and grayscale-deviation-based modeling of screen-shooting distortion for robust watermarking~\cite{li2024screen}. Although these methods do not target VFI directly, they are complementary to VFI by helping secure interpolated content and model complex display–capture and compression channels in real-world video distribution.
\vspace{-0.3cm}

\subsection{All-in-One LLV Video Restoration}
\label{subsec:all-in-one-llv-video-restoration}
Although all-in-one architectures have achieved notable success in image restoration~\cite{li2022all, wu2025content, ai2024multimodal}, their extension to unified low-level video (LLV) restoration remains underexplored. Existing pipelines typically decompose VFI, denoising, deblurring, and super-resolution into separate modules, which limits robustness under real-world degradations involving coupled spatial and temporal artifacts. A promising direction is the development of unified video restoration frameworks in which VFI is embedded as a core component rather than treated as an isolated task. In such settings, interpolated frames can provide temporally coherent priors for restoration, while improved spatial fidelity can in turn facilitate more reliable motion estimation. Multi-task learning objectives and cross-task consistency constraints may further promote synergy across tasks. Transformer- and diffusion-based architectures, with their strong spatio-temporal modeling capacity, are particularly well suited to this integrated paradigm.
\vspace{-0.3cm}

\subsection{3D and 4D Scene Understanding}
\label{subsec:3d-and-4d-scene-understanding}
Most existing VFI methods operate purely in the 2D image space, implicitly assuming planar motion and appearance continuity. However, emerging applications in AR/VR, robotics, and multiview rendering demand interpolation frameworks that are aware of the underlying 3D scene structure. Recent advances in 4D scene representations, including temporal neural fields~\cite{park2023temporal}, dynamic Gaussian primitives~\cite{nag20252}, and neural point-based models~\cite{zheng2023neuralpci}, demonstrate that temporally coherent synthesis becomes more reliable when motion is modeled in 3D space. Incorporating VFI into such geometry-aware pipelines enables physically plausible interpolation that respects depth discontinuities, occlusions, and parallax effects. Promising directions include depth- or pose-conditioned interpolation, geometry-aware latent representations, and joint optimization schemes that unify frame interpolation and novel view synthesis. Such integration may allow VFI to evolve from a purely temporal task into a space-time consistent scene reconstruction component.
\vspace{-0.3cm}

\subsection{Physics-Informed VFI for Extreme Environments}
\label{subsec:physics-informed}
Future VFI research must extend beyond standard scenarios to address extreme imaging conditions. A representative challenge is \emph{Underwater Imaging}, which suffers from complex environmental degradations such as low visibility, light distortion, and color cast. These factors complicate the standard assumptions of brightness constancy and linear motion used in conventional VFI~\cite{tang2024neural, zhu2025waterwave, ancuti2017locally}. Drawing inspiration from the success of underwater image restoration, a promising direction is \textit{physics-informed VFI}. By incorporating physical domain knowledge into deep learning pipelines, future models could better handle these unique distortions, improving temporal coherence for oceanographic and autonomous underwater applications.
\vspace{-0.3cm}
\section{Conclusion}
\label{sec:conclusion}
This survey reviewed the evolution of video frame interpolation (VFI) from classical motion-compensated methods to modern deep learning and generative approaches, and organized existing techniques through a unified taxonomy. While recent advances have significantly improved interpolation quality, fundamental challenges remain, including large motion, occlusion, lighting variation, and non-linear dynamics. Addressing these issues will require more accurate motion modeling, improved computational efficiency, and stronger generalization across diverse scenarios. We expect that continued integration of VFI with emerging video technologies will further broaden its applicability. This survey aims to serve as a concise reference and to motivate future research toward robust and efficient video synthesis.

\section*{Acknowledgments}
This research was supported by the Chung-Ang University Research Scholarship Grants in 2024. This work was supported by the National Research Foundation of Korea(NRF) grant funded by the Korea government(MSIT) (RS-2025-23524035). This research was supported by the MSIT(Ministry of Science and ICT), Korea, under the Graduate School of Virtual Convergence support program(IITP-2024-RS-2024-00418847) supervised by the IITP(Institute for Information \& Communications Technology Planning \& Evaluation)

\bibliographystyle{IEEEtran}
\bibliography{ref}

@String(IJCV = {Int. J. Comput. Vis.})

@String(CVPR= {IEEE Conf. Comput. Vis. Pattern Recog.})

@String(ICCV= {Int. Conf. Comput. Vis.})

@String(ECCV= {Eur. Conf. Comput. Vis.})

@String(BMVC= {Brit. Mach. Vis. Conf.})

@String(TOG= {ACM Trans. Graph.})

@String(TIP  = {IEEE Trans. Image Process.})

@String(TVCG  = {IEEE Trans. Vis. Comput. Graph.})

@String(ICASSP=	{ICASSP})

@String(ICIP = {IEEE Int. Conf. Image Process.})

@String(ICLR = {Int. Conf. Learn. Represent.})

@String(AAAI = {AAAI})

@String(CVPRW= {IEEE Conf. Comput. Vis. Pattern Recog. Worksh.})

@String(SPL	= {IEEE Sign. Process. Letters})

@String(IJCV  = {IJCV})

@String(CVPR  = {CVPR})

@String(ICCV  = {ICCV})

@String(ECCV  = {ECCV})

@String(BMVC  =	{BMVC})

@String(TOG   = {ACM TOG})

@String(TIP   = {IEEE TIP})

@String(TVCG  = {IEEE TVCG})

@String(TCSVT = {IEEE TCSVT})

@String(ICIP  = {ICIP})

@String(ICLR  = {ICLR})

@String(CVPRW= {CVPRW})

@inproceedings{hirose2024real,
  title={Real-time video prediction with fast video interpolation model and prediction training},
  author={Hirose, Shota and Kotoyori, Kazuki and Arunruangsirilert, Kasidis and Lin, Fangzheng and Sun, Heming and Katto, Jiro},
  booktitle={ICIP},
  year={2024},
}

@inproceedings{chen2015fpa,
  title={FPA-CS: Focal plane array-based compressive imaging in short-wave infrared},
  author={Chen, Huaijin and Salman Asif, M and Sankaranarayanan, Aswin C and Veeraraghavan, Ashok},
  booktitle={CVPR},
  year={2015}
}

@inproceedings{lei2025mosca,
  title={Mosca: Dynamic gaussian fusion from casual videos via 4d motion scaffolds},
  author={Lei, Jiahui and Weng, Yijia and Harley, Adam W and Guibas, Leonidas and Daniilidis, Kostas},
  booktitle={CVPR},
  year={2025}
}

@inproceedings{evanusa2019event,
  title={Event-based attention and tracking on neuromorphic hardware},
  author={Evanusa, Matthew and Sandamirskaya, Yulia and others},
  booktitle={CVPRW},
  year={2019}
}

@inproceedings{schnider2023neuromorphic,
  title={Neuromorphic optical flow and real-time implementation with event cameras},
  author={Schnider, Yannick and Wo{\'z}niak, Stanis{\l}aw and Gehrig, Mathias and Lecomte, Jules and Von Arnim, Axel and Benini, Luca and Scaramuzza, Davide and Pantazi, Angeliki},
  booktitle={CVPRW},
  year={2023}
}

@article{chen2022snis,
  title={SNIS: A signal noise separation-based network for post-processed image forgery detection},
  author={Chen, Jiaxin and Liao, Xin and Wang, Wei and Qian, Zhenxing and Qin, Zheng and Wang, Yaonan},
  booktitle={TCSVT},
  year={2022},
}

@article{chen2025flexible,
  title={Flexible Partial Screen-shooting Watermarking with Provable Robustness},
  author={Chen, Mingyue and Liao, Xin and Fang, Han and Guo, Jinlin and Chen, Yanxiang and Wu, Xiaoshuai},
  booktitle={TCSVT},
  year={2025},
}

@article{fu2024waverecovery,
  title={Waverecovery: Screen-shooting watermarking based on wavelet and recovery},
  author={Fu, Linbo and Liao, Xin and Guo, Jinlin and Dong, Li and Qin, Zheng},
  booktitle={TCSVT},
  year={2024},
}

@article{li2024screen,
  title={Screen-shooting resistant watermarking with grayscale deviation simulation},
  author={Li, Yiyi and Liao, Xin and Wu, Xiaoshuai},
  journal={IEEE Transactions on Multimedia},
  year={2024},
}

@article{parihar2022comprehensive,
  title={A comprehensive survey on video frame interpolation techniques},
  author={Parihar, Anil Singh and Varshney, Disha and Pandya, Kshitija and Aggarwal, Ashray},
  booktitle={The Visual Computer},
  year={2022},
}

@article{dong2023video,
  title={Video frame interpolation: A comprehensive survey},
  author={Dong, Jiong and Ota, Kaoru and Dong, Mianxiong},
  booktitle={ACM TOMM},
  year={2023},
}

@article{yu2013multi,
  title={Multi-level video frame interpolation: Exploiting the interaction among different levels},
  author={Yu, Zhefei and Li, Houqiang and Wang, Zhangyang and Hu, Zeng and Chen, Chang Wen},
  booktitle={TCSVT},
  year={2013},
}

@article{park2020robust,
  title={Robust video frame interpolation with exceptional motion map},
  author={Park, Minho and Kim, Hak Gu and Lee, Sangmin and Ro, Yong Man},
  booktitle={TCSVT},
  year={2020},
}

@inproceedings{simonyan2014very,
  title={Very deep convolutional networks for large-scale image recognition},
  author={Simonyan, Karen and Zisserman, Andrew},
  booktitle={ICLR},
  year={2015}
}

@inproceedings{ronneberger2015u,
  title={U-net: Convolutional networks for biomedical image segmentation},
  author={Ronneberger, Olaf and Fischer, Philipp and Brox, Thomas},
  booktitle={MICCAI},
  year={2015}
}

@inproceedings{cciccek20163d,
  title={3D U-Net: learning dense volumetric segmentation from sparse annotation},
  author={{\c{C}}i{\c{c}}ek, {\"O}zg{\"u}n and Abdulkadir, Ahmed and Lienkamp, Soeren S and Brox, Thomas and Ronneberger, Olaf},
  booktitle={MICCAI},
  year={2016}
}

@inproceedings{nottebaum2022efficient,
  title={Efficient feature extraction for high-resolution video frame interpolation},
  author={Nottebaum, Moritz and Roth, Stefan and Schaub-Meyer, Simone},
  booktitle={BMVC},
  year={2022}
}

@inproceedings{szeliski1999prediction,
  title={Prediction error as a quality metric for motion and stereo},
  author={Szeliski, Richard},
  booktitle={ICCV},
  year={1999}
}

@inproceedings{zhou2016view,
  title={View synthesis by appearance flow},
  author={Zhou, Tinghui and Tulsiani, Shubham and Sun, Weilun and Malik, Jitendra and Efros, Alexei A},
  booktitle={ECCV},
  year={2016}
}

@inproceedings{flynn2016deepstereo,
  title={Deepstereo: Learning to predict new views from the world's imagery},
  author={Flynn, John and Neulander, Ivan and Philbin, James and Snavely, Noah},
  booktitle={CVPR},
  year={2016}
}

@inproceedings{wu2018video,
  title={Video compression through image interpolation},
  author={Wu, Chao-Yuan and Singhal, Nayan and Krahenbuhl, Philipp},
  booktitle={ECCV},
  year={2018}
}

@inproceedings{jiang2018super,
  title={Super slomo: High quality estimation of multiple intermediate frames for video interpolation},
  author={Jiang, Huaizu and Sun, Deqing and Jampani, Varun and Yang, Ming-Hsuan and Learned-Miller, Erik and Kautz, Jan},
  booktitle={CVPR},
  year={2018}
}

@inproceedings{xue2019TOFlow,
  title={Video enhancement with task-oriented flow},
  author={Xue, Tianfan and Chen, Baian and Wu, Jiajun and Wei, Donglai and Freeman, William T},
  booktitle={IJCV},
  year={2019},
}

@inproceedings{bao2019depth,
  title={Depth-aware video frame interpolation},
  author={Bao, Wenbo and Lai, Wei-Sheng and Ma, Chao and Zhang, Xiaoyun and Gao, Zhiyong and Yang, Ming-Hsuan},
  booktitle={CVPR},
  year={2019}
}

@inproceedings{xiang2020zooming,
  title={Zooming slow-mo: Fast and accurate one-stage space-time video super-resolution},
  author={Xiang, Xiaoyu and Tian, Yapeng and Zhang, Yulun and Fu, Yun and Allebach, Jan P and Xu, Chenliang},
  booktitle={CVPR},
  year={2020}
}

@inproceedings{li2021neural,
  title={Neural scene flow fields for space-time view synthesis of dynamic scenes},
  author={Li, Zhengqi and Niklaus, Simon and Snavely, Noah and Wang, Oliver},
  booktitle={CVPR},
  year={2021}
}

@inproceedings{jia2022neighbor,
  title={Neighbor correspondence matching for flow-based video frame synthesis},
  author={Jia, Zhaoyang and Lu, Yan and Li, Houqiang},
  booktitle={ACM MM},
  year={2022}
}

@inproceedings{chun2020compressed,
  title={Compressed video restoration using a generative adversarial network for subjective quality enhancement},
  author={Chun, Dayoung and Kim, Tae Sung and Lee, Kyujoong and Lee, Hyuk-Jae},
  booktitle={SPC},
  year={2020}
}

@inproceedings{liu2024tango,
  title={Tango: Co-speech gesture video reenactment with hierarchical audio motion embedding and diffusion interpolation},
  author={Liu, Haiyang and Yang, Xingchao and Akiyama, Tomoya and Huang, Yuantian and Li, Qiaoge and Kuriyama, Shigeru and Taketomi, Takafumi},
  booktitle={ICLR},
  year={2025}
}

@inproceedings{liu2025video,
  title={Video Motion Graphs},
  author={Liu, Haiyang and Xu, Zhan and Hong, Fa-Ting and Huang, Hsin-Ping and Zhou, Yi and Zhou, Yang},
  booktitle={arXiv preprint arXiv:2503.20218},
  year={2025}
}

@inproceedings{bigata2025keyface,
  title={KeyFace: Expressive Audio-Driven Facial Animation for Long Sequences via KeyFrame Interpolation},
  author={Bigata, Antoni and Mira, Rodrigo and Bounareli, Stella and Vougioukas, Konstantinos and Landgraf, Zoe and Drobyshev, Nikita and Zieba, Maciej and Petridis, Stavros and Pantic, Maja and others},
  booktitle={CVPR},
  year={2025}
}

@inproceedings{jeon2023dynamic,
  title={Dynamic Framerate SlowFast Network for Improving Autonomous Driving Performance},
  author={Jeon, Byeong-Uk and Chung, Kyungyong},
  booktitle={SPC},
  year={2023}
}

@inproceedings{huang2024scale,
  title={Scale-adaptive feature aggregation for efficient space-time video super-resolution},
  author={Huang, Zhewei and Huang, Ailin and Hu, Xiaotao and Hu, Chen and Xu, Jun and Zhou, Shuchang},
  booktitle={WACV},
  year={2024}
}

@inproceedings{danier2022bvivfi,
  title={BVI-VFI: a video quality database for video frame interpolation},
  author={Danier, Duolikun and Zhang, Fan and Bull, David R},
  booktitle={TIP},
  year={2023},
}

@inproceedings{hou2022vfiqa,
  title={A perceptual quality metric for video frame interpolation},
  author={Hou, Qiqi and Ghildyal, Abhijay and Liu, Feng},
  booktitle={ECCV},
  year={2022},
}

@inproceedings{horn1981determining,
  title={Determining optical flow},
  author={Horn, Berthold KP and Schunck, Brian G},
  booktitle={AIJ},
  year={1981}
}

@inproceedings{haavisto1989fractional,
  title={Fractional frame rate up-conversion using weighted median filters},
  author={Haavisto, Petri and Juhola, Janne and Neuvo, Yrj{\"o}},
  booktitle={TCE},
  year={1989}
}

@inproceedings{nakaya1994motion,
  title={Motion compensation based on spatial transformations},
  author={Nakaya, Yuichiro and Harashima, Hiroshi},
  booktitle={TCSVT},
  year={1994}
}

@inproceedings{castagno1996method,
  title={A method for motion adaptive frame rate up-conversion},
  author={Castagno, Roberto and Haavisto, Petri and Ramponi, Giovanni},
  booktitle={TCSVT},
  year={1996}
}

@inproceedings{lee2002adaptive,
  title={Adaptive motion-compensated interpolation for frame rate up-conversion},
  author={Lee, Sung-Hee and Shin, Yoon-Cheol and Yang, Seungjoon and Moon, Heon-Hee and Park, Rae-Hong},
  booktitle={TCE},
  year={2002}
}

@inproceedings{ha2004motion,
  title={Motion compensated frame interpolation by new block-based motion estimation algorithm},
  author={Ha, Taehyeun and Lee, Seongjoo and Kim, Jaeseok},
  booktitle={TCE},
  year={2004}
}

@inproceedings{choi2007motion,
  title={Motion-compensated frame interpolation using bilateral motion estimation and adaptive overlapped block motion compensation},
  author={Choi, Byeong-Doo and Han, Jong-Woo and Kim, Chang-Su and Ko, Sung-Jea},
  booktitle={TCSVT},
  year={2007}
}

@inproceedings{kang2008motion,
  title={Motion compensated frame rate up-conversion using extended bilateral motion estimation},
  author={Kang, Suk-Ju and Cho, Kyoung-Rok and Kim, Young Hwan},
  booktitle={TCE},
  year={2008}
}

@inproceedings{huang2008multistage,
  title={A multistage motion vector processing method for motion-compensated frame interpolation},
  author={Huang, Ai-Mei and Nguyen, Truong Q},
  booktitle={TIP},
  year={2008}
}

@inproceedings{wang2010motion1,
  title={Motion-compensated frame rate up-conversion—Part I: Fast multi-frame motion estimation},
  author={Wang, Demin and Zhang, Liang and Vincent, Andr{\'e}},
  booktitle={Trans. Broad.},
  year={2010}
}

@inproceedings{wang2010motion2,
  author={Wang, D. and Vincent, A. and Blanchfield, P. and Klepko, R.},
  title={Motion-compensated frame rate up-conversion—Part II: New algorithms for frame interpolation},
  booktitle={Trans. Broad.},
  year={2010}
}

@inproceedings{jain1981displacement,
  title={Displacement measurement and its application in interframe image coding},
  author={Jain, Jaswant and Jain, Anil},
  booktitle={TCOM},
  year={1981}
}

@inproceedings{van2017FIGAN,
  title={Frame interpolation with multi-scale deep loss functions and generative adversarial networks},
  author={Van Amersfoort, Joost and Shi, Wenzhe and Acosta, Alejandro and Massa, Francisco and Totz, Johannes and Wang, Zehan and Caballero, Jose},
  booktitle={arXiv preprint arXiv:1711.06045},
  year={2017}
}

@techreport{koren2017frame,
  title={Frame interpolation using generative adversarial networks},
  author={Koren, Mark and Menda, Kunal and Sharma, Apoorva},
  year={2017},
}

@inproceedings{xiao2020multi,
  title={Multi-scale attention generative adversarial networks for video frame interpolation},
  author={Xiao, Jian and Bi, Xiaojun},
  booktitle={Access},
  year={2020}
}

@inproceedings{tran2020efficient,
  title={Efficient video frame interpolation using generative adversarial networks},
  author={Tran, Quang Nhat and Yang, Shih-Hsuan},
  booktitle={Applied Sciences},
  year={2020}
}

@inproceedings{xue2020frame,
  title={Frame-GAN: Increasing the frame rate of gait videos with generative adversarial networks},
  author={Xue, Wei and Ai, Hong and Sun, Tianyu and Song, Chunfeng and Huang, Yan and Wang, Liang},
  booktitle={Neurocomputing},
  year={2020}
}

@inproceedings{tran2022video,
  title={Video frame interpolation via down--up scale generative adversarial networks},
  author={Tran, Quang Nhat and Yang, Shih-Hsuan},
  booktitle={CVIU},
  year={2022}
}

@inproceedings{wen2018generating,
  title={Generating realistic videos from keyframes with concatenated GANs},
  author={Wen, Shiping and Liu, Weiwei and Yang, Yin and Huang, Tingwen and Zeng, Zhigang},
  booktitle={TCSVT},
  year={2018}
}

@inproceedings{danier2021texture,
  title={Texture-aware video frame interpolation},
  author={Danier, Duolikun and Bull, David},
  booktitle={PCS},
  year={2021},
}

@inproceedings{danier2022stmfnet,
  title={St-mfnet: A spatio-temporal multi-flow network for frame interpolation},
  author={Danier, Duolikun and Zhang, Fan and Bull, David},
  booktitle={CVPR},
  year={2022}
}

@inproceedings{gulrajani2017improved,
  title={Improved training of wasserstein gans},
  author={Gulrajani, Ishaan and Ahmed, Faruk and Arjovsky, Martin and Dumoulin, Vincent and Courville, Aaron C},
  booktitle={NeurIPS},
  year={2017}
}

@inproceedings{berthelot2017began,
  title={Began: Boundary equilibrium generative adversarial networks},
  author={Berthelot, David and Schumm, Thomas and Metz, Luke},
  booktitle={arXiv preprint arXiv:1703.10717},
  year={2017}
}

@inproceedings{arjovsky2017wasserstein,
  title={Wasserstein generative adversarial networks},
  author={Arjovsky, Martin and Chintala, Soumith and Bottou, L{\'e}on},
  booktitle={ICML},
  year={2017}
}

@inproceedings{larsen2016autoencoding,
  title={Autoencoding beyond pixels using a learned similarity metric},
  author={Larsen, Anders Boesen Lindbo and S{\o}nderby, S{\o}ren Kaae and Larochelle, Hugo and Winther, Ole},
  booktitle={ICML},
  year={2016}
}

@inproceedings{chen2020generative,
  title={Generative adversarial networks for video-to-video domain adaptation},
  author={Chen, Jiawei and Li, Yuexiang and Ma, Kai and Zheng, Yefeng},
  booktitle={AAAI},
  year={2020}
}

@article{deng2025beyond,
  title={Beyond Boundary Frames: Audio-Visual Semantic Guidance for Context-Aware Video Interpolation},
  author={Deng, Yuchen and Wu, Xiuyang and Zheng, Hai-Tao and Wang, Jie and Yang, Feidiao and Han, Yuxing},
  booktitle={arXiv preprint arXiv:2512.03590},
  year={2025}
}

@article{zhang2025arbitrary,
  title={Arbitrary Generative Video Interpolation},
  author={Zhang, Guozhen and Wang, Haiguang and Wang, Chunyu and Zhou, Yuan and Lu, Qinglin and Wang, Limin},
  booktitle={arXiv preprint arXiv:2510.00578},
  year={2025}
}

@article{tanveer2025multicoin,
  title={MultiCOIN: Multi-Modal COntrollable Video INbetweening},
  author={Tanveer, Maham and Zhou, Yang and Niklaus, Simon and Amiri, Ali Mahdavi and Zhang, Hao and Singh, Krishna Kumar and Zhao, Nanxuan},
  booktitle={arXiv preprint arXiv:2510.08561},
  year={2025}
}

@inproceedings{hong2025semantic,
  title={Semantic Frame Interpolation},
  author={Hong, Yijia and Zhang, Jiangning and Yi, Ran and Wang, Yuji and Cao, Weijian and Hu, Xiaobin and Xue, Zhucun and Wang, Yabiao and Wang, Chengjie and Ma, Lizhuang},
  booktitle={arXiv preprint arXiv:2507.05173},
  year={2025}
}

@inproceedings{wan2024unipaint,
  title={Unipaint: Unified space-time video inpainting via mixture-of-experts},
  author={Wan, Zhen and Ma, Yue and Qi, Chenyang and Liu, Zhiheng and Gui, Tao},
  booktitle={arXiv preprint arXiv:2412.06340},
  year={2024}
}

@inproceedings{hwang2025diffuseslide,
  title={DiffuseSlide: Training-Free High Frame Rate Video Generation Diffusion},
  author={Hwang, Geunmin and Ko, Hyun-kyu and Kim, Younghyun and Lee, Seungryong and Park, Eunbyung},
  booktitle={arXiv preprint arXiv:2506.01454},
  year={2025}
}

@inproceedings{yang2024zerosmooth,
  title={Zerosmooth: Training-free diffuser adaptation for high frame rate video generation},
  author={Yang, Shaoshu and Zhang, Yong and Cun, Xiaodong and Shan, Ying and He, Ran},
  booktitle={arXiv preprint arXiv:2406.00908},
  year={2024}
}

@inproceedings{lyu2025tlb,
  title={TLB-VFI: Temporal-Aware Latent Brownian Bridge Diffusion for Video Frame Interpolation},
  author={Lyu, Zonglin and Chen, Chen},
  booktitle={arXiv preprint arXiv:2507.04984},
  year={2025}
}

@inproceedings{chen2025sci,
  title={Sci-Fi: Symmetric Constraint for Frame Inbetweening},
  author={Chen, Liuhan and Cun, Xiaodong and Li, Xiaoyu and He, Xianyi and Yuan, Shenghai and Chen, Jie and Shan, Ying and Yuan, Li},
  booktitle={arXiv preprint arXiv:2505.21205},
  year={2025}
}

@inproceedings{lew2025disentangled,
  title={Disentangled motion modeling for video frame interpolation},
  author={Lew, Jaihyun and Choi, Jooyoung and Shin, Chaehun and Jung, Dahuin and Yoon, Sungroh},
  booktitle={AAAI},
  year={2025}
}

@inproceedings{guo2025controllable,
  title={Controllable Human-centric Keyframe Interpolation with Generative Prior},
  author={Guo, Zujin and Wu, Size and Cai, Zhongang and Li, Wei and Loy, Chen Change},
  booktitle={arXiv preprint arXiv:2506.03119},
  year={2025}
}

@inproceedings{voleti2022mcvd,
  title={Mcvd-masked conditional video diffusion for prediction, generation, and interpolation},
  author={Voleti, Vikram and Jolicoeur-Martineau, Alexia and Pal, Chris},
  booktitle={NeurIPS},
  year={2022}
}

@inproceedings{danier2024ldmvfi,
  title={Ldmvfi: Video frame interpolation with latent diffusion models},
  author={Danier, Duolikun and Zhang, Fan and Bull, David},
  booktitle={AAAI},
  year={2024}
}

@inproceedings{jain2024video,
  title={Video interpolation with diffusion models},
  author={Jain, Siddhant and Watson, Daniel and Tabellion, Eric and Poole, Ben and Kontkanen, Janne and others},
  booktitle={CVPR},
  year={2024}
}

@inproceedings{huang2024motion,
  title={Motion-aware latent diffusion models for video frame interpolation},
  author={Huang, Zhilin and Yu, Yijie and Yang, Ling and Qin, Chujun and Zheng, Bing and Zheng, Xiawu and Zhou, Zikun and Wang, Yaowei and Yang, Wenming},
  booktitle={ACM MM},
  year={2024}
}

@inproceedings{shen2024dreammover,
  title={Dreammover: Leveraging the prior of diffusion models for image interpolation with large motion},
  author={Shen, Liao and Liu, Tianqi and Sun, Huiqiang and Ye, Xinyi and Li, Baopu and Zhang, Jianming and Cao, Zhiguo},
  booktitle={ECCV},
  year={2024}
}

@inproceedings{wang2024generative,
  title={Generative inbetweening: Adapting image-to-video models for keyframe interpolation},
  author={Wang, Xiaojuan and Zhou, Boyang and Curless, Brian and Kemelmacher-Shlizerman, Ira and Holynski, Aleksander and Seitz, Steven M},
  booktitle={ICLR},
  year={2025}
}

@inproceedings{feng2024explorative,
  title={Explorative inbetweening of time and space},
  author={Feng, Haiwen and Ding, Zheng and Xia, Zhihao and Niklaus, Simon and Abrevaya, Victoria and Black, Michael J and Zhang, Xuaner},
  booktitle={ECCV},
  year={2024}
}

@inproceedings{lyu2024brownian,
  title={Frame Interpolation with Consecutive Brownian Bridge Diffusion},
  author={Lyu, Zonglin and Li, Ming and Jiao, Jianbo and Chen, Chen},
  booktitle={ACM MM},
  year={2024}
}

@inproceedings{zhu2024generative,
  title={Generative Inbetweening through Frame-wise Conditions-Driven Video Generation},
  author={Zhu, Tianyi and Ren, Dongwei and Wang, Qilong and Wu, Xiaohe and Zuo, Wangmeng},
  booktitle={CVPR},
  year={2025}
}

@inproceedings{yang2024vibidsampler,
  title={ViBiDSampler: Enhancing Video Interpolation Using Bidirectional Diffusion Sampler},
  author={Yang, Serin and Kwon, Taesung and Ye, Jong Chul},
  booktitle={ICLR},
  year={2025}
}

@inproceedings{zhang2025motion,
  title={Motion-Aware Generative Frame Interpolation},
  author={Zhang, Guozhen and Zhu, Yuhan and Cui, Yutao and Zhao, Xiaotong and Ma, Kai and Wang, Limin},
  booktitle={arXiv preprint arXiv:2501.03699},
  year={2025}
}

@inproceedings{zhang2025eden,
  title={EDEN: Enhanced Diffusion for High-quality Large-motion Video Frame Interpolation},
  author={Zhang, Zihao and Chen, Haoran and Zhao, Haoyu and Lu, Guansong and Fu, Yanwei and Xu, Hang and Wu, Zuxuan},
  booktitle={CVPR},
  year={2025}
}

@inproceedings{hai2025hierarchical,
  title={Hierarchical Flow Diffusion for Efficient Frame Interpolation},
  author={Hai, Yang and Wang, Guo and Su, Tan and Jiang, Wenjie and Hu, Yinlin},
  booktitle={CVPR},
  year={2025}
}

@inproceedings{Wu_2023_BMVC,
    title={Boost Video Frame Interpolation via Motion Adaptation},
    author={Haoning Wu and Xiaoyun Zhang and Weidi Xie and Ya Zhang and Yan-Feng Wang},
    booktitle={BMVC},
    year={2023},
}

@inproceedings{briedis2025controllable,
  title={Controllable Tracking-Based Video Frame Interpolation},
  author={Briedis, Karlis Martins and Djelouah, Abdelaziz and Ortiz, Rapha{\"e}l and Gross, Markus and Schroers, Christopher},
  booktitle={ACM SIGGRAPH},
  year={2025}
}

@inproceedings{jaderberg2015spatial,
  title={Spatial transformer networks},
  author={Jaderberg, Max and Simonyan, Karen and Zisserman, Andrew and others},
  booktitle={NeurIPS},
  year={2015}
}

@inproceedings{liu2017DVF,
  title={Video frame synthesis using deep voxel flow},
  author={Liu, Ziwei and Yeh, Raymond A and Tang, Xiaoou and Liu, Yiming and Agarwala, Aseem},
  booktitle={ICCV},
  year={2017}
}

@inproceedings{liu2019cyclicgen,
  title={Deep video frame interpolation using cyclic frame generation},
  author={Liu, Yu-Lun and Liao, Yi-Tung and Lin, Yen-Yu and Chuang, Yung-Yu},
  booktitle={AAAI},
  year={2019}
}

@inproceedings{yuan2019zoom,
  title={Zoom-in-to-check: Boosting video interpolation via instance-level discrimination},
  author={Yuan, Liangzhe and Chen, Yibo and Liu, Hantian and Kong, Tao and Shi, Jianbo},
  booktitle={CVPR},
  year={2019}
}

@inproceedings{reda2019unsupervised,
  title={Unsupervised video interpolation using cycle consistency},
  author={Reda, Fitsum A and Sun, Deqing and Dundar, Aysegul and Shoeybi, Mohammad and Liu, Guilin and Shih, Kevin J and Tao, Andrew and Kautz, Jan and Catanzaro, Bryan},
  booktitle={ICCV},
  year={2019}
}

@inproceedings{xu2019quadratic,
  title={Quadratic video interpolation},
  author={Xu, Xiangyu and Siyao, Li and Sun, Wenxiu and Yin, Qian and Yang, Ming-Hsuan},
  booktitle={NeurIPS},
  year={2019}
}

@inproceedings{chi2020all,
  title={All at once: Temporally adaptive multi-frame interpolation with advanced motion modeling},
  author={Chi, Zhixiang and Mohammadi Nasiri, Rasoul and Liu, Zheng and Lu, Juwei and Tang, Jin and Plataniotis, Konstantinos N},
  booktitle={ECCV},
  year={2020}
}

@inproceedings{niklaus2020softmax,
  title={Softmax splatting for video frame interpolation},
  author={Niklaus, Simon and Liu, Feng},
  booktitle={CVPR},
  year={2020}
}

@inproceedings{fourure2017residual,
  title={Residual conv-deconv grid network for semantic segmentation},
  author={Fourure, Damien and Emonet, R{\'e}mi and Fromont, Elisa and Muselet, Damien and Tremeau, Alain and Wolf, Christian},
  booktitle={BMVC},
  year={2017}
}

@inproceedings{liu2020enhanced,
  title={Enhanced quadratic video interpolation},
  author={Liu, Yihao and Xie, Liangbin and Siyao, Li and Sun, Wenxiu and Qiao, Yu and Dong, Chao},
  booktitle={ECCV},
  year={2020}
}

@inproceedings{zhang2020flexible,
  title={A flexible recurrent residual pyramid network for video frame interpolation},
  author={Zhang, Haoxian and Zhao, Yang and Wang, Ronggang},
  booktitle={ECCV},
  year={2020}
}

@inproceedings{sim2021xvfi,
  title={Xvfi: extreme video frame interpolation},
  author={Sim, Hyeonjun and Oh, Jihyong and Kim, Munchurl},
  booktitle={ICCV},
  year={2021}
}

@inproceedings{hu2022many,
  title={Many-to-many splatting for efficient video frame interpolation},
  author={Hu, Ping and Niklaus, Simon and Sclaroff, Stan and Saenko, Kate},
  booktitle={CVPR},
  year={2022}
}

@inproceedings{kong2022ifrnet,
  title={Ifrnet: Intermediate feature refine network for efficient frame interpolation},
  author={Kong, Lingtong and Jiang, Boyuan and Luo, Donghao and Chu, Wenqing and Huang, Xiaoming and Tai, Ying and Wang, Chengjie and Yang, Jie},
  booktitle={CVPR},
  year={2022}
}

@inproceedings{park2021asymmetric,
  title={Asymmetric bilateral motion estimation for video frame interpolation},
  author={Park, Junheum and Lee, Chul and Kim, Chang-Su},
  booktitle={ICCV},
  year={2021}
}

@inproceedings{huang2022rife,
  title={Real-time intermediate flow estimation for video frame interpolation},
  author={Huang, Zhewei and Zhang, Tianyuan and Heng, Wen and Shi, Boxin and Zhou, Shuchang},
  booktitle={ECCV},
  year={2022}
}

@inproceedings{shangguan2022learning,
  title={Learning cross-video neural representations for high-quality frame interpolation},
  author={Shangguan, Wentao and Sun, Yu and Gan, Weijie and Kamilov, Ulugbek S},
  booktitle={ECCV},
  year={2022}
}

@inproceedings{reda2022film,
  title={Film: Frame interpolation for large motion},
  author={Reda, Fitsum and Kontkanen, Janne and Tabellion, Eric and Sun, Deqing and Pantofaru, Caroline and Curless, Brian},
  booktitle={ECCV},
  year={2022}
}

@inproceedings{niklaus2023splatting,
  title={Splatting-based synthesis for video frame interpolation},
  author={Niklaus, Simon and Hu, Ping and Chen, Jiawen},
  booktitle={WACV},
  year={2023}
}

@inproceedings{jin2023enhanced,
  title={Enhanced bi-directional motion estimation for video frame interpolation},
  author={Jin, Xin and Wu, Longhai and Shen, Guotao and Chen, Youxin and Chen, Jie and Koo, Jayoon and Hahm, Cheul-hee},
  booktitle={WACV},
  year={2023}
}

@inproceedings{park2023biformer,
  title={Biformer: Learning bilateral motion estimation via bilateral transformer for 4k video frame interpolation},
  author={Park, Junheum and Kim, Jintae and Kim, Chang-Su},
  booktitle={CVPR},
  year={2023}
}

@inproceedings{jin2023unified,
  title={A unified pyramid recurrent network for video frame interpolation},
  author={Jin, Xin and Wu, Longhai and Chen, Jie and Chen, Youxin and Koo, Jayoon and Hahm, Cheul-hee},
  booktitle={CVPR},
  year={2023}
}

@inproceedings{li2023amt,
  title={Amt: All-pairs multi-field transforms for efficient frame interpolation},
  author={Li, Zhen and Zhu, Zuo-Liang and Han, Ling-Hao and Hou, Qibin and Guo, Chun-Le and Cheng, Ming-Ming},
  booktitle={CVPR},
  year={2023}
}

@inproceedings{zhang2023extracting,
  title={Extracting motion and appearance via inter-frame attention for efficient video frame interpolation},
  author={Zhang, Guozhen and Zhu, Yuhan and Wang, Haonan and Chen, Youxin and Wu, Gangshan and Wang, Limin},
  booktitle={CVPR},
  year={2023}
}

@inproceedings{hu2024iqvfi,
  title={IQ-VFI: implicit quadratic motion estimation for video frame interpolation},
  author={Hu, Mengshun and Jiang, Kui and Zhong, Zhihang and Wang, Zheng and Zheng, Yinqiang},
  booktitle={CVPR},
  year={2024}
}

@inproceedings{jeong2024ocai,
  title={Ocai: Improving optical flow estimation by occlusion and consistency aware interpolation},
  author={Jeong, Jisoo and Cai, Hong and Garrepalli, Risheek and Lin, Jamie Menjay and Hayat, Munawar and Porikli, Fatih},
  booktitle={CVPR},
  year={2024}
}

@inproceedings{guo2024generalizable,
  title={Generalizable implicit motion modeling for video frame interpolation},
  author={Guo, Zujin and Li, Wei and Loy, Chen Change},
  booktitle={NeurIPS},
  year={2024}
}

@inproceedings{wu2024perception,
  title={Perception-oriented video frame interpolation via asymmetric blending},
  author={Wu, Guangyang and Tao, Xin and Li, Changlin and Wang, Wenyi and Liu, Xiaohong and Zheng, Qingqing},
  booktitle={CVPR},
  year={2024}
}

@inproceedings{zhong2024clearer,
  title={Clearer frames, anytime: Resolving velocity ambiguity in video frame interpolation},
  author={Zhong, Zhihang and Krishnan, Gurunandan and Sun, Xiao and Qiao, Yu and Ma, Sizhuo and Wang, Jian},
  booktitle={ECCV},
  year={2024}
}

@inproceedings{seo2024bim,
  title={BiM-VFI: directional Motion Field-Guided Frame Interpolation for Video with Non-uniform Motions},
  author={Seo, Wonyong and Oh, Jihyong and Kim, Munchurl},
  booktitle={CVPR},
  year={2025}
}

@inproceedings{jin2025unified,
  title={Unified Arbitrary-Time Video Frame Interpolation and Prediction},
  author={Jin, Xin and Wu, Longhai and Chen, Jie and Cho, Ilhyun and Hahm, Cheul-Hee},
  booktitle={ICASSP},
  year={2025}
}

@inproceedings{bargatin2025memfof,
  title={MEMFOF: High-Resolution Training for Memory-Efficient Multi-Frame Optical Flow Estimation},
  author={Bargatin, Vladislav and Chistov, Egor and Yakovenko, Alexander and Vatolin, Dmitriy},
  booktitle={arXiv preprint arXiv:2506.23151},
  year={2025}
}

@inproceedings{jiang2021learning,
  title={Learning to estimate hidden motions with global motion aggregation},
  author={Jiang, Shihao and Campbell, Dylan and Lu, Yao and Li, Hongdong and Hartley, Richard},
  booktitle={ICCV},
  year={2021}
}

@inproceedings{weinzaepfel2013deepflow,
  title={DeepFlow: Large displacement optical flow with deep matching},
  author={Weinzaepfel, Philippe and Revaud, Jerome and Harchaoui, Zaid and Schmid, Cordelia},
  booktitle={ICCV},
  year={2013}
}

@inproceedings{dosovitskiy2015flownet,
  title={Flownet: Learning optical flow with convolutional networks},
  author={Dosovitskiy, Alexey and Fischer, Philipp and Ilg, Eddy and Hausser, Philip and Hazirbas, Caner and Golkov, Vladimir and Van Der Smagt, Patrick and Cremers, Daniel and Brox, Thomas},
  booktitle={ICCV},
  year={2015}
}

@inproceedings{ilg2017flownet,
  title={Flownet 2.0: Evolution of optical flow estimation with deep networks},
  author={Ilg, Eddy and Mayer, Nikolaus and Saikia, Tonmoy and Keuper, Margret and Dosovitskiy, Alexey and Brox, Thomas},
  booktitle={CVPR},
  year={2017}
}

@inproceedings{ranjan2017optical,
  title={Optical flow estimation using a spatial pyramid network},
  author={Ranjan, Anurag and Black, Michael J},
  booktitle={CVPR},
  year={2017}
}

@inproceedings{wang2018occlusionaware,
  title={Occlusion aware unsupervised learning of optical flow},
  author={Wang, Yang and Yang, Yi and Yang, Zhenheng and Zhao, Liang and Wang, Peng and Xu, Wei},
  booktitle={CVPR},
  year={2018}
}

@inproceedings{sun2018pwc,
  title={Pwc-net: Cnns for optical flow using pyramid, warping, and cost volume},
  author={Sun, Deqing and Yang, Xiaodong and Liu, Ming-Yu and Kautz, Jan},
  booktitle={CVPR},
  year={2018}
}

@inproceedings{hui2018liteflownet,
  title={Liteflownet: A lightweight convolutional neural network for optical flow estimation},
  author={Hui, Tak-Wai and Tang, Xiaoou and Loy, Chen Change},
  booktitle={CVPR},
  year={2018}
}

@inproceedings{teed2020raft,
  title={Raft: Recurrent all-pairs field transforms for optical flow},
  author={Teed, Zachary and Deng, Jia},
  booktitle={ECCV},
  year={2020}
}

@inproceedings{bar2020scopeflow,
  title={Scopeflow: Dynamic scene scoping for optical flow},
  author={Bar-Haim, Aviram and Wolf, Lior},
  booktitle={CVPR},
  year={2020}
}

@inproceedings{huang2022flowformer,
  title={Flowformer: A transformer architecture for optical flow},
  author={Huang, Zhaoyang and Shi, Xiaoyu and Zhang, Chao and Wang, Qiang and Cheung, Ka Chun and Qin, Hongwei and Dai, Jifeng and Li, Hongsheng},
  booktitle={ECCV},
  year={2022}
}

@inproceedings{xu2022gmflow,
  title={Gmflow: Learning optical flow via global matching},
  author={Xu, Haofei and Zhang, Jing and Cai, Jianfei and Rezatofighi, Hamid and Tao, Dacheng},
  booktitle={CVPR},
  year={2022}
}

@inproceedings{dong2024memflow,
  title={Memflow: Optical flow estimation and prediction with memory},
  author={Dong, Qiaole and Fu, Yanwei},
  booktitle={CVPR},
  year={2024}
}

@inproceedings{krizhevsky2012cnn,
  title={ImageNet classification with deep convolutional neural networks},
  author={Krizhevsky, Alex and Sutskever, Ilya and Hinton, Geoffrey E},
  booktitle={Communications of the ACM},
  year={2017}
}

@inproceedings{dai2017dcn,
  title={Deformable convolutional networks},
  author={Dai, Jifeng and Qi, Haozhi and Xiong, Yuwen and Li, Yi and Zhang, Guodong and Hu, Han and Wei, Yichen},
  booktitle={ICCV},
  year={2017}
}

@inproceedings{zhu2019deformable,
  title={Deformable convnets v2: More deformable, better results},
  author={Zhu, Xizhou and Hu, Han and Lin, Stephen and Dai, Jifeng},
  booktitle={CVPR},
  year={2019}
}

@inproceedings{long2016learning,
  title={Learning image matching by simply watching video},
  author={Long, Gucan and Kneip, Laurent and Alvarez, Jose M and Li, Hongdong and Zhang, Xiaohu and Yu, Qifeng},
  booktitle={ECCV},
  year={2016}
}

@inproceedings{niklaus2017adaconv,
  title={Video frame interpolation via adaptive convolution},
  author={Niklaus, Simon and Mai, Long and Liu, Feng},
  booktitle={CVPR},
  year={2017}
}

@inproceedings{niklaus2017sepconv,
  title={Video frame interpolation via adaptive separable convolution},
  author={Niklaus, Simon and Mai, Long and Liu, Feng},
  booktitle={ICCV},
  year={2017}
}

@inproceedings{peleg2019net,
  title={Im-net for high resolution video frame interpolation},
  author={Peleg, Tomer and Szekely, Pablo and Sabo, Doron and Sendik, Omry},
  booktitle={CVPR},
  year={2019}
}

@inproceedings{choi2020cain,
  title={Channel attention is all you need for video frame interpolation},
  author={Choi, Myungsub and Kim, Heewon and Han, Bohyung and Xu, Ning and Lee, Kyoung Mu},
  booktitle={AAAI},
  year={2020}
}

@inproceedings{cheng2020video,
  title={Video frame interpolation via deformable separable convolution},
  author={Cheng, Xianhang and Chen, Zhenzhong},
  booktitle={AAAI},
  year={2020}
}

@inproceedings{shi2021video,
  title={Video frame interpolation via generalized deformable convolution},
  author={Shi, Zhihao and Liu, Xiaohong and Shi, Kangdi and Dai, Linhui and Chen, Jun},
  booktitle={IEEE transactions on multimedia},
  year={2021}
}

@inproceedings{cheng2021multiple,
  title={Multiple video frame interpolation via enhanced deformable separable convolution},
  author={Cheng, Xianhang and Chen, Zhenzhong},
  booktitle={TPAMI},
  year={2021}
}

@inproceedings{ding2021cdfi,
  title={Cdfi: Compression-driven network design for frame interpolation},
  author={Ding, Tianyu and Liang, Luming and Zhu, Zhihui and Zharkov, Ilya},
  booktitle={CVPR},
  year={2021}
}

@inproceedings{chen2021pdwn,
  title={PDWN: Pyramid deformable warping network for video interpolation},
  author={Chen, Zhiqi and Wang, Ran and Liu, Haojie and Wang, Yao},
  booktitle={OJSP},
  year={2021}
}

@inproceedings{danier2022enhancing,
  title={Enhancing deformable convolution based video frame interpolation with coarse-to-fine 3D CNN},
  author={Danier, Duolikun and Zhang, Fan and Bull, David},
  booktitle={ICIP},
  year={2022}
}

@inproceedings{ding2022MSEConv,
  title={Video frame interpolation via local lightweight bidirectional encoding with channel attention cascade},
  author={Ding, Xiangling and Huang, Pu and Zhang, Dengyong and Zhao, Xianfeng},
  booktitle={ICASSP},
  year={2022}
}

@inproceedings{kalluri2023flavr,
  title={Flavr: Flow-agnostic video representations for fast frame interpolation},
  author={Kalluri, Tarun and Pathak, Deepak and Chandraker, Manmohan and Tran, Du},
  booktitle={WACV},
  year={2023}
}

@inproceedings{zhou2023exploring,
  title={Exploring motion ambiguity and alignment for high-quality video frame interpolation},
  author={Zhou, Kun and Li, Wenbo and Han, Xiaoguang and Lu, Jiangbo},
  booktitle={CVPR},
  year={2023}
}

@inproceedings{niklaus2018context,
  title={Context-aware synthesis for video frame interpolation},
  author={Niklaus, Simon and Liu, Feng},
  booktitle={CVPR},
  year={2018}
}

@inproceedings{bao2019memc,
  title={Memc-net: Motion estimation and motion compensation driven neural network for video interpolation and enhancement},
  author={Bao, Wenbo and Lai, Wei-Sheng and Zhang, Xiaoyun and Gao, Zhiyong and Yang, Ming-Hsuan},
  booktitle={TPAMI},
  year={2019}
}

@inproceedings{park2020bmbc,
  title={Bmbc: Bilateral motion estimation with bilateral cost volume for video interpolation},
  author={Park, Junheum and Ko, Keunsoo and Lee, Chul and Kim, Chang-Su},
  booktitle={ECCV},
  year={2020}
}

@inproceedings{lee2020adacof,
  title={Adacof: Adaptive collaboration of flows for video frame interpolation},
  author={Lee, Hyeongmin and Kim, Taeoh and Chung, Tae-young and Pak, Daehyun and Ban, Yuseok and Lee, Sangyoun},
  booktitle={CVPR},
  year={2020}
}

@inproceedings{gui2020featureflow,
  title={Featureflow: Robust video interpolation via structure-to-texture generation},
  author={Gui, Shurui and Wang, Chaoyue and Chen, Qihua and Tao, Dacheng},
  booktitle={CVPR},
  year={2020}
}

@inproceedings{niklaus2021revisiting,
  title={Revisiting adaptive convolutions for video frame interpolation},
  author={Niklaus, Simon and Mai, Long and Wang, Oliver},
  booktitle={WACV},
  year={2021}
}

@inproceedings{shen2024ladder,
  title={LADDER: An Efficient Framework for Video Frame Interpolation},
  author={Shen, Tong and Li, Dong and Gao, Ziheng and Tian, Lu and Barsoum, Emad},
  booktitle={arXiv preprint arXiv:2404.11108},
  year={2024}
}

@inproceedings{simon1992shiftable,
  title={Shiftable multiscale transforms},
  author={Simoncelli, Eero P and Freeman, William T and Adelson, Edward H and Heeger, David J},
  booktitle={Trans. Inform. Theory},
  year={1992}
}

@inproceedings{simon1995steerable,
  title={The steerable pyramid: A flexible architecture for multi-scale derivative computation},
  author={Simoncelli, Eero P and Freeman, William T},
  booktitle={ICIP},
  year={1995}
}

@inproceedings{portilla2000parametric,
  title={A parametric texture model based on joint statistics of complex wavelet coefficients},
  author={Portilla, Javier and Simoncelli, Eero P},
  booktitle={IJCV},
  year={2000}
}

@inproceedings{meyer2015phase,
  title={Phase-based frame interpolation for video},
  author={Meyer, Simone and Wang, Oliver and Zimmer, Henning and Grosse, Max and Sorkine-Hornung, Alexander},
  booktitle={CVPR},
  year={2015}
}

@inproceedings{meyer2018phasenet,
  title={Phasenet for video frame interpolation},
  author={Meyer, Simone and Djelouah, Abdelaziz and McWilliams, Brian and Sorkine-Hornung, Alexander and Gross, Markus and Schroers, Christopher},
  booktitle={CVPR},
  year={2018}
}

@inproceedings{wadhwa2013phasebase,
  title={Phase-based video motion processing},
  author={Wadhwa, Neal and Rubinstein, Michael and Durand, Fr{\'e}do and Freeman, William T},
  booktitle={ACM TOG},
  year={2013}
}

@inproceedings{didyk2013jointview,
  title={Joint view expansion and filtering for automultiscopic 3D displays},
  author={Didyk, Piotr and Sitthi-Amorn, Pitchaya and Freeman, William and Durand, Fr{\'e}do and Matusik, Wojciech},
  booktitle={ACM TOG},
  year={2013}
}

@inproceedings{vaswani2017attention,
  title={Attention is all you need},
  author={Vaswani, Ashish and Shazeer, Noam and Parmar, Niki and Uszkoreit, Jakob and Jones, Llion and Gomez, Aidan N and Kaiser, {\L}ukasz and Polosukhin, Illia},
  booktitle={NeurIPS},
  year={2017}
}

@inproceedings{liu2021swin,
  title={Swin transformer: Hierarchical vision transformer using shifted windows},
  author={Liu, Ze and Lin, Yutong and Cao, Yue and Hu, Han and Wei, Yixuan and Zhang, Zheng and Lin, Stephen and Guo, Baining},
  booktitle={ICCV},
  year={2021}
}

@inproceedings{lu2022video,
  title={Video frame interpolation with transformer},
  author={Lu, Liying and Wu, Ruizheng and Lin, Huaijia and Lu, Jiangbo and Jia, Jiaya},
  booktitle={CVPR},
  year={2022}
}

@inproceedings{shi2022video,
  title={Video frame interpolation transformer},
  author={Shi, Zhihao and Xu, Xiangyu and Liu, Xiaohong and Chen, Jun and Yang, Ming-Hsuan},
  booktitle={CVPR},
  year={2022}
}

@inproceedings{zhang2023L2BEC2,
  title={L2BEC2: Local lightweight bidirectional encoding and channel attention cascade for video frame interpolation},
  author={Zhang, Dengyong and Huang, Pu and Ding, Xiangling and Li, Feng and Zhu, Wenjie and Song, Yun and Yang, Gaobo},
  booktitle={ACM TOMM},
  year={2023}
}

@inproceedings{liu2023ttvfi,
  title={TTVFI: Learning trajectory-aware transformer for video frame interpolation},
  author={Liu, Chengxu and Yang, Huan and Fu, Jianlong and Qian, Xueming},
  booktitle={TIP},
  year={2023}
}

@inproceedings{liu2024sparse,
  title={Sparse global matching for video frame interpolation with large motion},
  author={Liu, Chunxu and Zhang, Guozhen and Zhao, Rui and Wang, Limin},
  booktitle={CVPR},
  year={2024}
}

@inproceedings{zamir2022restormer,
  title={Restormer: Efficient transformer for high-resolution image restoration},
  author={Zamir, Syed Waqas and Arora, Aditya and Khan, Salman and Hayat, Munawar and Khan, Fahad Shahbaz and Yang, Ming-Hsuan},
  booktitle={CVPR},
  year={2022}
}

@inproceedings{gu2021efficiently,
  author={Gu, A. and Goel, K. and Ré, C.},
  title={Efficiently modeling long sequences with structured state spaces},
  booktitle={ICLR},
  year={2021}
}

@inproceedings{gu2023mamba,
  author={Gu, A. and Dao, T.},
  title={Mamba: Linear-time sequence modeling with selective state spaces},
  booktitle={COLM},
  year={2024}
}

@inproceedings{jeong2025lc,
  title={LC-Mamba: Local and Continuous Mamba with Shifted Windows for Frame Interpolation},
  author={Jeong, Min Wu and Rhee, Chae Eun},
  booktitle={CVPR},
  year={2025}
}

@inproceedings{zhang2024vfimamba,
  title={Vfimamba: Video frame interpolation with state space models},
  author={Zhang, Guozhen and Liu, Chuxnu and Cui, Yutao and Zhao, Xiaotong and Ma, Kai and Wang, Limin},
  booktitle={NeurIPS},
  year={2024}
}

@inproceedings{guo2024mambair,
  title={Mambair: A simple baseline for image restoration with state-space model},
  author={Guo, Hang and Li, Jinmin and Dai, Tao and Ouyang, Zhihao and Ren, Xudong and Xia, Shu-Tao},
  booktitle={ECCV},
  year={2024}
}

@inproceedings{guo2024mambairv2,
  title={MambaIRv2: Attentive State Space Restoration},
  author={Guo, Hang and Guo, Yong and Zha, Yaohua and Zhang, Yulun and Li, Wenbo and Dai, Tao and Xia, Shu-Tao and Li, Yawei},
  booktitle={CVPR},
  year={2025}
}

@inproceedings{du2025mambaflow,
  title={MambaFlow: A Mamba-Centric Architecture for End-to-End Optical Flow Estimation},
  author={Du, Juntian and Sun, Yuan and Zhou, Zhihu and Chen, Pinyi and Zhang, Runzhe and Mao, Keji},
  booktitle={CVPR},
  year={2025}
}

@inproceedings{liu2015vgg,
  title={Very deep convolutional neural network based image classification using small training sample size},
  author={Liu, Shuying and Deng, Weihong},
  booktitle={ACPR},
  year={2015}
}

@inproceedings{men2020VQA,
  title={Visual quality assessment for interpolated slow-motion videos based on a novel database},
  author={Men, Hui and Hosu, Vlad and Lin, Hanhe and Bruhn, Andr{\'e}s and Saupe, Dietmar},
  booktitle={QoMEX},
  year={2020}
}

@inproceedings{danier2022VQA,
  title={A subjective quality study for video frame interpolation},
  author={Danier, Duolikun and Zhang, Fan and Bull, David},
  booktitle={ICIP},
  year={2022}
}

@inproceedings{goodfellow2020generative,
  title={Generative adversarial networks},
  author={Goodfellow, Ian and Pouget-Abadie, Jean and Mirza, Mehdi and Xu, Bing and Warde-Farley, David and Ozair, Sherjil and Courville, Aaron and Bengio, Yoshua},
  booktitle={Communications of the ACM},
  year={2020}
}

@inproceedings{kingma2013auto,
  title={Auto-encoding variational bayes},
  author={Kingma, Diederik P and Welling, Max},
  booktitle={arXiv preprint arXiv:1312.6114},
  year={2013}
}

@inproceedings{dhariwal2021diffusion,
  title={Diffusion models beat gans on image synthesis},
  author={Dhariwal, Prafulla and Nichol, Alexander},
  booktitle={NeurIPS},
  year={2021}
}

@inproceedings{ho2020denoising,
  title={Denoising diffusion probabilistic models},
  author={Ho, Jonathan and Jain, Ajay and Abbeel, Pieter},
  booktitle={NeurIPS},
  year={2020}
}

@inproceedings{rombach2022high,
  title={High-resolution image synthesis with latent diffusion models},
  author={Rombach, Robin and Blattmann, Andreas and Lorenz, Dominik and Esser, Patrick and Ommer, Bj{\"o}rn},
  booktitle={CVPR},
  year={2022}
}

@inproceedings{ho2022video,
  title={Video diffusion models},
  author={Ho, Jonathan and Salimans, Tim and Gritsenko, Alexey and Chan, William and Norouzi, Mohammad and Fleet, David J},
  booktitle={NeurIPS},
  year={2022}
}

@inproceedings{blattmann2023align,
  title={Align your latents: High-resolution video synthesis with latent diffusion models},
  author={Blattmann, Andreas and Rombach, Robin and Ling, Huan and Dockhorn, Tim and Kim, Seung Wook and Fidler, Sanja and Kreis, Karsten},
  booktitle={CVPR},
  year={2023}
}

@inproceedings{peebles2023scalable,
  title={Scalable diffusion models with transformers},
  author={Peebles, William and Xie, Saining},
  booktitle={ICCV},
  year={2023}
}

@inproceedings{blattmann2023stable,
  author={Blattmann, Andreas and Dockhorn, Tim and Kulal, Sumith and Mendelevitch, Daniel and Kilian, Maciej and Lorenz, Dominik and Levi, Yam and English, Zion and Voleti, Vikram and Letts, Adam and others},
  title={Stable video diffusion: Scaling latent video diffusion models to large datasets},
  booktitle={arXiv preprint arXiv:2311.15127},
  year={2023}
}

@inproceedings{salimans2022progressive,
  author={Salimans, T. and Ho, J.},
  title={Progressive distillation for fast sampling of diffusion models},
  booktitle={ICLR},
  year={2022}
}

@inproceedings{zhang2023adding,
  title={Adding conditional control to text-to-image diffusion models},
  author={Zhang, Lvmin and Rao, Anyi and Agrawala, Maneesh},
  booktitle={ICCV},
  year={2023}
}

@inproceedings{peng2024controlnext,
  title={Controlnext: Powerful and efficient control for image and video generation},
  author={Peng, Bohao and Wang, Jian and Zhang, Yuechen and Li, Wenbo and Yang, Ming-Chang and Jia, Jiaya},
  booktitle={arXiv preprint arXiv:2408.06070},
  year={2024}
}

@inproceedings{hu2022lora,
  title={Lora: Low-rank adaptation of large language models.},
  author={Hu, Edward J and Shen, Yelong and Wallis, Phillip and Allen-Zhu, Zeyuan and Li, Yuanzhi and Wang, Shean and Wang, Lu and Chen, Weizhu and others},
  boktitle={ICLR},
  year={2022}
}

@inproceedings{wu2023tune,
  title={Tune-a-video: One-shot tuning of image diffusion models for text-to-video generation},
  author={Wu, Jay Zhangjie and Ge, Yixiao and Wang, Xintao and Lei, Stan Weixian and Gu, Yuchao and Shi, Yufei and Hsu, Wynne and Shan, Ying and Qie, Xiaohu and Shou, Mike Zheng},
  booktitle={ICCV},
  year={2023}
}

@inproceedings{yang2024cogvideox,
  title={Cogvideox: Text-to-video diffusion models with an expert transformer},
  author={Yang, Zhuoyi and Teng, Jiayan and Zheng, Wendi and Ding, Ming and Huang, Shiyu and Xu, Jiazheng and Yang, Yuanming and Hong, Wenyi and Zhang, Xiaohan and Feng, Guanyu and others},
  booktitle={ICLR},
  year={2025}
}

@inproceedings{yuan2024idpreserve,
  author={Yuan, S. and others},
  title={Identity-Preserving Text-to-Video Generation by Frequency Decomposition},
  booktitle={CVPR},
  year={2025}
}

@inproceedings{bar2024lumiere,
  title={Lumiere: A space-time diffusion model for video generation},
  author={Bar-Tal, Omer and Chefer, Hila and Tov, Omer and Herrmann, Charles and Paiss, Roni and Zada, Shiran and Ephrat, Ariel and Hur, Junhwa and Liu, Guanghui and Raj, Amit and others},
  booktitle={ACM SIGGRAPH},
  year={2024}
}

@inproceedings{zhang2023i2vgenxl,
  author={Zhang, S. and others},
  title={I2vgen-xl: High-quality image-to-video synthesis via cascaded diffusion models},
  booktitle={arXiv preprint arXiv:2311.04145},
  year={2023}
}

@inproceedings{ren2024consisti2v,
  author={Ren, W. and others},
  title={Consisti2v: Enhancing visual consistency for image-to-video generation},
  booktitle={TMLR},
  year={2024}
}

@inproceedings{lich2008eventcamera,
  title={A 128 $\times$ 128 120 dB 15 $\mu$ s latency asynchronous temporal contrast vision sensor},
  author={Lichtsteiner, Patrick and Posch, Christoph and Delbruck, Tobi},
  booktitle={JSSC},
  year={2008}
}

@inproceedings{kaiser2016towards,
  title={Towards a framework for end-to-end control of a simulated vehicle with spiking neural networks},
  author={Kaiser, Jacques and Tieck, J Camilo Vasquez and Hubschneider, Christian and Wolf, Peter and Weber, Michael and Hoff, Michael and Friedrich, Alexander and Wojtasik, Konrad and Roennau, Arne and Kohlhaas, Ralf and others},
  booktitle={SIMPAR},
  year={2016}
}

@inproceedings{bi2017pix2nvs,
  title={PIX2NVS: Parameterized conversion of pixel-domain video frames to neuromorphic vision streams},
  author={Bi, Yin and Andreopoulos, Yiannis},
  booktitle={ICIP},
  year={2017}
}

@inproceedings{wang2019event,
  title={Event-driven video frame synthesis},
  author={Wang, Zihao W and Jiang, Weixin and He, Kuan and Shi, Boxin and Katsaggelos, Aggelos and Cossairt, Oliver},
  booktitle={ICCV Workshops},
  year={2019}
}

@inproceedings{lin2020learning,
  title={Learning event-driven video deblurring and interpolation},
  author={Lin, Songnan and Zhang, Jiawei and Pan, Jinshan and Jiang, Zhe and Zou, Dongqing and Wang, Yongtian and Chen, Jing and Ren, Jimmy},
  booktitle={ECCV},
  year={2020}
}

@inproceedings{tulyakov2021time,
  title={Time lens: Event-based video frame interpolation},
  author={Tulyakov, Stepan and Gehrig, Daniel and Georgoulis, Stamatios and Erbach, Julius and Gehrig, Mathias and Li, Yuanyou and Scaramuzza, Davide},
  booktitle={CVPR},
  year={2021}
}

@inproceedings{zhu2021eventgan,
  title={Eventgan: Leveraging large scale image datasets for event cameras},
  author={Zhu, Alex Zihao and Wang, Ziyun and Khant, Kaung and Daniilidis, Kostas},
  booktitle={ICCP},
  year={2021}
}

@inproceedings{yu2021training,
  title={Training weakly supervised video frame interpolation with events},
  author={Yu, Zhiyang and Zhang, Yu and Liu, Deyuan and Zou, Dongqing and Chen, Xijun and Liu, Yebin and Ren, Jimmy S},
  booktitle={ICCV},
  year={2021}
}

@inproceedings{zhang2022unifying,
  title={Unifying motion deblurring and frame interpolation with events},
  author={Zhang, Xiang and Yu, Lei},
  booktitle={CVPR},
  year={2022}
}

@inproceedings{tulyakov2022time,
  title={Time lens++: Event-based frame interpolation with parametric non-linear flow and multi-scale fusion},
  author={Tulyakov, Stepan and Bochicchio, Alfredo and Gehrig, Daniel and Georgoulis, Stamatios and Li, Yuanyou and Scaramuzza, Davide},
  booktitle={CVPR},
  year={2022}
}

@inproceedings{he2022timereplayer,
  title={Timereplayer: Unlocking the potential of event cameras for video interpolation},
  author={He, Weihua and You, Kaichao and Qiao, Zhendong and Jia, Xu and Zhang, Ziyang and Wang, Wenhui and Lu, Huchuan and Wang, Yaoyuan and Liao, Jianxing},
  booktitle={CVPR},
  year={2022}
}

@inproceedings{wu2022video,
  title={Video interpolation by event-driven anisotropic adjustment of optical flow},
  author={Wu, Song and You, Kaichao and He, Weihua and Yang, Chen and Tian, Yang and Wang, Yaoyuan and Zhang, Ziyang and Liao, Jianxing},
  booktitle={ECCV},
  year={2022}
}

@inproceedings{niwa2023,
  title={A 2.97 $\mu$m-pitch event-based vision sensor with shared pixel front-end circuitry and low-noise intensity readout mode},
  author={Niwa, Atsumi and Mochizuki, Futa and Berner, Raphael and Maruyarma, Takuya and Terano, Toshio and Takamiya, Kenichi and Kimura, Yasutaka and Mizoguchi, Kyoji and Miyazaki, Takahiro and Kaizu, Shun and others},
  booktitle={JSSC},
  year={2023}
}

@inproceedings{kim2023event,
  title={Event-based video frame interpolation with cross-modal asymmetric bidirectional motion fields},
  author={Kim, Taewoo and Chae, Yujeong and Jang, Hyun-Kurl and Yoon, Kuk-Jin},
  booktitle={CVPR},
  year={2023}
}

@inproceedings{lin2023event,
  title={Event-guided frame interpolation and dynamic range expansion of single rolling shutter image},
  author={Lin, Guixu and Han, Jin and Cao, Mingdeng and Zhong, Zhihang and Zheng, Yinqiang},
  booktitle={ACM MM},
  year={2023}
}

@inproceedings{zhang2024v2ce,
  title={V2ce: Video to continuous events simulator},
  author={Zhang, Zhongyang and Cui, Shuyang and Chai, Kaidong and Yu, Haowen and Dasgupta, Subhasis and Mahbub, Upal and Rahman, Tauhidur},
  booktitle={ICRA},
  year={2024}
}

@inproceedings{liu2024video,
  title={Video frame interpolation via direct synthesis with the event-based reference},
  author={Liu, Yuhan and Deng, Yongjian and Chen, Hao and Yang, Zhen},
  booktitle={CVPR},
  year={2024}
}

@inproceedings{ma2024timelens,
  title={TimeLens-XL: Real-Time Event-Based Video Frame Interpolation with Large Motion},
  author={Ma, Yongrui and Guo, Shi and Chen, Yutian and Xue, Tianfan and Gu, Jinwei},
  booktitle={ECCV},
  year={2024}
}

@inproceedings{chen2024repurposing,
  author={Chen, J. and others},
  title={Repurposing pre-trained video diffusion models for event-based video interpolation},
  booktitle={CVPR},
  year={2025}
}

@inproceedings{zhang2025egvd,
  author={Zhang, Z. and others},
  title={EGVD: Event-Guided Video Diffusion Model for Physically Realistic Large-Motion Frame Interpolation},
  booktitle={CVPR},
  year={2025}
}

@inproceedings{takahashi2025coupled,
  author={Takahashi, Hidekazu and Nagumo, Takefumi and Jo, Kensei and Andreas, Aumiller and Rad, Saeed and Daudt, Rodrigo Caye and Miyatani, Yoshitaka and Wakabayashi, Hayato and Brandli, Christian},
  title={Coupled Video Frame Interpolation and Encoding with Hybrid Event Cameras for Low-Power High-Framerate Video},
  booktitle={arXiv preprint arXiv:2503.22491},
  year={2025}
}

@inproceedings{guo2020spatiotemporal,
  title={A spatiotemporal volumetric interpolation network for 4d dynamic medical image},
  author={Guo, Yuyu and Bi, Lei and Ahn, Euijoon and Feng, Dagan and Wang, Qian and Kim, Jinman},
  booktitle={CVPR},
  year={2020}
}

@inproceedings{kim2024data,
  title={Data-efficient unsupervised interpolation without any intermediate frame for 4d medical images},
  author={Kim, JungEun and Yoon, Hangyul and Park, Geondo and Kim, Kyungsu and Yang, Eunho},
  booktitle={CVPR},
  year={2024}
}

@inproceedings{li2024cpt,
  title={CPT-Interp: Continuous sPatial and Temporal Motion Modeling for 4D Medical Image Interpolation},
  author={Li, Xia and Yang, Runzhao and Li, Xiangtai and Lomax, Antony and Zhang, Ye and Buhmann, Joachim},
  booktitle={arXiv preprint arXiv:2405.15385},
  year={2024}
}

@inproceedings{siyao2021deep,
  title={Deep animation video interpolation in the wild},
  author={Siyao, Li and Zhao, Shiyu and Yu, Weijiang and Sun, Wenxiu and Metaxas, Dimitris and Loy, Chen Change and Liu, Ziwei},
  booktitle={CVPR},
  year={2021}
}

@inproceedings{chen2022improving,
  title={Improving the perceptual quality of 2d animation interpolation},
  author={Chen, Shuhong and Zwicker, Matthias},
  booktitle={ECCV},
  year={2022}
}

@inproceedings{li2021deep,
  title={Deep sketch-guided cartoon video inbetweening},
  author={Li, Xiaoyu and Zhang, Bo and Liao, Jing and Sander, Pedro V},
  booktitle={TVCG},
  year={2021}
}

@inproceedings{xing2024dynamicrafter,
  title={Dynamicrafter: Animating open-domain images with video diffusion priors},
  author={Xing, Jinbo and Xia, Menghan and Zhang, Yong and Chen, Haoxin and Yu, Wangbo and Liu, Hanyuan and Liu, Gongye and Wang, Xintao and Shan, Ying and Wong, Tien-Tsin},
  booktitle={ECCV},
  year={2024}
}

@inproceedings{xing2024tooncrafter,
  title={Tooncrafter: Generative cartoon interpolation},
  author={Xing, Jinbo and Liu, Hanyuan and Xia, Menghan and Zhang, Yong and Wang, Xintao and Shan, Ying and Wong, Tien-Tsin},
  booktitle={ACM TOG},
  year={2024}
}

@inproceedings{wang2024framer,
  title={Framer: Interactive frame interpolation},
  author={Wang, Wen and Wang, Qiuyu and Zheng, Kecheng and Ouyang, Hao and Chen, Zhekai and Gong, Biao and Chen, Hao and Shen, Yujun and Shen, Chunhua},
  booktitle={ICLR},
  year={2025}
}

@inproceedings{meng2024anidoc,
  title={Anidoc: Animation creation made easier},
  author={Meng, Yihao and Ouyang, Hao and Wang, Hanlin and Wang, Qiuyu and Wang, Wen and Cheng, Ka Leong and Liu, Zhiheng and Shen, Yujun and Qu, Huamin},
  booktitle={CVPR},
  year={2025}
}

@inproceedings{yang2025layeranimate,
  author={Yang, Y. and Fan, L. and Lin, Z. and Wang, F. and Zhang, Z.},
  title={LayerAnimate: Layer-specific control for animation},
  booktitle={arXiv preprint arXiv:2501.08295},
  year={2025}
}

@inproceedings{xie2025physanimator,
  title={PhysAnimator: Physics-Guided Generative Cartoon Animation},
  author={Xie, Tianyi and Zhao, Yiwei and Jiang, Ying and Jiang, Chenfanfu},
  booktitle={CVPR},
  year={2025}
}

@inproceedings{chavez2025time,
  author={Chavez, Victor Fonte and Esteves, Claudia and Hayet, Jean-Bernard},
  title={Time-adaptive Video Frame Interpolation based on Residual Diffusion},
  booktitle={ACM SIGGRAPH},
  year={2025}
}

@inproceedings{hur2025high,
  title={High-Resolution Frame Interpolation with Patch-based Cascaded Diffusion},
  author={Hur, Junhwa and Herrmann, Charles and Saxena, Saurabh and Kontkanen, Janne and Lai, Wei-Sheng and Shih, Yichang and Rubinstein, Michael and Fleet, David J and Sun, Deqing},
  booktitle={AAAI},
  year={2025}
}

@inproceedings{shechtman2002stsr,
  title={Increasing space-time resolution in video},
  author={Shechtman, Eli and Caspi, Yaron and Irani, Michal},
  booktitle={ECCV},
  year={2002}
}

@inproceedings{kim2020fisr,
  title={Fisr: Deep joint frame interpolation and super-resolution with a multi-scale temporal loss},
  author={Kim, Soo Ye and Oh, Jihyong and Kim, Munchurl},
  booktitle={AAAI},
  year={2020}
}

@inproceedings{haris2020space,
  title={Space-time-aware multi-resolution video enhancement},
  author={Haris, Muhammad and Shakhnarovich, Greg and Ukita, Norimichi},
  booktitle={CVPR},
  year={2020}
}

@inproceedings{xu2021temporal,
  title={Temporal modulation network for controllable space-time video super-resolution},
  author={Xu, Gang and Xu, Jun and Li, Zhen and Wang, Liang and Sun, Xing and Cheng, Ming-Ming},
  booktitle={CVPR},
  year={2021}
}

@inproceedings{chen2023motif,
  title={Motif: Learning motion trajectories with local implicit neural functions for continuous space-time video super-resolution},
  author={Chen, Yi-Hsin and Chen, Si-Cun and Lin, Yen-Yu and Peng, Wen-Hsiao},
  booktitle={ICCV},
  year={2023}
}

@inproceedings{kim2025bf,
  title={BF-STVSR: B-Splines and Fourier---Best Friends for High Fidelity Spatial-Temporal Video Super-Resolution},
  author={Kim, Eunjin and Kim, Hyeonjin and Jin, Kyong Hwan and Yoo, Jaejun},
  booktitle={CVPR},
  year={2025}
}

@inproceedings{zhang2020video,
  title={Video frame interpolation without temporal priors},
  author={Zhang, Youjian and Wang, Chaoyue and Tao, Dacheng},
  booktitle={NeurIPS},
  year={2020}
}

@inproceedings{shen2020video,
  title={Video frame interpolation and enhancement via pyramid recurrent framework},
  author={Shen, Wang and Bao, Wenbo and Zhai, Guangtao and Chen, Li and Min, Xiongkuo and Gao, Zhiyong},
  booktitle={TIP},
  year={2020}
}

@inproceedings{shen2020blurry,
  title={Blurry video frame interpolation},
  author={Shen, Wang and Bao, Wenbo and Zhai, Guangtao and Chen, Li and Min, Xiongkuo and Gao, Zhiyong},
  booktitle={CVPR},
  year={2020}
}

@inproceedings{zhong2022animation,
  title={Animation from blur: Multi-modal blur decomposition with motion guidance},
  author={Zhong, Zhihang and Sun, Xiao and Wu, Zhirong and Zheng, Yinqiang and Lin, Stephen and Sato, Imari},
  booktitle={ECCV},
  year={2022}
}

@inproceedings{oh2022demfi,
  title={Demfi: deep joint deblurring and multi-frame interpolation with flow-guided attentive correlation and recursive boosting},
  author={Oh, Jihyong and Kim, Munchurl},
  booktitle={ECCV},
  year={2022}
}

@inproceedings{shang2023joint,
  title={Joint video multi-frame interpolation and deblurring under unknown exposure time},
  author={Shang, Wei and Ren, Dongwei and Yang, Yi and Zhang, Hongzhi and Ma, Kede and Zuo, Wangmeng},
  booktitle={CVPR},
  year={2023}
}

@inproceedings{yang2024latency,
  title={Latency correction for event-guided deblurring and frame interpolation},
  author={Yang, Yixin and Liang, Jinxiu and Yu, Bohan and Chen, Yan and Ren, Jimmy S and Shi, Boxin},
  booktitle={CVPR},
  year={2024}
}

@inproceedings{charbonnier1994two,
  title={Two deterministic half-quadratic regularization algorithms for computed imaging},
  author={Charbonnier, Pierre and Blanc-Feraud, Laure and Aubert, Gilles and Barlaud, Michel},
  booktitle={ICIP},
  year={1994}
}

@inproceedings{mathieu2015deep,
  title={Deep multi-scale video prediction beyond mean square error},
  author={Mathieu, Michael and Couprie, Camille and LeCun, Yann},
  booktitle={ICLR},
  year={2015}
}

@inproceedings{bojanowski2017optimizing,
  title={Optimizing the latent space of generative networks},
  author={Bojanowski, Piotr and Joulin, Armand and Lopez-Paz, David and Szlam, Arthur},
  booktitle={ICLR},
  year={2018}
}

@inproceedings{zabih1994non,
  author={Zabih, Ramin and Woodfill, John},
  title={Non-parametric local transforms for computing visual correspondence},
  booktitle={ECCV},
  year={1994}
}

@inproceedings{meister2018unflow,
  title={Unflow: Unsupervised learning of optical flow with a bidirectional census loss},
  author={Meister, Simon and Hur, Junhwa and Roth, Stefan},
  booktitle={AAAI},
  year={2018}
}

@misc{montgomery1994xiph,
  author={Montgomery, C.},
  title={Xiph.org video test media (derf's collection)},
  howpublished={Online, Available: https://media.xiph.org/video/derf/},
  year={1994}
}

@inproceedings{baker2011middlebury,
  author={Baker, S. and Scharstein, D. and Lewis, J. P. and Roth, S. and Black, M. J. and Szeliski, R.},
  title={A database and evaluation methodology for optical flow},
  booktitle={IJCV},
  year={2011}
}

@inproceedings{soomro2012ucf101,
  author={Soomro, K. and Zamir, A. R. and Shah, M.},
  title={UCF101: A dataset of 101 human actions classes from videos in the wild},
  booktitle={CRV},
  year={2012}
}

@inproceedings{geiger2012kitti,
  author={Geiger, A. and Lenz, P. and Urtasun, R.},
  title={Are we ready for autonomous driving? the kitti vision benchmark suite},
  booktitle={CVPR},
  year={2012}
}

@inproceedings{butler2012naturalistic,
  author={Butler, Daniel J and Wulff, Jonas and Stanley, Garrett B and Black, Michael J},
  title={A naturalistic open source movie for optical flow evaluation},
  booktitle={ECCV},
  year={2012}
}

@inproceedings{perazzi2016benchmark,
  author={Perazzi, F. and Pont-Tuset, J. and McWilliams, B. and Van Gool, L. and Gross, M. and Sorkine-Hornung, A.},
  title={A benchmark dataset and evaluation methodology for video object segmentation},
  booktitle={CVPR},
  year={2016}
}

@inproceedings{su2017adobe,
  author={Su, S. and Delbracio, M. and Wang, J. and Sapiro, G. and Heidrich, W. and Wang, O.},
  title={Deep video deblurring for hand-held cameras},
  booktitle={CVPR},
  year={2017}
}

@inproceedings{nah2017gopro,
  author={Nah, Seungjun and Hyun Kim, Tae and Mu Lee, Kyoung},
  title={Deep multi-scale convolutional neural network for dynamic scene deblurring},
  booktitle={CVPR},
  year={2017}
}

@inproceedings{bain2021frozen,
  title={Frozen in time: A joint video and image encoder for end-to-end retrieval},
  author={Bain, Max and Nagrani, Arsha and Varol, G{\"u}l and Zisserman, Andrew},
  booktitle={ICCV},
  year={2021}
}

@inproceedings{stergiou2024lavib,
  title={LAVIB: A Large-scale Video Interpolation Benchmark},
  author={Stergiou, Alexandros},
  booktitle={NeurIPS},
  year={2024}
}

@inproceedings{nan2024openvid,
  author={Nan, K. and Xie, R. and Zhou, P. and Fan, T. and Yang, Z. and Chen, Z. and Li, X. and Yang, J. and Tai, Y.},
  title={Openvid-1M: A large-scale high-quality dataset for text-to-video generation},
  booktitle={ICLR},
  year={2024}
}

@inproceedings{wang2004image,
  title={Image quality assessment: from error visibility to structural similarity},
  author={Wang, Zhou and Bovik, Alan C and Sheikh, Hamid R and Simoncelli, Eero P},
  booktitle={TIP},
  year={2004}
}

@inproceedings{mittal2012making,
  title={Making a “completely blind” image quality analyzer},
  author={Mittal, Anish and Soundararajan, Rajiv and Bovik, Alan C},
  booktitle={SPL},
  year={2012}
}

@inproceedings{heusel2017gans,
  author={Heusel, Martin and Ramsauer, Hubert and Unterthiner, Thomas and Nessler, Bernhard and Hochreiter, Sepp},
  title={Gans trained by a two time-scale update rule converge to a local nash equilibrium},
  booktitle={NeurIPS},
  year={2017}
}

@inproceedings{szegedy2016rethinking,
  title={Rethinking the inception architecture for computer vision},
  author={Szegedy, Christian and Vanhoucke, Vincent and Ioffe, Sergey and Shlens, Jon and Wojna, Zbigniew},
  booktitle={CVPR},
  year={2016}
}

@inproceedings{zhang2018lpips,
  author={Zhang, Richard and Isola, Phillip and Efros, Alexei A and Shechtman, Eli and Wang, Oliver},
  title={The unreasonable effectiveness of deep features as a perceptual metric},
  booktitle={CVPR},
  year={2018}
}

@inproceedings{chu2020learning,
  title={Learning temporal coherence via self-supervision for GAN-based video generation},
  author={Chu, Mengyu and Xie, You and Mayer, Jonas and Leal-Taix{\'e}, Laura and Thuerey, Nils},
  booktitle={ACM TOG},
  year={2020}
}

@inproceedings{unterthiner2018fvd,
  title={Towards accurate generative models of video: A new metric \& challenges},
  author={Unterthiner, Thomas and Van Steenkiste, Sjoerd and Kurach, Karol and Marinier, Raphael and Michalski, Marcin and Gelly, Sylvain},
  booktitle={ICLR Workshop},
  year={2019}
}

@inproceedings{carreira2017quo,
  title={Quo vadis, action recognition? a new model and the kinetics dataset},
  author={Carreira, Joao and Zisserman, Andrew},
  booktitle={CVPR},
  year={2017}
}

@inproceedings{liu2024fr,
  title={Fr$\backslash$'echet Video Motion Distance: A Metric for Evaluating Motion Consistency in Videos},
  author={Liu, Jiahe and Qu, Youran and Yan, Qi and Zeng, Xiaohui and Wang, Lele and Liao, Renjie},
  booktitle={ICML Workshop},
  year={2024}
}

@inproceedings{li2019quality,
  title={Quality assessment of in-the-wild videos},
  author={Li, Dingquan and Jiang, Tingting and Jiang, Ming},
  booktitle={Proceedings of the 27th ACM international conference on multimedia},
  year={2019}
}

@inproceedings{ding2020image,
  title={Image quality assessment: Unifying structure and texture similarity},
  author={Ding, Keyan and Ma, Kede and Wang, Shiqi and Simoncelli, Eero P},
  booktitle={TPAMI},
  year={2020}
}

@inproceedings{ghildyal2022shift,
  title={Shift-tolerant perceptual similarity metric},
  author={Ghildyal, Abhijay and Liu, Feng},
  booktitle={ECCV},
  year={2022}
}

@inproceedings{danier2022flolpips,
  author={Danier, Duolikun and Zhang, Fan and Bull, David},
  title={FloLPIPS: A bespoke video quality metric for frame interpolation},
  booktitle={PCS},
  year={2022}
}

@inproceedings{huang2024vbench,
  author={Huang, Ziqi and He, Yinan and Yu, Jiashuo and Zhang, Fan and Si, Chenyang and Jiang, Yuming and Zhang, Yuanhan and Wu, Tianxing and Jin, Qingyang and Chanpaisit, Nattapol and others},
  title={Vbench: Comprehensive benchmark suite for video generative models},
  booktitle={CVPR},
  year={2024}
}

@inproceedings{ancuti2017locally,
  title={Locally adaptive color correction for underwater image dehazing and matching},
  author={Ancuti, Codruta O and Ancuti, Cosmin and De Vleeschouwer, Christophe and Garcia, Rafael},
  booktitle={CVPRW},
  year={2017}
}

@article{zhu2025waterwave,
  title={WaterWave: Bridging Underwater Image Enhancement into Video Streams via Wavelet-based Temporal Consistency Field},
  author={Zhu, Qi and Zhang, Jingyi and Zheng, Naishan and Yu, Wei and Zhang, Jinghao and Ji, Deyi and Zhao, Feng},
  booktitle={arXiv preprint arXiv:2512.05492},
  year={2025}
}

@inproceedings{tang2024neural,
  title={Neural underwater scene representation},
  author={Tang, Yunkai and Zhu, Chengxuan and Wan, Renjie and Xu, Chao and Shi, Boxin},
  booktitle={CVPR},
  year={2024}
}

@inproceedings{yu2022deep,
  title={Deep bayesian video frame interpolation},
  author={Yu, Zhiyang and Zhang, Yu and Xiang, Xujie and Zou, Dongqing and Chen, Xijun and Ren, Jimmy S},
  booktitle={ECCV},
  year={2022},
}

@inproceedings{li2022all,
  title={All-in-one image restoration for unknown corruption},
  author={Li, Boyun and Liu, Xiao and Hu, Peng and Wu, Zhongqin and Lv, Jiancheng and Peng, Xi},
  booktitle={CVPR},
  year={2022}
}

@inproceedings{wu2025content,
  title={Content-Aware Transformer for All-in-one Image Restoration},
  author={Wu, Gang and Jiang, Junjun and Jiang, Kui and Liu, Xianming},
  booktitle={arXiv preprint arXiv:2504.04869},
  year={2025}
}

@inproceedings{ai2024multimodal,
  title={Multimodal prompt perceiver: Empower adaptiveness generalizability and fidelity for all-in-one image restoration},
  author={Ai, Yuang and Huang, Huaibo and Zhou, Xiaoqiang and Wang, Jiexiang and He, Ran},
  booktitle={CVPR},
  year={2024}
}

@inproceedings{nag20252,
  title={In-2-4D: Inbetweening from Two Single-View Images to 4D Generation},
  author={Nag, Sauradip and Cohen-Or, Daniel and Zhang, Hao and Mahdavi-Amiri, Ali},
  booktitle={arXiv preprint arXiv:2504.08366},
  year={2025}
}

@inproceedings{park2023temporal,
  title={Temporal interpolation is all you need for dynamic neural radiance fields},
  author={Park, Sungheon and Son, Minjung and Jang, Seokhwan and Ahn, Young Chun and Kim, Ji-Yeon and Kang, Nahyup},
  booktitle={CVPR},
  year={2023}
}

@inproceedings{zheng2023neuralpci,
  title={Neuralpci: Spatio-temporal neural field for 3d point cloud multi-frame non-linear interpolation},
  author={Zheng, Zehan and Wu, Danni and Lu, Ruisi and Lu, Fan and Chen, Guang and Jiang, Changjun},
  booktitle={CVPR},
  year={2023}
}
\vspace{-43pt}

\begin{IEEEbiography}[{\includegraphics[width=1in,height=1.25in,clip,keepaspectratio]{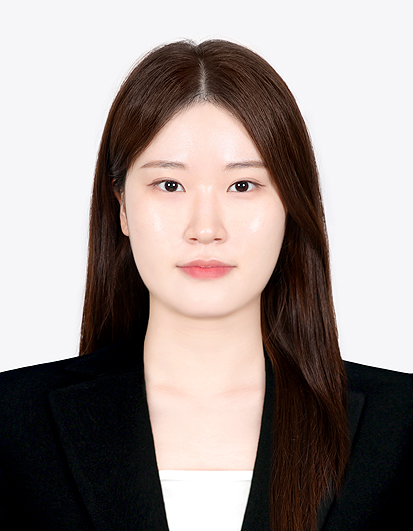}}]{Dahyeon Kye}
is currently working toward the Ph.D. degree in the Graduate School of Advanced Imaging Science, Multimedia \& Film (GSAIM) at Chung-Ang University (CAU), Seoul, South Korea. She received the B.E. degree in computer engineering from Sejong University (SJU), Seoul, South Korea, in 2023, and the M.E. degree from GSAIM, Chung-Ang University, under the supervision of Prof. Jihyong Oh. Her research interests include low-level vision and generative AI.
\end{IEEEbiography}
\vspace{-40pt}

\begin{IEEEbiography}[{\includegraphics[width=1in,height=1.25in,clip,keepaspectratio]{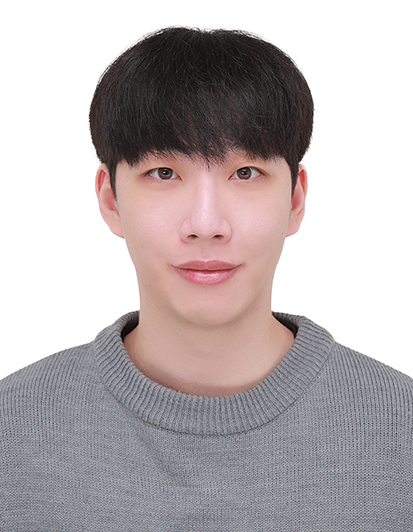}}]{Changhyun Roh} is an M.S. student at The Graduate School of Advanced Imaging Science (GSAIM), Chung-Ang University (CAU), advised by Prof. Jihyong Oh at CMLAB. He received B.S. degree in engineering from Tech University of Korea (TUKOREA). His current researsch interets include generative models, with an emphasis on diffusion-based image generation and personalization.
\end{IEEEbiography}
\vspace{-40pt}

\begin{IEEEbiography}[{\includegraphics[width=1in,height=1.25in,clip,keepaspectratio]{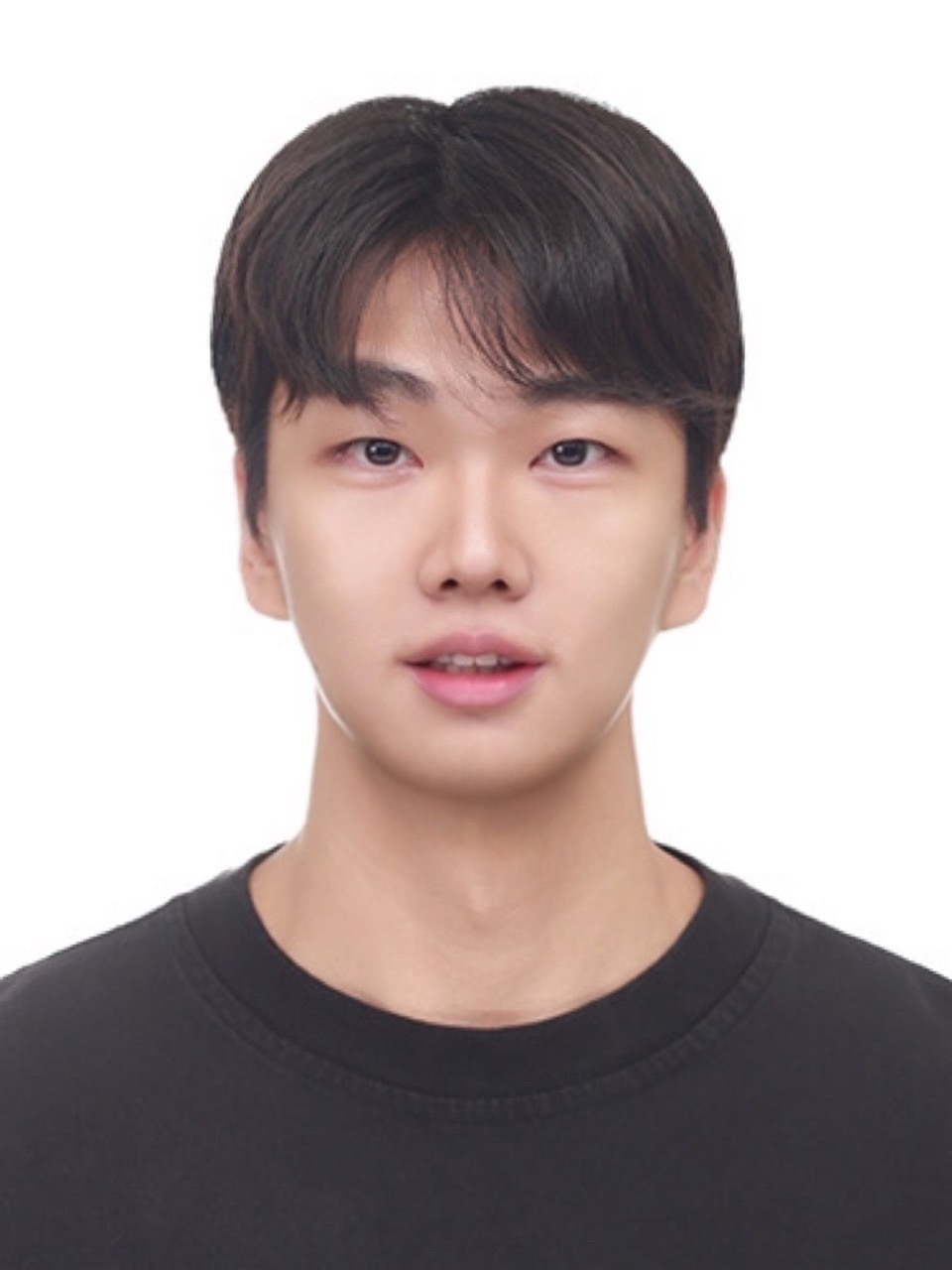}}]{Sukhun Ko} received the B.S. degree in Big Data Convergence and began pursuing the M.S. degree in Imaging Science at the Graduate School of Advanced Imaging Science, Multimedia \& Film (GSAIM), Chung-Ang University, Seoul, South Korea, in March 2025. His research interests include low-level vision tasks, image generation, and implicit neural representations. He is currently a member of the Creative Vision and Multimedia Lab (CMLab, \url{https://cmlab.cau.ac.kr/}) at Chung-Ang University.
\end{IEEEbiography}
\vspace{-40pt}

\begin{IEEEbiography}[{\includegraphics[width=1in,height=1.25in,clip,keepaspectratio]{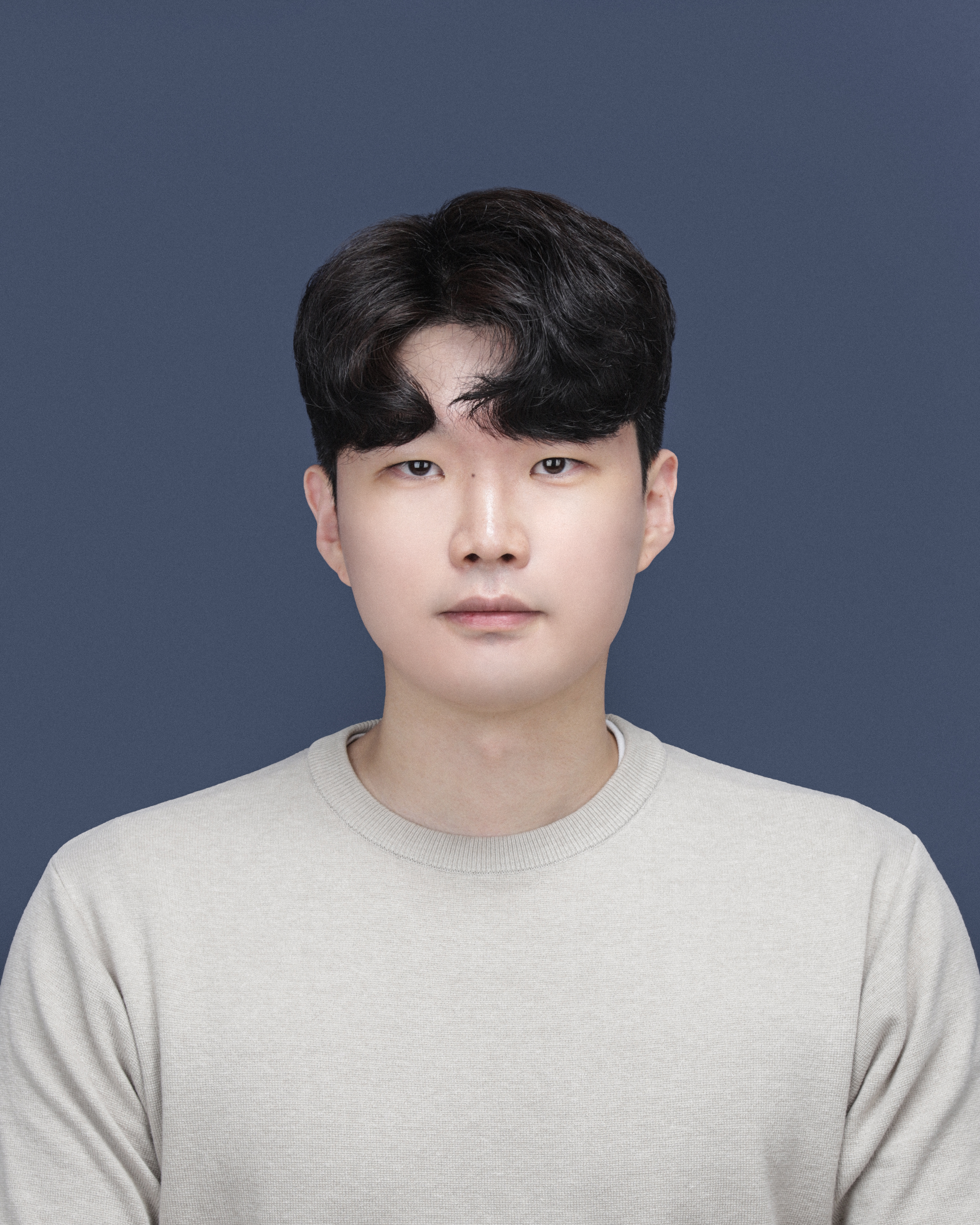}}]{Chanho Eom} is an Assistant Professor at GSAIM, Chung-Ang University in Seoul, Korea. He received his B.S. and Ph.D. degrees in Electrical and Electronic Engineering from Yonsei University in 2017 and 2023, respectively. He previously worked as a researcher at the Samsung Advanced Institute of Technology (SAIT). His research interests include computer vision and deep learning, particularly in retrieval, person re-identification, and video analysis, both in theory and applications.
\end{IEEEbiography}
\vspace{-40pt}

\begin{IEEEbiography}[{\includegraphics[width=1in,height=1.25in,clip,keepaspectratio]{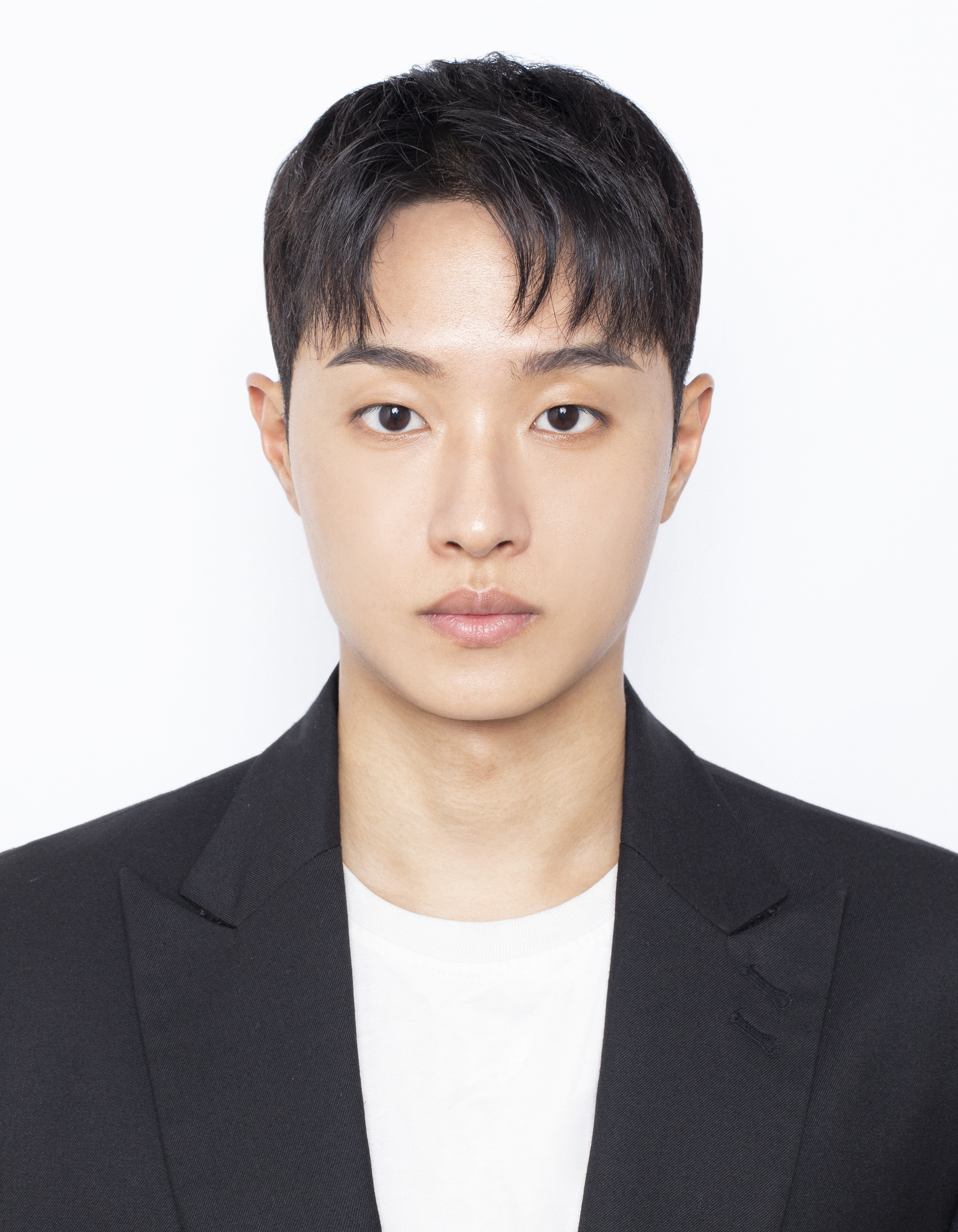}}]{Jihyong Oh} is an Assistant Professor at the Graduate School of Advanced Imaging Science, Multimedia \& Film (GSAIM) at Chung-Ang University (CAU; Seoul, South Korea) and has led the Creative Vision and Multimedia Lab (CMLab: \url{https://cmlab.cau.ac.kr/}) since September 2023. He received his B.E., M.E., and Ph.D. degrees in Electrical Engineering from KAIST in 2017, 2019, and 2023, respectively. He previously worked as a post-doctoral researcher at VICLAB of KAIST and as a research intern at Meta (Facebook) Reality Labs in 2022. His research primarily focuses on low-level vision, image/video restoration, 3D vision, and generative AI.
\end{IEEEbiography}

\vspace{-33pt}

\end{document}